\documentclass[preprint, 5p]{elsarticle}
\usepackage{times}
\usepackage{amsmath,amsfonts,bm}
\usepackage{url}
\usepackage{hyperref}
\usepackage{graphicx}
\usepackage{refcount}
\usepackage{xstring}
\usepackage{scrextend} 
\usepackage{geometry}
\usepackage{adjustbox}
\usepackage{amsthm}
\usepackage{amssymb}
\usepackage{mathtools}
\usepackage[table]{xcolor}
\usepackage{xcolor}
\usepackage{multirow}
\usepackage{stmaryrd}
\usepackage{standalone}
\usepackage{natbib}
\usepackage{makecell}
\usepackage{rotating}
\usepackage[capitalise]{cleveref}
\usepackage{amsthm}
\usepackage{thmtools, thm-restate}
\usepackage{xcolor}
\usepackage[cal=boondoxo]{mathalfa} 
\usepackage[normalem]{ulem} 
\usepackage{enumitem}
\usepackage{float}
\usepackage{xr}
\usepackage{balance}
\usepackage{xspace}
\usepackage{url}
\usepackage{expl3}
\usepackage{amsmath}
\usepackage{amssymb}
\usepackage{mathtools}
\usepackage{dsfont}
\usepackage{upgreek}
\usepackage{bbm}
\usepackage{graphicx}
\usepackage{epstopdf}
\usepackage{grffile}
\usepackage{tikz}
\usepackage{float}
\usepackage{svg}
\usepackage{algorithm}
\usepackage{algorithmic}
\usepackage[labelfont = bf]{caption}
\usepackage{subcaption}
\usepackage{relsize}  
\usepackage{makecell}
\usepackage{times}
\usepackage{url}
\usepackage{hyperref}
\usepackage{graphicx}
\usepackage[capitalise]{cleveref}
\usepackage{refcount}
\usepackage{xstring}
\usepackage{geometry}
\usepackage{amsmath}
\usepackage{adjustbox}
\usepackage{amsthm}
\usepackage{amssymb}
\usepackage{mathtools}
\usepackage[table]{xcolor}
\usepackage{xcolor}
\usepackage{multirow}
\usepackage{stmaryrd}
\usepackage{standalone}
\usepackage{natbib}
\usepackage{rotating}

\newtheorem{definition}{Definition}

\newtheorem{property}{Property}

\newtheorem{corollary}{Corollary}

\definecolor{blue1}{HTML}{2851CC}
\definecolor{red1}{HTML}{FF0000}
\definecolor{green1}{HTML}{3BB273}
\definecolor{orange1}{HTML}{F96C39}
\definecolor{pink1}{HTML}{F056BF}
\definecolor{gray1}{HTML}{696A62}
\definecolor{gray2}{HTML}{E3E4DB}

\definecolor{todocolor}{HTML}{F17105}
\definecolor{newcolor}{HTML}{682ACB}
\definecolor{newakcolor}{HTML}{07B053}

\newcommand{\norm}[1]{\left\lVert#1\right\rVert}
\newcommand{\abs}[1]{\left|#1\right|}

\newcommand{\indep}{\perp \!\!\!\! \perp}
\newcommand{\R}{\mathbb{R}}
\newcommand{\supp}{\operatorname{supp}}

\newcommand{\mystd}[1]{{{\color{black} \scriptsize ({#1})}}}
\newcommand{\mybest}[1]{\cellcolor{gray1!25}\textbf{#1}}
\newcommand{\mysecond}[1]{\cellcolor{gray1!15}\underline{#1}}

\newcommand{\smallg}[1]{{\small #1}}

\newcounter{marginNoteCounter}

\newcounter{marginNoteCounterGw}

\newcommand{\ie}   			{i.e.\@\xspace}
\newcommand{\eg}   			{e.g.\@\xspace}

\newcommand{\ind}       {\mathds{1}}
\newcommand{\Ind}[1]    {\ind_{\{#1\}} }

\newcommand{\algComment}[1] 	{{\hfill{$\displaystyle \triangleright$}~\,{#1}}}
\renewcommand*{\top}{{\mkern-1.5mu\mathsf{T}}}
\newcommand{\dist} {\operatorname{dist}}

\newcommand{\inlinetitle}[2]  {\vspace{4pt}\noindent\textbf{\emph{#1}{#2}}}

\newcommand{\tr}[1]{\operatorname{tr}(#1)}

\newcommand{\padded}[1] {\,#1\,}
\newcommand{\void}[1] {\padded{\cdot}}

\makeatletter
\def\adl@drawiv#1#2#3{%
        \hskip.5\tabcolsep
        \xleaders#3{#2.5\@tempdimb #1{1}#2.5\@tempdimb}%
                #2\z@ plus1fil minus1fil\relax
        \hskip.5\tabcolsep}
\newcommand{\cmidruledashed}[1]{%
  \noalign{\vskip\aboverulesep
           \global\let\@dashdrawstore\adl@draw
           \global\let\adl@draw\adl@drawiv}
  \cdashline{#1}%
	\noalign{\vskip-\belowrulesep}
	\noalign{\vskip-\belowrulesep}
}
\makeatother

\makeatletter
\newcommand{\pushright}[1]{\ifmeasuring@#1\else\omit\hfill$\displaystyle#1$\fi\ignorespaces}
\newcommand{\pushleft}[1]{\ifmeasuring@#1\else\omit$\displaystyle#1$\hfill\fi\ignorespaces}
\makeatother

\makeatletter
\let\KV@Gin@trim@old\KV@Gin@trim
\define@key{Gin}{trim}{%
  \begingroup
  \edef\x{\endgroup
    \noexpand\setkeys{Gin}{trim@old={#1}}%
  }\x
}
\makeatother
\makeatletter
\let\KV@Gin@viewport@old\KV@Gin@viewport
\define@key{Gin}{viewport}{%
  \begingroup
  \edef\x{\endgroup
    \noexpand\setkeys{Gin}{viewport@old={#1}}%
  }\x
}
\makeatother

\newcommand{\VOID}[1]{}

\newcommand{\overbar}[1] {\mkern 1.5mu\overline{\mkern-3.5mu#1\mkern-1.0mu}\mkern 1.5mu}

\def\eqref#1{equation~\ref{#1}}

\def\1{\bm{1}}

\def\rmA{{\mathbf{A}}}
\def\rmB{{\mathbf{B}}}

\def\rmD{{\mathbf{D}}}

\def\rmI{{\mathbf{I}}}

\def\rmK{{\mathbf{K}}}

\def\rmM{{\mathbf{M}}}
\def\rmN{{\mathbf{N}}}

\def\rmP{{\mathbf{P}}}
\def\rmQ{{\mathbf{Q}}}

\def\rmT{{\mathbf{T}}}
\def\rmU{{\mathbf{U}}}
\def\rmV{{\mathbf{V}}}
\def\rmW{{\mathbf{W}}}
\def\rmX{{\mathbf{X}}}

\def\rmSigma{{\mathbf{\Sigma}}}
\def\rmLambda{{\mathbf{\Lambda}}}
\def\rmPi{{\mathbf{\Pi}}}

\def\rmPsi{{\mathbf{\Psi}}}
\def\rmPhi{{\mathbf{\Phi}}}

\def\mXi{{\bm{\Xi}}}
\def\mLambda{{\bm{\Lambda}}}
\def\mSigma{{\bm{\Sigma}}}

\DeclareMathAlphabet{\mathsfit}{\encodingdefault}{\sfdefault}{m}{sl}
\SetMathAlphabet{\mathsfit}{bold}{\encodingdefault}{\sfdefault}{bx}{n}

\def\gD{{\mathcal{D}}}

\def\gN{{\mathcal{N}}}

\def\gS{{\mathcal{S}}}

\def\gU{{\mathcal{U}}}

\def\sN{{\mathbb{N}}}

\newcommand{\E}{\mathbb{E}}

\DeclareMathOperator*{\argmax}{arg\,max}

\hypersetup{
  colorlinks=true,breaklinks,
  linktoc=section,
  linkcolor=red1,
  citecolor=blue1,
  urlcolor=green1,
  }

\crefformat{appendix}{#2#1#3}
\crefname{equation}{Eq.}{Eq.}
\crefname{section}{Sec.}{Sec.}
\crefname{algorithm}{Alg.}{Alg.}
\crefname{table}{Tab.}{Tab.}
\crefname{definition}{Def.}{Def.}
\crefname{theorem}{Thm.}{Thm.}
\crefname{property}{Prop.}{Prop.}
\creflabelformat{equation}{#2\textup{#1}#3}

\newcommand{\MDT}{MDT\xspace}
\newcommand{\MDTs}{MDTs\xspace}
\newcommand{\MDTcvx}{\textsc{MDT-Cvx-Rand}\xspace}
\newcommand{\MDTrand}{\textsc{MDT-Rand}\xspace}
\newcommand{\MDTchs}{\textsc{MDT-Direct}\xspace}

\newcommand{\MDTCST}{\textsc{MDT-Cst}\xspace}
\newcommand{\MDTbsc}{\textsc{MDT-Bsc}\xspace} 

\newcommand{\CrD}{\textsc{Cr-Diff}\xspace}
\newcommand{\ComD}{\textsc{Com-Diff}\xspace}
\newcommand{\IDM}{\textsc{ID}\xspace}
\newcommand{\ADM}{\textsc{AD}\xspace}
\newcommand{\DM}{\textsc{DM}\xspace}

\newcommand{\pADM}{\textsc{p-AD}\xspace}
\newcommand{\MVDM}{\textsc{MVD}\xspace}

\newcommand{\Vnum}{V}
\newcommand{\Pset}{\mathcal{P}}
\newcommand{\kernelFunc}{\kappa}

\newcommand*{\defeq}{=}

\newcommand*{\iter}{t}
\newcommand*{\Wnorm}{\rmA}
\renewcommand{\epsilon}{\varepsilon}

\newcommand{\Psetc} {{\Pset_\text{c}}}
\newcommand{\Psetcvx} {{\Pset_{\textup{cvx}}}}
\newcommand{\Psetinit} {{\Pset_{\textup{init}}}}
\newcommand{\Vnuminit} {{\Vnum_{\textup{init}}}}

\newcommand{\appref}[1]{%
  Appendix~\hyperref[#1]{\!\!\!
    \StrBehind{\getrefnumber{#1}}{Appendix }[\appref@letter]%
    \appref@letter%
  }%
}

\makeatletter
\def\Wopnorm{\@ifnextchar[{\@Wopnormwith}{\@Wopnormwithout}}
\def\@Wopnormwith[#1][#2]{\overbar{W}_{#1;#2}}
\def\@Wopnormwithout#1{\overbar{W}_{#1}}
\def\P{\@ifnextchar[{\@Pwith}{\@Pwithout}}
\def\@Pwith[#1][#2]{\bm{P}_{#1;#2}}
\def\@Pwithout#1{\bm{P}_{#1}}
\def\Wop{\@ifnextchar[{\@Wopwith}{\@Wopwithout}}
\def\@Wopwith[#1][#2]{\rmW_{#1;#2}}
\def\@Wopwithout#1{\rmW_{#1}}
\makeatother
\makeatletter

\journal{Information Fusion}
\begin{document}
\begin{frontmatter}
  \title{Multi-view diffusion geometry using intertwined diffusion trajectories}

  \author[1]{Gwendal Debaussart-Joniec\corref{cor1}}
  \author[1]{Argyris Kalogeratos\corref{cor1}}

  \cortext[cor1]{Corresponding authors: \{name.surname\}@institution.org}

  \affiliation[1]{organization={École Normale Supérieure Paris-Saclay, Centre Borelli, CNRS},
  addressline={4 Av. des Sciences},
  postcode={91190},
  city={Gif-Sur-Yvette},
  country={France}
	}

  \date{}

  \begin{abstract}
  This paper introduces a comprehensive unified framework for constructing multi-view diffusion geometries through intertwined \emph{multi-view diffusion trajectories} (MDTs), a class of inhomogeneous diffusion processes that iteratively combine the random walk operators of multiple data views. Each \MDT defines a trajectory-dependent diffusion operator with a clear probabilistic and geometric interpretation, capturing over time the interplay between data views. Our formulation encompasses existing multi-view diffusion models, while providing new degrees of freedom for view interaction and fusion. We establish theoretical properties under mild assumptions, including ergodicity of both the point-wise operator and the process in itself. We also derive \MDT-based diffusion distances, and associated embeddings via singular value decompositions. Finally, we propose various strategies for learning \MDT operators within the defined operator space, guided by internal quality measures. Beyond enabling flexible model design, MDTs also offer a neutral baseline for evaluating diffusion-based approaches through comparison with randomly selected \MDTs. Experiments show the practical impact of the \MDT operators in a manifold learning and data clustering context.
  \end{abstract}


\end{frontmatter}

\section{Introduction}

Progress in data acquisition and distributed computing architectures has allowed widespread access to multi-view data, where each view provides a different representation of the same data objects. In such contexts, views may correspond to multiple data modalities (\eg text, images, sensor data), multiple measuring instruments, or may be produced via transformations of an original input data view (\eg projections, texture-edge-frequency decompositions and related construction). Two approaches for combining multi-view representations exist: \emph{representation fusion}, which seeks a single common representation, and \emph{representation alignment}, which aims to produce complementary representations informed by the multiple views. This work will focus on representation fusion. Several approaches have been developed for this purpose; see surveys

Random walk-based representation learning relies on pairwise affinity measures to form graphs over point-cloud data, followed by a random walk defined via the random walk Laplacian. This operator is tightly connected to the graph-diffusion operator that models heat flow over the discrete domain encoded by the graph's vertices and edges. Powers of the transition matrix reveal relationships between points at different time scales and effectively denoises the data by pushing representations toward prominent directions captured by low-frequency eigenvectors \citep{dijk_recovering_2018, sevi2022clustering}. Approaches such as Diffusion Maps (DM) \cite{coifman_diffusion_2006} belong to the kernel eigenmap family of manifold learning methods that use spectral decompositions to reveal geometric structure. In the single-view setting, it is known that under mild conditions the diffusion operator converges to the heat kernel on a continuous underlying manifold \citep{coifman_diffusion_2006}. Since the heat kernel encodes rich geometric information \citep{berline_heat_1992, grigoryan_heat_2009}, this observation is key for the diffusion-based framework. Diffusion operators have been employed in manifold learning \citep{coifman_diffusion_2006}, clustering \citep{sevi_generalized_2025, sevi2022clustering}, denoising, and visualization \citep{fernandez_diffusion_2015, haghverdi_diffusion_2015}.

The reliability and interpretability of single-view diffusion methods have motivated their extension to the multi-view setting. A central challenge in this context is to design that can meaningfully integrate heterogeneous information from multiple views. Existing approaches typically rely on fixed rules for coupling view-specific diffusion operators into a single composite operator. Multi-view Diffusion Maps (\MVDM) \citep{lindenbaum_multi-view_2020} use products of view-specific kernels to promote cross-view transitions, following multi-view spectral clustering ideas \citep{de_sa_spectral_2018} that encode interactions through block structures built from cross-kernel terms.Alternating Diffusion (\ADM) \citep{katz_alternating_2019} constructs a composite operator by alternating between view-specific transition matrices at each step, effectively enforcing the random walk to switch views iteratively. Integrated Diffusion (\IDM) \citep{kuchroo_multimodal_2022} first applies diffusion within each view independently to denoise the data, and then combines the resulting operators to capture inter-view relationships. Other methods, such as Cross-Diffusion (\CrD) \citep{wang_unsupervised_2012} and Composite Diffusion (\ComD) \citep{shnitzer_recovering_2018}, utilize both forward and backward diffusion operators to model complex interactions between views. Finally, diffusion-inspired intuition has also been applied in the multi-view setting through a sequence of graph shift operators interpreted as convolutional filters \citep{butler_convolutional_2022}.

\begin{figure}[t]
  \centering
    \includegraphics[width=0.49\textwidth]{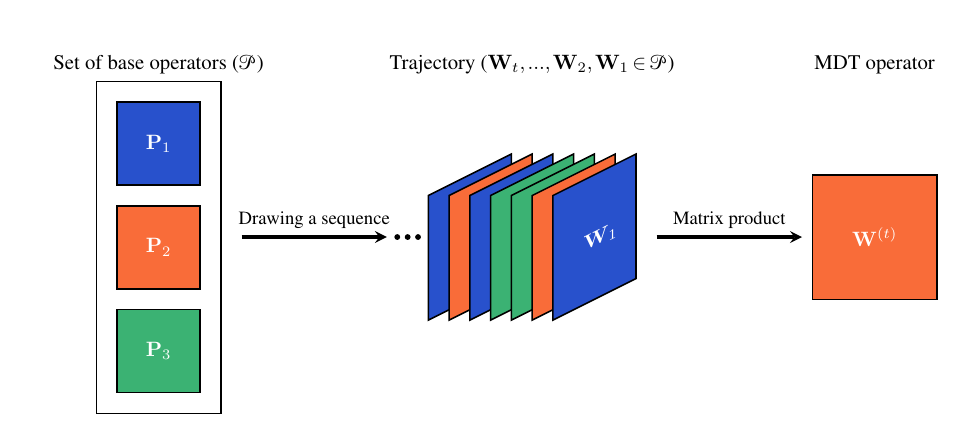}
    \caption{\textbf{MDT Operator.} Constructing a diffusion operator based on a multi-view diffusion trajectory (\MDT). From the set of base random walk operators $\Pset$ appearing on the left, a sequence $(\rmW_i)_{i \in \sN}$ of operators is drawn. The final \MDT operator $\rmW^{(t)}$ is defined as the left-product of the sequence of operators.}
  \label{fig:MDT}
\end{figure}

While the proposed work focuses on operator-based techniques  estimating the global behavior of the random walk through iterated transition matrices, a separate family operates by sampling local random walks (\ie explicit vertex sequences), such as node2vec \citep{grover_node2vec_2016} and its variants. Extensions to multi-graph settings have also been explored \citep{roy_learning_2020, huang_random_2022, valentini_hetnode2vec_2023}. Our main contributions are summarized below:
\begin{itemize}[leftmargin=1.5em, itemsep=0em, topsep=0.3em]
    \item \textbf{Multi-view Diffusion Trajectories (\MDTs).}~We introduce a flexible framework for defining diffusion geometry in multi-view settings based on time-inhomogeneous diffusion processes that intertwine the random walk operators of different views. The construction relies on an operator space derived from the input data. Each \MDT defines a diffusion operator that admits a clear probabilistic and geometric interpretation. Our framework is detailed in \cref{sec:mdt}, an illustrative example is given in \cref{fig:MDT}.
    \item \textbf{Unified theoretical foundation.}~We establish theoretical properties under mild conditions: ergodicity of point-wise operators, ergodicity of the \MDT process, and the existence of trajectory-dependent diffusion distances and embeddings obtained via singular value decompositions. Several existing multi-view diffusion schemes (\eg Alternating Diffusion \cite{katz_alternating_2019}, Integrated Diffusion \cite{kuchroo_multimodal_2022}) arise naturally as special cases.
    \item \textbf{Learning \MDT operators.}~The proposed formulation enables learning diffusion operators within the admissible operator space. We develop unsupervised strategies guided by internal quality measures and examine configurations relevant for clustering and manifold learning. Experiments on synthetic and real-world datasets show that \MDTs match or exceed the performance of state-of-the-art diffusion-based fusion methods.
   \item \textbf{Neutral baseline for diffusion-based evaluation.}~As the MDT operator space includes established approaches as particular cases, we suggest using randomly selected \MDTs as a neutral, principled baseline for the evaluation of diffusion-based multi-view methods. Experimental results show that, provided the time parameter ($t$) is well-chosen in an unsupervised way, random \MDTs achieve competitive and often superior performance compared to many sophisticated existing models. This highlights the importance of comparing against such baselines in future studies.
\end{itemize}

\section{Notations and background}\label{sec:background}
Denote $e_i$ the vector having all $0$'s and a $1$ in its $i$-th component, and $\mathbf{1}$ is the all-ones vector. Suppose that we have $N$ datapoints represented by a multi-view dataset $\{\rmX_v \}_{v=1}^\Vnum$ that is a collection of individual datasets $\rmX_v \in \R^{d_v \times N}$, where $d_v$ is the dimension of the data view $v$. We denote by $x_j$ the $j$-th individual object of the dataset.

\subsection{Diffusion Geometry}\label{sec:diffusion-geometry-one-view}
Given a dataset $\rmX \in \R^{d \times N}$, we denote by $\kernelFunc$ the associated kernel function, verifying that $\kernelFunc : \R^{d}\times\R^{d} \to \R_+$ is symmetric and self-positive, $\kernelFunc (x, x) > 0$. A common choice for the selection of a kernel is the ``Gaussian kernel'' $k_\sigma (x,y) = \exp(-\frac{\dist(x,y)}{2 \sigma^2})$, where $\sigma$ is a bandwidth parameter. We denote by $\rmK$ the kernel matrix obtained from this kernel $(\rmK)_{i j} = \kernelFunc(x_i,x_j)$. Let $\rmD : (\rmD)_{ii} = \sum_j (\rmK)_{ij}$, be the vertex degree matrix. Normalizing $\rmK$ by $\rmD$ defines the following row-stochastic matrix:
\begin{equation} \label{eqdef:diffusion_operator}
  \rmP \defeq \rmD^{-1}\rmK \in \R^{N \times N}.
\end{equation}
The operator $\rmP$ is known as the diffusion or random walk operator, as it defines a homogeneous Markov Chain on $G$, and $\rmP^t$ is the operator after $t$ consecutive steps. This operator can be used to visualize or infer the intrinsic organization of the data (\eg through clustering) \citep{coifman_diffusion_2006,nadler_diffusion_2006,barth_diffusion_2008}. Let $\gD^t(i,j)$ be the \emph{diffusion distance} \citep{coifman_diffusion_2006} defined by:
\begin{equation} \label{eqdef:diffusion_distance}
  \left(\gD^t(i, j)\right)^2 \defeq \sum_{k = 1,...,N} \frac{1}{\pi(k)}\left[(\rmP^t)_{i k} - (\rmP^t)_{jk} \right]^2,
\end{equation}
where $\pi(k)$ is the stationary distribution of the Markov chain. The main feature of the diffusion distance is that it uses the structure of the data to provide smoothed the distance measurements, modulated by the time parameter, that are less sensitive to noise and distortions. As $\rmP$ is a normalized symmetric matrix, it admits a real eigenvalue decomposition $\rmP = \rmU \rmLambda \rmU^\top$. This eigenvalue decomposition can be used to embed the datapoints into an Euclidean space via the mapping:
\begin{equation}
  \Psi^t (x_i) \defeq e_i^\top \rmU \rmLambda^t.
\end{equation}
For this new embedding, called \emph{Diffusion Map}~(\DM), it has been shown that the Euclidean distance in the embedded space is equal to the diffusion distance \cite{coifman_diffusion_2006}, more formally:
\begin{equation}
  \left(\gD^t(i, j)\right)^2 = \norm{\Psi^t(x_i) - \Psi^t(x_j)}^2.
\end{equation}
This shows that the diffusion distance is indeed a metric distance. To achieve a compact embedding of the datapoints, the eigenmap can be truncated up to some order, by taking only the first eigenvalues. The choice of how many eigenvalues should be kept is a difficult problem, and heuristics based either on the spectral decay of the eigenvalues (\eg the elbow rule) or on the number of clusters (in the context of clustering) are usually employed. Similar heuristics are also employed for the selection of diffusion time parameter ($t$), which controls the scale at which data geometry is explored. The diffusion maps procedure is summarized in \cref{alg:DM}.

\begin{algorithm}[t]
  \small
  \caption{\textbf{--} Diffusion Maps for a single data view}\label{alg:DM}
  \begin{algorithmic}
    \STATE {\bfseries Input:}
    $\rmX$: a high dimensional dataset \\
    \STATE \hspace*{\algorithmicindent}\hspace*{\algorithmicindent}\quad $l$: dimension of the output \\
    \STATE \hspace*{\algorithmicindent}\hspace*{\algorithmicindent}\quad $t$: number of steps \\
    \STATE {\bfseries Output:} $\rmPsi^{l,t}$: a low dimensional embedding of the datapoints \\
    \vspace{1.3mm}
    \hrule
    \vspace{1.3mm}
    \STATE {Construct the diffusion operator $\rmP$ according to \cref{eqdef:diffusion_operator}}
    \STATE {Compute the spectral decomposition $\rmPhi, \rmLambda$ of $\rmP$}
    \STATE {Define a new embedding of the datapoints according to \\ \quad $\rmPsi^{l,t}(x_i) = [\lambda^t_1 \phi_1(i), ..., \lambda_l^t \phi_l(i)]^\top$}
	\STATE \textbf{return} $\rmPsi^{l,t}$
  \end{algorithmic}
\end{algorithm}

\subsection{Extensions for multiple views}\label{sec:existing-methods}
The process described in \cref{sec:diffusion-geometry-one-view} can be applied independently to each available data view, $v=1,...,\Vnum$. In that case, all important elements are indexed accordingly, \ie $\rmX_v$, $d_v$, $\kernelFunc_v$, $\dist_v$, $\rmP_v$, $\rmK_v$, $\Psi_v$, and $\gD_v^t$. Eventually, doing so yields to the set of operators $\Psetc \defeq \{\rmP_v \}_{v=1}^\Vnum$, which we call \emph{canonical diffusion operator set}. With the notion of canonical set at hand, we can see that existing works define diffusion maps for multi-view data mainly by combining the transition matrices of that set in a composite operator $\rmQ$. This defines a standard homogeneous Markov chain, whereas each step in fact consists of multiple successive steps using the different transition matrices in $\Psetc$. For example, in a two-view setting, suppose the set $\Psetc = \{\rmP_1, \rmP_2\}$ and the composite operator $\rmQ = \rmP_1 \rmP_1 \rmP_2$, then its iteration for $t=2$ would be $\rmQ^{2} = \rmP_1 \rmP_1 \rmP_2 \rmP_1 \rmP_1 \rmP_2$. The analysis of operators constructed this way can be done using the standard homogeneous Markov chain theory. On the downside, this requires a prefixed fusion of the views into $\rmQ$, leading to a lack of flexibility. A number of options have been proposed in the literature, whereas there is hardly any theoretical foundation how to make such a crucial choice. \cref{tab:existing-methods} gives an overview of those methods. For simplicity, in what follows we will be expressing those methods using $2$ views.

Alternating Diffusion (\ADM) \citep{lederman_alternating_2015} builds the operator $\rmQ_{\text{AD}} = \rmP_1\rmP_2$, which has been analyzed theoretically in subsequent works \citep{lederman_learning_2018, talmon_latent_2017, ding_spectral_2020} and also in terms of computational efficiency \citep{yeh_landmark_2024} through the usage of `landmark' alternating diffusion. The transition matrix induced by \ADM can be viewed as a Markov chain that alternates between the first and second views at each iteration. Integrated Diffusion (\IDM) \cite{kuchroo_multimodal_2022} proposes the operator $\rmQ_{\text{ID}} = \rmP_1^{t_1} \rmP_2^{t_v}$, which can be applied directly or iterated as a whole. \IDM can also be though of as an \ADM process, on the already iterated transition matrices $\{\rmP_v^{t_v}\}$, for $v = 1, 2$. Essentially, this process first denoises each views before composing them via alternating diffusion.

Multi-View Diffusion (\MVDM) \citep{lindenbaum_multi-view_2020}, instead relies on the computation of a new matrix from the kernels $\rmK_1,\rmK_2$:
\begin{align}
  \rmK_{\text{\MVDM}} &= \begin{bsmallmatrix} 0 & \rmK_1 \rmK_2 \\ \rmK_2 \rmK_1 & 0 \end{bsmallmatrix}, \\
  \rmQ_{\text{\MVDM}} &= \rmD_{\text{\MVDM}}^{-1}\rmK_{\text{\MVDM}},
\end{align}
where $(\rmD_{\text{\MVDM}})_{ii} = \sum_j (\rmK_{\text{\MVDM}})_{ij}$ is the `degree' of $\rmK_{\text{\MVDM}}$. The idea stems from De Sa's approach \citep{de_sa_spectral_2018} and is conceptually similar to \ADM, as the process alternates between the two views. A key difference is that $\rmQ_{\text{\MVDM}}$ admits a real eigenvalue decomposition.

Other methods use the transpose of transition matrices to obtain operators that admit real eigenvalues. While this departs from the standard framework of stochastic transition matrices\footnote{
  \emph{Iterating row-stochastic matrices}:~If $\rmP$ and $\rmQ$ are two row-stochastic matrices, the product $\rmP \rmQ^\top$ is not generally stochastic: $\sum_j (\rmP \rmQ^\top)_{ij} = \sum_{j,k} \rmP_{ik} \rmQ_{jk} = \sum_k \rmP_{ik} (\sum_j \rmQ_{jk})$, which sums to $1$ only if $\sum_j \rmQ_{jk} = 1$ for all $k$. This holds only when $\rmQ$ is doubly stochastic, which is not assumed here.}, these approaches still build on the diffusion intuition.

In particular, Cross-Diffusion (\CrD) \citep{wang_unsupervised_2012} constructs two simultaneous `diffusion' processes:
\begin{align}
  \rmQ_{\text{CrD},1}^{(t+1)} =& \rmP_1 \rmQ_{\text{CrD},2}^{(t)} \rmP_2^\top \\
  \rmQ_{\text{CrD},2}^{(t+1)} =& \rmP_2 \rmQ_{\text{CrD},1}^{(t)} \rmP_2^\top,
\end{align}
where $\rmQ_{\text{CrD}, v}^{(1)} = \rmP_v, v =1, 2 $, which are then fused as:
\begin{equation}
  \rmQ_{\text{CrD}}^{(t)} = \frac{1}{2}\left(\rmQ_{\text{CrD},1}^{(t)} + \rmQ_{\text{CrD},2}^{(t)}\right).
\end{equation}

Similarly, Composite Diffusion (\ComD) \citep{shnitzer_recovering_2018} defines two operators:
\begin{align}
  \rmQ_{\text{ComD},1} &= \rmP_2 \rmP_1^\top + \rmP_1 \rmP_2^\top, \\
  \rmQ_{\text{ComD},2} &= \rmP_2 \rmP_1^\top - \rmP_1 \rmP_2^\top,
\end{align}
where $\rmQ_{\text{ComD},1}$ is symmetric and thus admits a real eigenvalue decomposition, while $\rmQ_{\text{ComD},2}$ is anti-symmetric (\ie $\rmQ_{\text{ComD},2}^\top = - \rmQ_{\text{ComD},2}$) and admits a purely imaginary eigenvalue decomposition. These operators are then used to compute embeddings: $\rmQ_{\text{ComD},1}$ captures similarity between views, whereas $\rmQ_{\text{ComD},2}$ encodes their dissimilarity. For clustering tasks, one can rely on the first operator $\rmQ_{\text{ComD},1}$.
\begin{table*}[h]\footnotesize\centering
	\begin{tabular}{r c c c c c}
		\Xhline{1pt}
		\multirow{2}{*}{\textbf{Method}\qquad\qquad\qquad\qquad} & \textbf{Row-stochastic} & \textbf{Extensible to} & \textbf{Tuning of the} & \textbf{Real} \\
		& \textbf{transition matrices} & \textbf{more than $2$ views} & \textbf{time parameter}&  \textbf{eigenvalues} & \textbf{encompassed by MDTs}\\
		\hline
		\ \ \ \ Multi-View Diffusion (\MVDM) \citep{lindenbaum_multi-view_2020} (2020) & $\checkmark$ & $\sim$       &  --         & $\checkmark$ & --\\
		\ \ \ \ \ \ \ \ \ Integrated Diffusion (\IDM) \citep{kuchroo_multimodal_2022} (2022) & $\checkmark$ & $\checkmark$ & $\checkmark$ & -- & $\checkmark$ \\
		\ \ \ \ \ \ \ \ Alternating Diffusion (\ADM) \citep{katz_alternating_2019} (2019) & $\checkmark$ & $\checkmark$ &  --         & -- & $\checkmark$ \\
		\ \ \ Cross-Diffusion (\CrD) \citep{wang_unsupervised_2012} (2012) & -- &  $\checkmark$   &--      &  --         & -- \\
		Composite Diffusion (\ComD) \citep{shnitzer_recovering_2018} (2018) & -- &  --         &  --         & $\checkmark$ & --\\
		\Xhline{0.5pt}
		Multi-view Diffusion Trajectory (\MDT) (ours)  & $\checkmark$ & $\checkmark$ & $\checkmark$ & -- & --\\
		\Xhline{1pt}
		&\multicolumn{5}{c}{\footnotesize{$\checkmark$: offers the feature; \qquad $\sim$: does not scale well with respect to the number of views.}}
	\end{tabular}
		\caption{\textbf{Operator-based multi-view diffusion methods.} Overview of methods compared to the proposed \MDT (and its variants).}
	\label{tab:existing-methods}
\end{table*}

\section{Multi-view Diffusion Trajectories (\MDTs)}\label{sec:mdt}

This section introduces the proposed framework for constructing a diffusion geometry across multiple data views. Our goal is to generalize the single-view diffusion operator into a family of \emph{intertwined} operators that jointly explore all views through time-inhomogeneous diffusion processes.

\subsection{The proposed view-intertwining diffusion model}\label{subsec:proposed-dm}
The central object of our framework is the following definition; see also the illustration in \cref{fig:MDT}.
\begin{definition}[Multi-view diffusion trajectory -- \MDT] \label{def:diff_traj}
Let $\Pset$ be a set of transition matrices representing the diffusion operators related to different data-views, and $(\rmW_i)_{i\in\mathbb{N}}$ a sequence with $\rmW_i\in\Pset$. The operator of a \emph{multi-view diffusion trajectory} of length $t\in\mathbb{N}^\ast$ is
\begin{equation}
    \rmW^{(t)} = \rmW_t \rmW_{t-1} \cdots \rmW_1, \quad \text{with } \ \rmW^{(1)} = \rmW_1.
\end{equation}
\end{definition}
Each $\rmW_i$ determines the view in which diffusion occurs at step~$i$. Because matrix multiplication is non-commutative, the sequence order defines a specific interaction pattern among views. Intuitively, $\rmW^{(t)}$ describes the cumulative effect of performing $t$ random-walk steps, each possibly performed in a different view. If $V=1$, the process reduces to the standard homogeneous diffusion operator $\rmP^t$. For multiple views, however, the process is \emph{time-inhomogeneous}: the transition law changes with each step, thereby capturing more complex interactions between the views. This flexibility allows the design of different variants of diffusion trajectories by specifying both the composition the set $\Pset$ and the sequence in which the matrix products are applied. Further details on this construction are given in \cref{sec:the_Pset}.

\subsection{Basic properties}

To ensure that the diffusion process is well-defined for any sequence, we first state simple conditions that guarantee ergodicity and convergence. Proofs are given in \appref{app:proofs}.

\begin{property}[Stability of aperiodicity and irreducibility] \label{prop:W_irreducible}
If every $\rmP \in \Pset$ is a row-stochastic matrix with strictly positive diagonal entries,
and the associated Markov chains are aperiodic and irreducible,
then any product $\rmW^{(t)}$ obtained from them also defines an aperiodic and irreducible chain.
\end{property}
Therefore, alternating or mixing well-connected views does not break connectivity: the joint diffusion preserves the ability to reach any node from any other in a finite number of steps.
\begin{corollary}[Existence of a stationary distribution] \label{cor:W_irreducible}
Under the same assumptions:
\begin{enumerate}
  \item Each trajectory $\rmW^{(t)}$ admits a unique stationary distribution $\pi_t$ such that $\pi_t^\top \rmW^{(t)} = \pi_t^\top$.
\item As $t\!\to\!\infty$, $\rmW^{(t)}$ converges to a rank-one matrix, denoted $\rmW^{(\infty)}=\mathbf{1}\pi_\infty^\top$.
\end{enumerate}
\end{corollary}
Note that, even if $\rmW^{(t)}$ is aperiodic and irreducible, it is generally not self-adjoint in $\langle \cdot, \cdot \rangle_{1/\pi_t}$ nor reversible. This corollary says that the inhomogeneous diffusion process converges to a stationary regime, similarly to classical diffusion maps. This implies that the parameter $t$ controls the temporal scale of smoothing: short trajectories emphasize local structure, whereas longer ones capture a form of global organization, common to multiple views. Due to the time-inhomogeneous nature of the process, however, the interpretation of $t$ is more subtle than in the single-view case. In particular, while increasing $t$ generally leads to more global smoothing, different trajectories might yields different rates of convergences, depending on the sequence of views selected. Moreover, the decay of eigenvalues of $\rmW^{(t)}$ might be non-monotonic with $t$.

\subsection{Trajectory-dependent diffusion geometry}

Given the \MDT operator $\rmW^{(t)}$, we can define a trajectory-dependent diffusion distance that measures the connectivity between datapoints, as determined by the multi-view trajectory at hand.
\begin{definition}[Trajectory-dependent diffusion distance]\label{def:traj_diff_dist}
Let $\pi_t$ be the stationary distribution of the \MDT operator $\rmW^{(t)}$. The trajectory-dependent diffusion distance between datapoints $x_i$ and $x_j$ is:
\begin{equation}
    \gD_{\rmW^{(t)}}(x_i,x_j)
    = \sum_{k=1}^{N}\frac{1}{\pi_t(k)}\big(\rmW^{(t)}_{ik}-\rmW^{(t)}_{jk}\big)^2.
    \label{eq:trajectory-distance}
\end{equation}
\end{definition}
\cref{def:traj_diff_dist} generalizes the standard diffusion distance \citep{coifman_diffusion_2006}: two datapoints are close if their transition probability profiles are similar \emph{across all views explored by the trajectory}, in this sense, this distance is also less sensitive noise inside views, as it aggregates information across their probability profiles. $\rmW^{(t)}$ is not necessarily symmetric or diagonalizable, hence we rely on its singular value decomposition (SVD) to construct an embedding.
\begin{definition}[Trajectory-dependent diffusion map]
Let $\rmW^{(t)}$ be a \MDT operator with SVD $\rmW^{(t)} = \rmU_t\rmSigma_t \rmV_t^\top$. The trajectory-dependent diffusion map is defined as:
\begin{equation}
    \rmPsi^{(t)}(x_i)=e_i^\top \rmU_t\rmSigma_t,
\end{equation}
where $e_i$ is the $i$-th canonical vector.
\end{definition}
\begin{property}[Diffusion distance preservation] \label{prop:diff_dist_map}
The trajectory-dependent diffusion map $\Psi^{(t)}$ satisfies:
  \begin{equation}
    \gD_{\rmW^{(t)}}(x_i,x_j) = \norm{\rmPsi^{(t)}(x_i) - \rmPsi^{(t)}(x_j)}_2.
  \end{equation}
\end{property}
This embedding preserves the trajectory-dependent diffusion distance, thus generalizing diffusion maps to the proposed time-inhomogeneous process. A key difference from \citep{coifman_diffusion_2006} is that they consider the eigenvalue decomposition of $\rmP$, relying on the direct relation between the eigenvalues of $\rmP$ and $\rmP^t$. In contrast, our \MDT operator $\rmW^{(t)}$ is heterogeneous, so the singular values of $\rmW^{(t)}$ are generally not directly related to those of $\rmW^{(t')}$ for $t' \neq t$. \cref{prop:diff_dist_map} also shows that $\gD_{\rmW^{(t)}}$ defines a metric on the data points for any $t \in \sN^*$. Moreover, this property enables robust handling of noise via SVD truncation, allowing a balance between preserving the accuracy of the trajectory-dependent diffusion distance and reducing the impact of noise in the dataset.

\begin{algorithm}[t]
  \small
  \caption{\textbf{--} \MDT-based diffusion map}\label{alg:MDT}
  \begin{algorithmic}[1]
    \STATE {\bfseries Input:}  \ \ \ $\rmX_1, \rmX_2, ..., \rmX_\Vnum$: a collection of data views \\
    \STATE \hspace*{\algorithmicindent}\hspace*{\algorithmicindent}\quad \ \ \ $q$: a quality measure \\
    \STATE {\bfseries Output:} $\rmW^{(t)}$: a diffusion operator \\
    \vspace{1.3mm}
    \hrule
    \vspace{1.3mm}
    \STATE Consider a diffusion operator $\rmP_i$ for each view $i$
    \STATE Construct the set of base diffusion operators $\Pset$ \algComment{see \cref{sec:the_Pset}}
    \STATE Determine a sequence $\rmW^{(t)}$ based on the quality measure $q$: \\
		\quad\  $\rmW^{(t)} \in \argmax_{\rmU \in \Pset^t} q(\rmU)$ \hfill \algComment{see \cref{sec:determine-sequence}}\\
    \STATE \textbf{return} $\rmW^{(t)}$
  \end{algorithmic}
\end{algorithm}

\subsection{Designs for \texorpdfstring{$\Pset$}{P} and associated trajectory spaces} \label{sec:the_Pset}
The flexibility of the \MDT framework rely on both the definition of $\Pset$ and the selection of the sequence $(\rmW_i)$. In particular, assumptions that are made in the definition of the $\Pset$ are weak enough so that many variations of designs can be explored. We discuss here baseline designs for $\Pset$, relationship of \MDTs with existing multi-view frameworks are further discussed in \cref{sec:mdt-vs-existing}.

\inlinetitle{Discrete designs}{.}~A straightforward approach is to define $\Pset$ as a countable set of operators, using either all or a subset of the views.
\begin{definition}[Discrete Multi-view Diffusion Trajectory]
  We say that $\rmW^{(t)}$ is a \emph{discrete multi-view diffusion trajectory} if for all $t$, $\rmW_t \in \Pset$ and $\Pset$ is countable.
\end{definition}
A natural choice is to take the \emph{canonical diffusion operator set} $\Psetc = \{\rmP_v\}_{v=1}^V$, defined in \cref{sec:background}, where each $\rmP_v$ is hte transition matrix for view~$v$. Trajectories may then follow deterministic or random sequences of these operators. When $V = 1$, we recover the standard single-view diffusion maps. Additional discrete operators can be included in $\Pset$ to enrich the trajectory space: the identity matrix $\rmI_N$ allows a `idle' step, the rank-one uniform operator $\mXi = \tfrac{1}{N}\mathbf{1}\mathbf{1}^\top$ can introduce teleportation effects when combined with canonical operators as in a PageRank-style diffusion $\rmP_{\mathrm{PR},v}(a)=a \rmP_v+(1-a)\mXi$. While directly taking $\rmW_i = \mXi$ collapses the product to rank one, it motivates including teleportation or perturbed operators in more general designs \citep{gleich_pagerank_2014}. Similarly, local-scale smoothing operators such as $\mXi_v = \rmP_v^{t'}$ for small $t' \in \sN^*$ can be added, either directly in $\Pset$ or as replacements for $\mXi$ in PageRank-like operators, to balance local and global diffusion effects \citep{Langville2004Deeper, gleich_pagerank_2014}.

\begin{figure*}[t!]
  \centering
  \begin{subfigure}[t]{0.5\textwidth}
    \centering
		\includegraphics[width=\textwidth, viewport=0 50 930 860, clip]{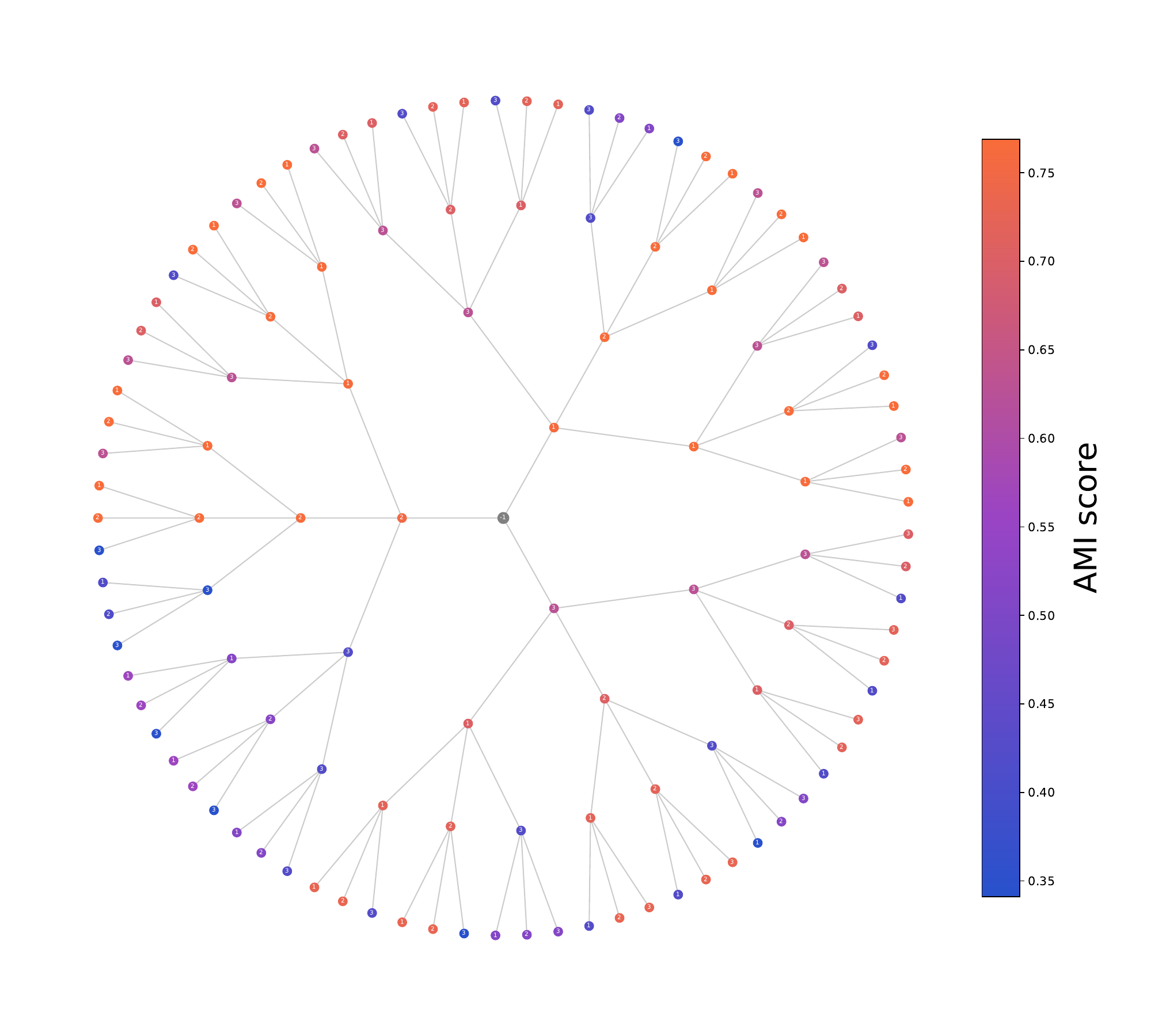}%
    \label{subfig:065MVTree}
  \end{subfigure}%
  \caption{\textbf{Tree structure of the discrete trajectory space.} Representation of the discrete \MDT trajectory space as a tree (see \cref{sec:the_Pset}) for the L-Isolet dataset that has $3$ views (see description in \cref{tab:datasets}). The space is illustrated as a tree in a circular layout; a trajectory starts from the root, which is the identity matrix, and then navigates in the space by choosing the next operator $\rmW_i\in\Pset$ that is added to the product sequence $\rmW^{(t)}$ (\cref{def:diff_traj}). The depth of each node indicates the length $t$ of the associated trajectory. Short trajectories emphasize local structure, whereas longer ones capture the global structure, common to multiple views. The tree shows also task-related information: node color indicates the clustering quality (AMI) achieved by spectral clustering using the associated \MDT operator.}
  \label{fig:tree-reg}
\end{figure*}

In the discrete setting, trajectories can be visualized as paths in a $\abs{\Pset}$-ary tree of depth $t$, where each node at depth $i$ represents the choice of operator $\rmW_i \in \Pset$. Consequently, the total number of distinct trajectories grows exponentially with both the number of views $V$ and the diffusion time $t$, yielding $V^t$ possible paths. \cref{fig:tree-reg} illustrates this tree structure for a 3-view dataset, with node colors indicating the clustering quality achieved by spectral clustering using the corresponding \MDT operators.

\inlinetitle{Continuous designs}{.}~%
A more flexible approach defines $\Pset$ as a continuous set of operators.
\begin{definition}[Continuous Multi-view Diffusion Trajectory]\label{def:continuous_MDT}
We say that $(\rmW_i)_{i}$ is a \emph{continuous multi-view diffusion trajectory} if $\Pset$ is continuous. A simple choice is the convex hull of the canonical set (\ie the convex set including its elements):
\begin{equation}\label{eq:def_Psetcvx}
    \Psetcvx = \Big\{\sum_{v=1}^V a_v \rmP_v \;\big|\; a_v \geq 0,\ \sum_v a_v = 1\Big\}.
\end{equation}
The associated \MDT operator is then denoted as $\rmW^{(a,t)}$.
\end{definition}
Each convex combination can vary at every step, \ie $a_{i,v} \neq a_{j,v}$ for $i \neq j$. Thus, a trajectory is represented by the sequence of weight vectors $(a_i)_{i=1}^t$, with $V-1$ degrees of freedom per step under the simplex constraint. This defines a smooth, continuous search space, enabling interpolation between views and facilitating optimization of trajectories (see \cref{sec:determine-sequence:optimization}). As in the discrete case, these convex combinations can include perturbed operators, such as locally smoothed versions of $\rmP_v$, teleportation variants or the rank-one uniform operator, allowing the trajectory to balance local and global diffusion effects.

In fact, this continuous formulation naturally generalizes the classical PageRank random walk: for a single view with $\Psetinit = \{\rmP, \mXi\}$ and using the convex set based of $\Psetinit$, an \MDT sequence can be written as
\begin{equation*}
\rmW^{(t)} = \rmW_t \rmW_{t-1} \dots \rmW_1 = \rmP_{\mathrm{PR}}(a_t) \rmP_{\mathrm{PR}}(a_{t-1}) \dots \rmP_{\mathrm{PR}}(a_1),
\end{equation*}
where for all $i$, $a_i \in [0,1]$ and $\rmP_{\mathrm{PR}}(a_i) = a_i \rmP + (1-a_i) \mXi$.
To further highlight \MDT's representation capacity, we could consider the aforementioned PageRank extensions that use local-scale smoothing operators for each view. In the multi-view case, there will be a set of such smoothing operators to be included, $\{\mXi_i\}_{i=1}^{\Vnuminit} \subset \Psetinit$, whose convex combination would allow $\rmW_t$ not only to fuse the transition matrices of the views, but also their associated smoothing operators.

\subsection{Relation to existing multi-view diffusion models}\label{sec:mdt-vs-existing}

Several existing diffusion map frameworks are recovered as special cases of the proposed \MDT framework:
\begin{itemize}[leftmargin=1.5em, itemsep=0em]
    \item \textit{Alternating Diffusion} (\ADM) \citep{katz_alternating_2019}: choosing the canonical set and alternating by $\rmW_i = \rmP_{(i\!\bmod\!V)+1}$, $\rmW^{(Vt)}$ yields the alternating diffusion operator at time $t$.
    \item \textit{Integrated Diffusion} (\IDM) \citep{kuchroo_multimodal_2022}: defining a trajectory where the process first diffuses within each view for $t_v$ steps before switching to another, \eg $\rmW^{(t_1 + t_2)} = \rmP_1^{t_1} \rmP_2^{t_2}$. Another interpretation is that \IDM is an \ADM process on the set of already iterated operators $\{\rmP_v^{t_v} | v=1 \dots V\}$.
    \item \textit{PageRank-inspired \MDTs}: This is one of our designs, which still builds on the well-known PageRank operator to define perturbed multi-view diffusion trajectories. Using convex combinations of the form $\rmW_i = \sum_v a_{i,v} \rmP_{\mathrm{PR},v}(\alpha)$, this interpolates between view-specific and global diffusion.
\end{itemize}
Therefore, the \MDT framework unifies in a comprehensible way a broad family of operator-based multi-view diffusion methods under a single probabilistic and geometric formulation. In that sense, approaches that construct a composite larger operator than the $N \times N$ operators of the views, such as that of \citep{lindenbaum_multi-view_2020} (see \cref{sec:background}), do not fall within our framework. The same way, approaches that do not rely on random-walk transition matrices, or uses backward operators such as \citep{wang_unsupervised_2012,shnitzer_recovering_2018}, are also outside what \MDT can encompass.

\section{Unsupervised diffusion trajectory learning}
\label{sec:determine-sequence}

\subsection{Problem formulation} \label{sec:determine-sequence:problem-formulation}

One of the most important advantages brought by the \MDT framework over the standard approaches, is that it enables \emph{operator learning}. The space of Multi-view Diffusion Trajectories (\MDTs), introduced in \cref{sec:mdt}, defines a broad family of diffusion operators obtained by combining the view-specific transition matrices contained in $\Pset$. However, not all trajectories are equally informative: different sequence of view operators (multi-view diffusion trajectories) emphasize distinct aspects of the underlying manifold and propagate different types of inter-view relations. Importantly, the quality of the resulting embedding depends not only on which operators are selected ($\Pset$), but also on how they are ordered ($(\rmW_i)$) and on the trajectory length (the number of diffusion steps $t$). This motivates the definition of an \emph{internal quality measure} to guide the trajectory selection. Therefore, the search for an `optimal' trajectory is constrained by three elements:
i)~the expressiveness of $\Pset$,
ii)~the diffusion time $t$, and
iii)~the chosen task-dependent quality measure. This section formulates the selection, or learning, of diffusion trajectories as an \emph{optimization problem} guided by a suitable internal measure, which seeks for effective optimization strategies to deal with the constraints introduced by the \MDT framework.

Given a dataset represented by the operator set $\Pset$, the objective is to determine a sequence of matrices $\tau = (\rmW_i)_{i=1}^t$ who's product $\rmW^{(t)}$ produces a meaningful (for the problem at hand) and well-structured embedding of the data. This search is guided by an \emph{internal quality measure}~$Q(\tau)$, defined as:
\begin{equation}
   \tau^\star = \argmax_{\tau\in\Pset^t\!,\,t \in \sN^\ast} Q(\tau),
\end{equation}
where $\Pset^t$ is the set of all possible sequences of length $t$ formed by the operators in $\Pset$. Although $t$ is a parameter in the problem formulation, in practice it is often fixed using heuristics \citep{kuchroo_multimodal_2022}. We will showcase later on how to adapt such heuristics to the \MDT framework.
Typical internal measures include clustering validity indices such as the Davies-Bouldin, Silhouette, and Calinski-Harabasz (CH) indices \citep{vinh_information_2009, desgraupes_clustering_2016}, which have all been extensively studied (\eg see \cite{gagolewski_are_2021}). Without relying on external labels, such indices can evaluate the compactness and separation of clusters formed in the diffusion space \citep{sevi_generalized_2025, sevi2022clustering}, which in the present case would be induced by a trajectory, and therefore can serve as guiding internal quality measures. Each index introduces its own bias, reflecting specific assumptions about what constitutes a `good' clustering, and should therefore be selected and interpreted with care. Although originally defined for single-view settings, a simple way to be extended to the multi-view case is by computing a weighted sum of the per-view scores, where the weights account for both the relative importance of each view and the scale of the corresponding criterion. In \cref{sec:experiments:clustering}, the CH index is adopted as the internal quality measure $Q(\tau)$ evaluated on the diffusion operators $\{\rmP_{i}\}_{i=1}^V$. CH is a computationally efficient index that has been shown to perform well in various clustering scenarios \citep{desgraupes_clustering_2016}. For a dataset $\rmX$ and a clustering $C$, the CH index is defined as:
\begin{align*}\label{eq:CH_index}
    \text{CH}(\rmX, C) = \frac{\tr{\rmB_C}}{\tr{\rmW_C}} \cdot \frac{N - |C|}{|C| - 1}, \\
    Q_{\text{CH}}(\tau) = \sum_{i=1}^\Vnum w_i \cdot \text{CH}\left(\rmX_v, C_i(\tau) \right),
\end{align*}
where $C_i(\tau)$ is the clustering obtained by applying $k$-means on the trajectory-dependant diffusion map obtained by $\tau$, $\rmB_C$ and $\rmW_C$ are the between- and within-cluster scatter matrices, respectively, and $w_i$ is a weight associated to view $i$ that is set to $w_i = 1/V$ in the absence of prior knowledge. Other single-view clustering validity indices can be used in place of CH within the definition of $Q_{\cdot}(\tau)$.
In \cref{sec:experiments:manifold}, we alternatively propose and employ a contrastive-inspired internal criterion to assess the quality of the learned embeddings.

\subsection{Optimization strategies for \MDT learning}\label{sec:determine-sequence:optimization}

The optimization strategies discussed in this section aim to select or learn effective diffusion trajectories within the \MDT framework. To do so, they rely on either the continuous or the discrete \MDT formulation, defined in \cref{sec:the_Pset}. This section's strategies are summarized in \cref{tab:MDT-variants}.

\inlinetitle{Random sampling}{.}~%
We first consider selecting diffusion trajectories through random sampling as a simple and computationally inexpensive baseline. This approach provides a reference distribution over trajectories that captures the average random behavior of multi-view diffusion without explicit optimization, and without relying on an internal quality measure, serving as a useful baseline to assess the impact of more sophisticated trajectory designs.
\begin{definition}[Random multi-view trajectory] \label{def:rand_traj}
A sequence of operators $(\rmW_i)_{i \in \sN}$ is called a \emph{random multi-view trajectory} if each operator $\rmW_i$ is independently drawn from a probability distribution $\mu_i$ supported on the operator set $\Pset$:%
\begin{align*}
  &\forall i,\ \rmW_i \sim \mu_i, \quad \supp(\mu_i) \subseteq \Pset \ \,\text{and } \ \,\forall i \neq j,\ \rmW_i \indep \rmW_j.
\end{align*}
\end{definition}
This stochastic process can be viewed as a discrete-time \emph{Markov Jump Process}~(MJP)~\citep{costa_discrete_2005}, where the states correspond to the operators in $\Pset$ and the transition law evolves according to the sequence of distributions $(\mu_i)_i$. When for all $i$, $\mu_i = \mu$, then $\rmW_i \stackrel{\textup{\tiny i.i.d.}}{\sim} \mu$, and the resulting expected operator satisfies $\mathbb{E}[\rmW^{(t)}] = \big(\mathbb{E}_{\rmW \sim \mu}[\rmW]\big)^t$, which represents the average diffusion geometry induced by random sampling. Although individual trajectories may yield noisy embeddings, the ensemble of such random paths approximates a mixture model over all possible diffusion trajectories. This property makes random sampling valuable as an unbiased baseline and as a stochastic regularization mechanism; this aspect is further developed in \cref{sec:experiments}. In practice, this approach enables fast generation of multiple trajectories at negligible cost, providing an empirical evaluating reference.

\inlinetitle{Criteria-based selection}{.}
Using the problem formulation in \cref{sec:determine-sequence:problem-formulation}, we can establish criteria for selecting promising trajectories based on their expected performance according to the chosen task-dependent quality measure $Q$. Depending on the trajectory space, different optimization strategies can be employed. In the \emph{discrete \MDT setting}, tree search algorithms can efficiently explore the combinatorial space of operator sequences. Techniques such as beam search, Monte Carlo Tree Search or $\epsilon$-greedy algorithms can be used to select \MDTs that achieve higher quality scores.
In the \emph{continuous \MDT setting}, gradient-based optimization methods or evolutionary algorithms can be utilized to navigate the continuous parameter space defined by $\Pset$. In the particular case of a convex \MDT, where the operator is expressed as a convex combination of the form $\rmW_i = \sum_{v=1}^\Vnuminit a_v \rmP_v$, these optimization methods refine iteratively the weight vectors $(a_i)$ to maximize the quality index $Q$, potentially uncovering nuanced combinations of views that better capture the underlying data structure. The choice of optimization strategy should consider both computational efficiency and the ability to escape local optima, ensuring a robust search for high-quality diffusion trajectories. In particular, for this purpose we consider Direct \citep{jones_direct_2001}, which has the advantage of being non-stochastic and hence has needs lower computational cost compared to stochastic methods, but works well when $t \times \abs{\Psetinit}$ is relatively small. Other methods can also be considered for the optimization, \eg \cite{hu_survey_2012, serre_stein_2025, martinez_bayesopt_2014, igel_evolutionary_2007}.

\inlinetitle{Selecting the diffusion time horizon ($t$)}{.}~%
The number of diffusion steps $t$ is a critical hyperparameter that controls the scale at which the data structure is analyzed: short diffusions emphasize local geometry, while longer ones over-smooth the manifold and eliminate fine details.
For determining $t$, we adapt an existing approach that tracks how the singular value spectrum of $\rmW^{(t)}$ evolves over time \cite{kuchroo_multimodal_2022}. To this end, we define the \emph{singular entropy} as the Shannon entropy of the normalized singular value distribution:
\begin{equation}
  S(\rmW^{(t)}) \defeq
  -\sum_i \sigma_i(\rmW^{(t)}) \log\big(\sigma_i(\rmW^{(t)})\big),
\end{equation}
where the singular values are normalized to $\sigma_i(\rmW^{(t)}) \in [0,1]$.
Low entropy indicates that most of the diffusion energy is concentrated in a few leading components (\ie the matrix is close to rank-one), suggesting that the dominant structure has emerged. High entropy reflects a more uniform spectrum, potentially corresponding to noise or unresolved substructures. The optimal diffusion time is chosen at the \emph{elbow point} of the entropy curve, balancing noise reduction and information retention.

Although $\rmW^{(t)}$ converges to a rank-one matrix asymptotically (\cref{cor:W_irreducible}), its singular values do not necessarily decrease monotonically due to the inhomogeneity of the process. To mitigate fluctuations, we compute the entropy of the expected operator $\mathbb{E}_{\mu}[\rmW^{(t)}]$ that yields smoother and more stable behavior in practice. When no prior information about the relative importance of views is available, the expectation is taken under an i.i.d.\ assumption with uniform measure $\mu=\frac{1}{|\Pset|}$ over $\Pset$. In this context:
\begin{align*}
	\E_\mu[\rmW^{(t)}] = (\E_\mu[\rmW_1])^t = \Bigg(\frac{1}{|\Pset|} \sum_{\rmP \in \Pset} \rmP\Bigg)^t.
\end{align*}
\cref{fig:singular_entropy} shows example entropy curves for the singular values.

\begin{table*}[t]\footnotesize\centering
	\begin{tabular}{l l c c c}
		\Xhline{1pt}
		\multirow{2}{*}{\textbf{\MDT variant}} & \multicolumn{1}{c}{\textbf{Trajectory}} & \multicolumn{1}{c}{\textbf{Internal}} & \multicolumn{1}{c}{\textbf{Optimization}} & \multicolumn{1}{c}{\textbf{Application}}\\
		& \multicolumn{1}{c}{\textbf{space}} & \multicolumn{1}{c}{\textbf{quality index}} & \multicolumn{1}{c}{\textbf{method}} & \multicolumn{1}{c}{\textbf{task}}\\
		\Xhline{0.5pt}
		\MDTCST & $\Psetcvx^t$ & Contrastive (\cref{eq:contrastive_loss}) & ADAM & Manifold learning \\
		\Xhline{0.25pt}
		\MDTrand & $\Psetc^t$     & -- & -- & Clustering\\
		\MDTcvx & $\Psetcvx^t$    & -- & -- & Clustering\\
		\MDTchs  & $\Pset^t$      & CH & Direct & Clustering\\
		\MDTbsc & $\Pset^t$ & CH & Beam-Search & Clustering\\
		\Xhline{1pt}
	\end{tabular}
	\caption{\textbf{\MDT variants.} Overview of the \MDT variants used in the experiments related to two tasks.}
	\label{tab:MDT-variants}
\end{table*}

\section{Experiments}\label{sec:experiments}

\subsection{General setup}\label{sec:experiments:general-setup}

This section showcases the proposed \MDT framework in the manifold learning and the data clustering tasks. We define the MDT variants presented in \cref{tab:MDT-variants}, which derive from the proposed \MDT framework (\cref{sec:determine-sequence}).
In the manifold learning context, we employ the convex \MDTCST that is obtained by optimizing a contrastive-like loss (\cref{sec:experiments:manifold}). In the clustering case, we choose the Calinski-Harabasz index (CH) as an internal clustering-based quality criterion to guide the \MDT construction. Four \MDT variants are considered for clustering: \MDTrand, \MDTcvx, \MDTchs, and \MDTbsc (\cref{sec:experiments:clustering}). \MDTrand and \MDTcvx represent randomly sampled trajectories in discrete and convex spaces, respectively. \MDTchs employs Direct optimization in the convex set $\Psetcvx$, while \MDTbsc uses beam search to explore the discrete trajectory space $\Psetc$.
We then compare against state-of-the-art multi-view diffusion methods, (\cref{sec:existing-methods}, \cref{tab:existing-methods}). The performances of single-view diffusions are reported in \appref{fig:viewperf}. Since identifying the best view in an unsupervised manner is generally not feasible, these single-view results are included only as references to contextualize the performance of multi-view methods.

For methods requiring a time parameter tuning (\IDM, \MDTrand, \MDTchs, \MDTcvx, and \pADM), we use the elbow of the singular/spectral entropy using the \texttt{Kneed} package \citep{satopaa_finding_2011}, while for \MDTbsc, the maximum width of the tree is set to $2$ times this elbow. \CrD does not come with a tuning mechanism for the time parameter, hence we follow the suggestion of its authors to set a high value of $t=20$. In the other cases, the time parameter is set to $t=1$. Multi-view diffusion \MVDM produces one embedding per view, since no specific view is preferred, we report the embedding associated to the first view.

In the following subsections, we present results obtained on toy examples and real-life datasets whose characteristics are summarized in \cref{tab:datasets}. We consider point-cloud data, and the transition matrices are built using Gaussian kernels with bandwidth $\sigma = \max_j \left( \min_{i, i \neq j} \dist_v(x_i, x_j) \right)$, with $\dist_v$ denoting the Euclidean distance in view $v$. A $K$-nearest neighbors ($K$-NN) graph is then constructed on the kernelized data ($K = \lceil \log(N) \rceil$). In the manifold learning setting, the embedding dimension is fixed to $2$ to facilitate visualization. In clustering experiments, the dimensionality of the embedding space is set equal to the number of target clusters ($k$), which is assumed to be a known hyperparameter. The implementation of the \MDT variants and all the compared methods has been included in an open-source repository\footnote{The repo is available at: \\ \href{https://github.com/Gwendal-Debaussart/mixed-diffusion-trajectory}{https://github.com/Gwendal-Debaussart/mixed-diffusion-trajectory}}.
  \begin{table*}[t]\small
    \centering
    \begin{tabular}{l@{\hspace{2pt}}lcccl}
    \Xhline{1pt}
    \multicolumn{2}{l}{\textbf{Datasets}} & \multicolumn{1}{c}{\textbf{Number of views} ($v$)} & \textbf{Structure} & \multicolumn{1}{c}{\textbf{Number of datapoints}} ($N$) & \multicolumn{1}{c}{\textbf{Learning task}}\\
    \Xhline{0.5pt}
     Helix A  & \citep{lindenbaum_multi-view_2020} & 2 & Circular & 1500 & Manifold learning\\
     Helix-B   & \citep{lindenbaum_multi-view_2020} & 2 & Circular & 1500 & Manifold learning\\
		 Deformed plane & [ours] & 2 & Plane & 3000 & Manifold learning \\
    \Xhline{0.5pt}
    K-MvMNIST  & \citep{kuchroo_multimodal_2022}    & 2 & 10 clusters & 6000 & Data clustering\\
    L-MvMNIST & \citep{lindenbaum_multi-view_2020}      & 2 & 10 clusters & 6000 & Data clustering\\
    Olivetti & [\href{https://cs.nyu.edu/~roweis/data/olivettifaces.mat}{\textsc{at\&t}}] & 2 & 40 clusters & 400 & Data clustering\\
    Yale & [\href{https://github.com/ChuanbinZhang/Multi-view-datasets/}{\textsc{Web}}] & 3 & 15 clusters & 165 & Data clustering\\
    100Leaves & [\href{https://github.com/ChuanbinZhang/Multi-view-datasets/}{\textsc{Web}}] & 3 & 100 clusters & 1600 & Data clustering\\
    L-Isolet & \citep{lindenbaum_multi-view_2020}     & 3 & 26 clusters & 1600 & Data clustering\\
    MSRC-v5 & [\href{https://github.com/ChuanbinZhang/Multi-view-datasets/}{\textsc{Web}}] & 5 & 7 clusters & 210 & Data clustering\\
		Multi-Feat & [\href{http://archive.ics.uci.edu/dataset/72/multiple+features}{\textsc{uci}}] & 6 & 10 clusters & 2000 & Data clustering\\
    Caltech101-7 & [\href{https://github.com/ChuanbinZhang/Multi-view-datasets/}{\textsc{Web}}] & 6 & 7 clusters & 1474 & Data clustering\\
    \Xhline{1pt}
    \end{tabular}
    \caption{\textbf{Datasets.} Overview of the datasets used in the experiments concerning two learning tasks.}
    \label{tab:datasets}
  \end{table*}
\subsection{Application to manifold learning}\label{sec:experiments:manifold}
\inlinetitle{Internal quality index.}~%
To learn \MDTs for this task, we propose the following quality index to guide the process, inspired from recent works on contrastive learning \citep{pan_multi-view_2021,hu_comprehensive_2024}:
\begin{align}
  & Q_X^v (a, t) = \sum_{i=1}^{N} \sum_{j \in \gN_i^{(m)}} - {\textstyle \log \left( \frac{\exp\left(\rmW^{(a,t)}_{i j}\right)}{ \sum_{k \neq i} \exp \left(\rmW^{(a,t)}_{i k}\right)}\right)}, \\
  & Q_X(a, t) = \sum_{v=1}^\Vnuminit \lambda_v Q_X^v (a,t), \label{eq:contrastive_loss}
\end{align}
where $\mathcal{N}_i^{(v)}$ denotes the neighbors of node $i$ in view $v$ (determined by the kernel matrix, generally after a $K$-NN step), and $\lambda_v$ is a weighting parameter representing the relative view importance. $\rmW^{(a,t)}$ is an \MDT operator associated to the convex set $\Psetcvx$ (see \cref{def:continuous_MDT}). For convenience, we set $\lambda_v = 1/v$ for all $v$, and choose $t$ using the heuristic defined in \cref{sec:determine-sequence:optimization}. To optimize with regards to the proposed loss, we employ the ADAM optimizer. This configuration is named as \MDTCST.

\begin{figure*}
  \centering
  \begin{subfigure}[t]{0.30\textwidth}
    \centering
    \includegraphics[width=\textwidth]{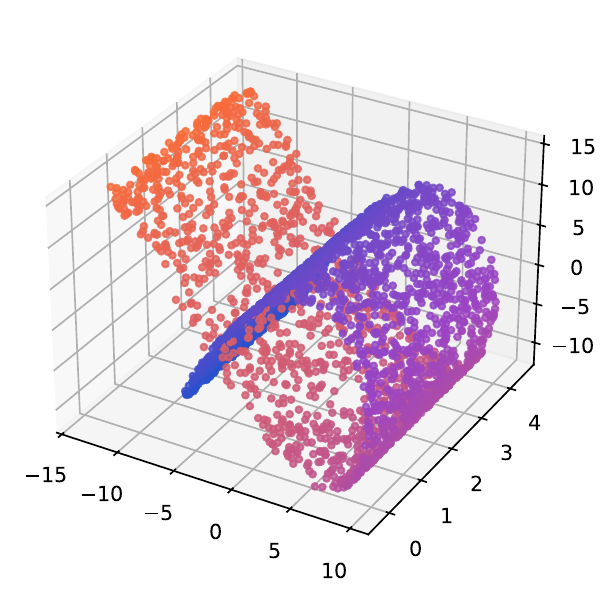}
    \caption{View $1$}
  \end{subfigure}\quad%
  \begin{subfigure}[t]{0.30\textwidth}
    \centering
    \includegraphics[width=\textwidth]{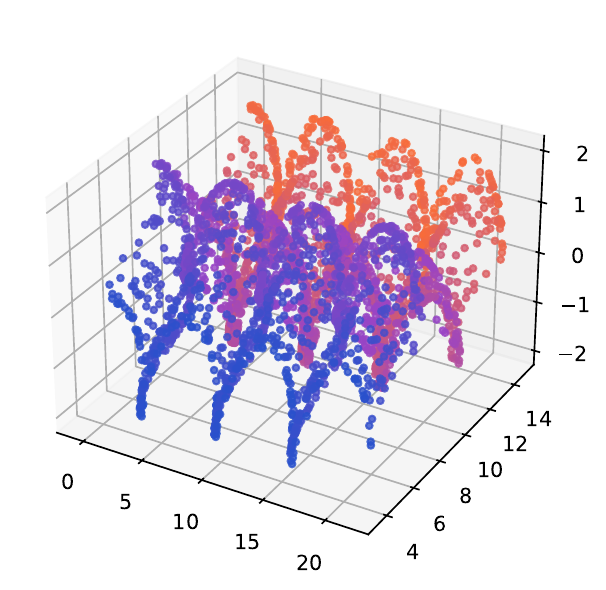}
    \caption{View $2$}
  \end{subfigure}
  \\
  \begin{subfigure}[b]{0.30\textwidth}
    \centering
    \includegraphics[width=\textwidth]{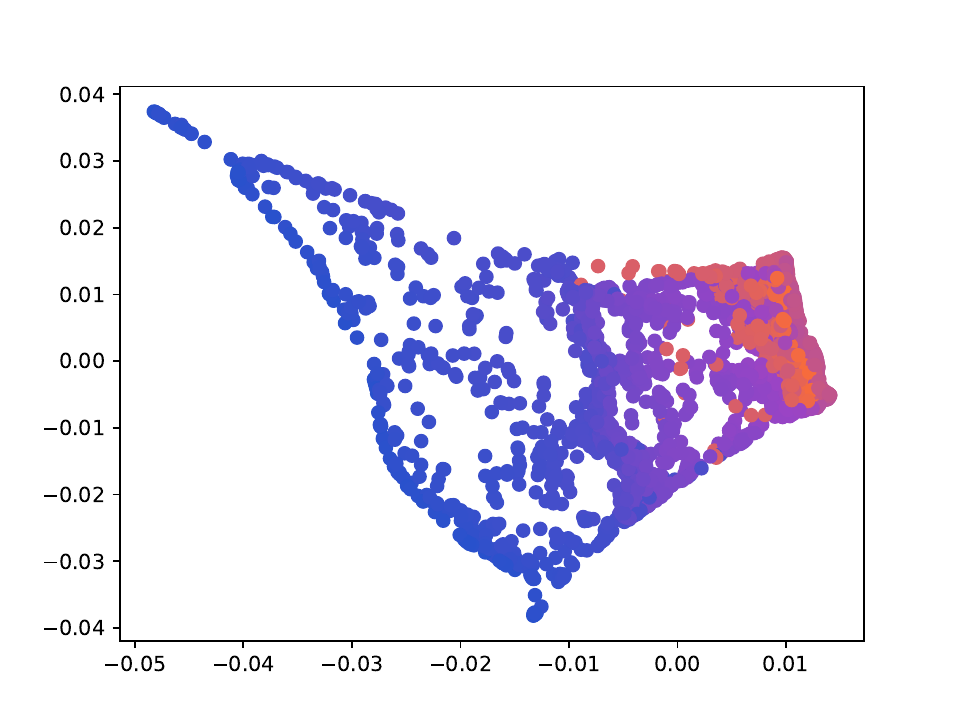}
    \caption{Multi-view Diffusion (\MVDM) -- View 1}
  \end{subfigure}\quad%
   \begin{subfigure}[b]{0.30\textwidth}
    \centering
    \includegraphics[width=\textwidth]{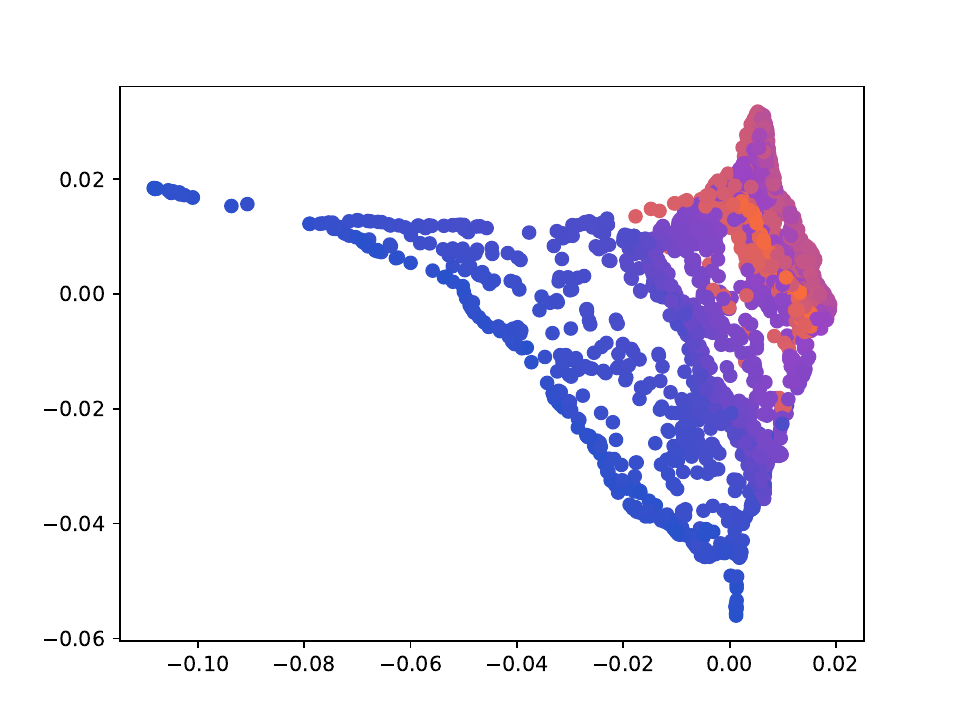}
    \caption{\ADM}
  \end{subfigure}\quad%
    \begin{subfigure}[b]{0.30\textwidth}
    \centering
    \includegraphics[width=\textwidth]{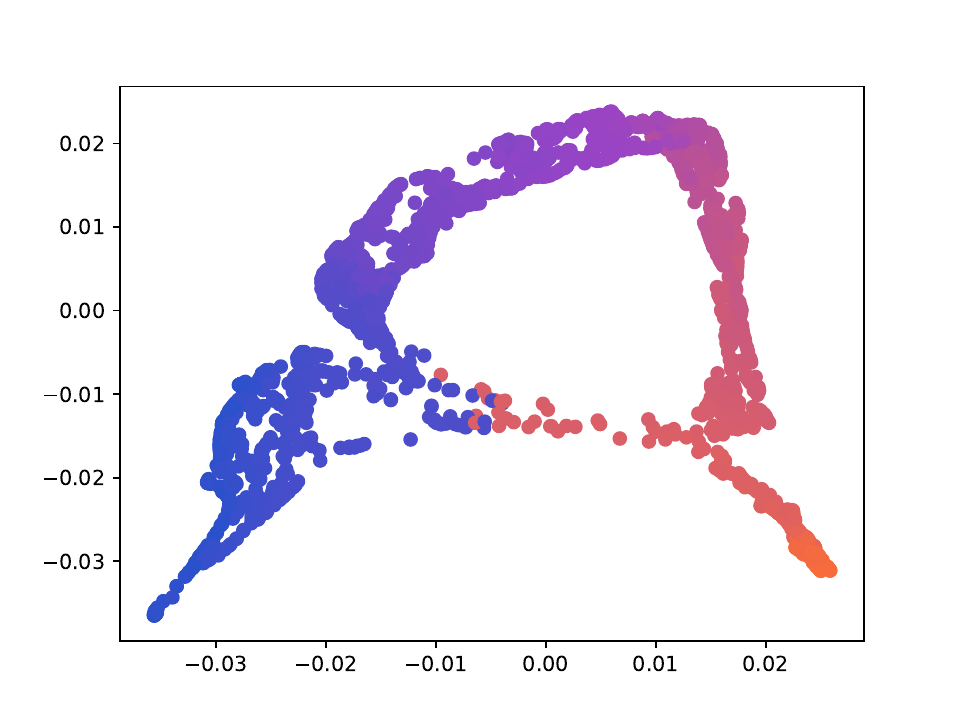}
    \caption{\IDM}
  \end{subfigure}
  \\
  \begin{subfigure}[b]{0.30\textwidth}
    \centering
    \begin{minipage}{\textwidth}
      \centering
      \begin{subfigure}[t]{0.48\textwidth}
        \includegraphics[width=\textwidth]{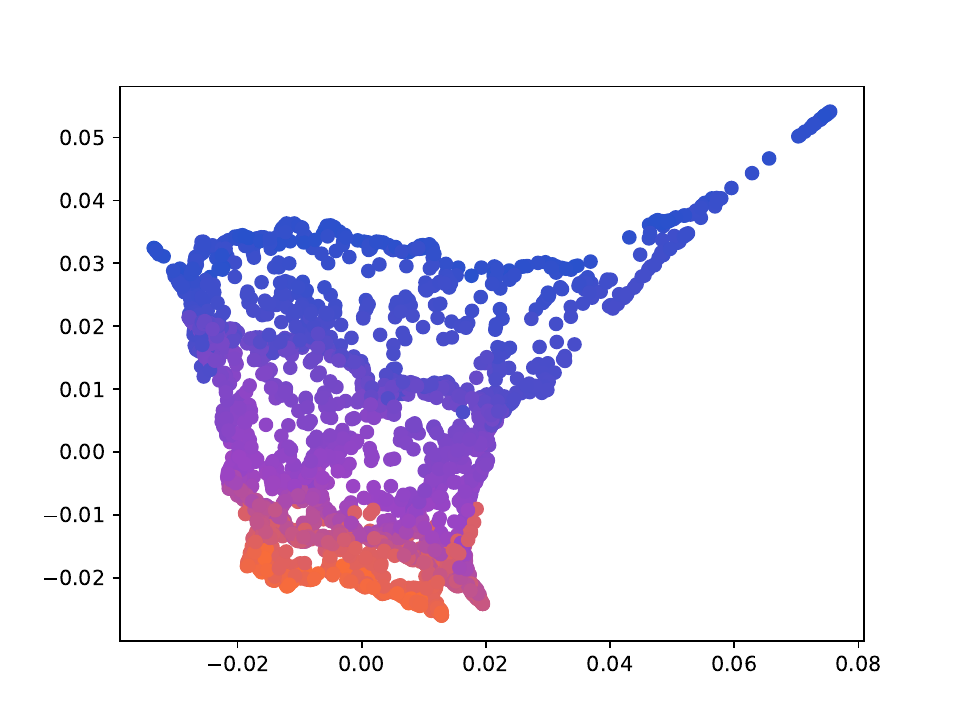}
      \end{subfigure}%
      \begin{subfigure}[t]{0.48\textwidth}
        \includegraphics[width=\textwidth]{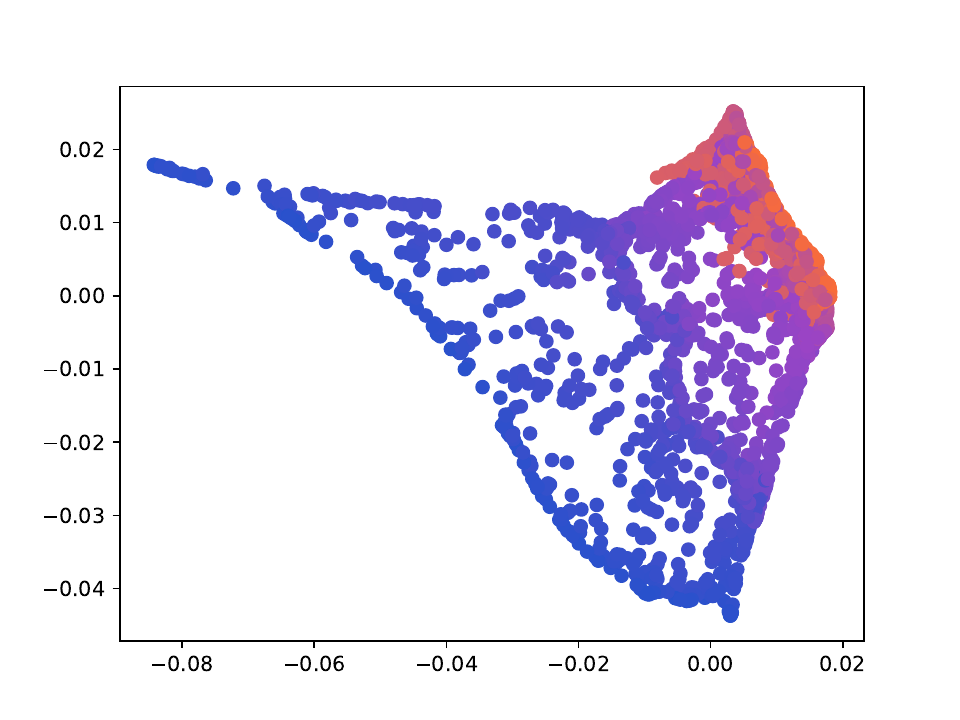}
      \end{subfigure} \\
      \begin{subfigure}[t]{0.48\textwidth}
        \includegraphics[width=\textwidth]{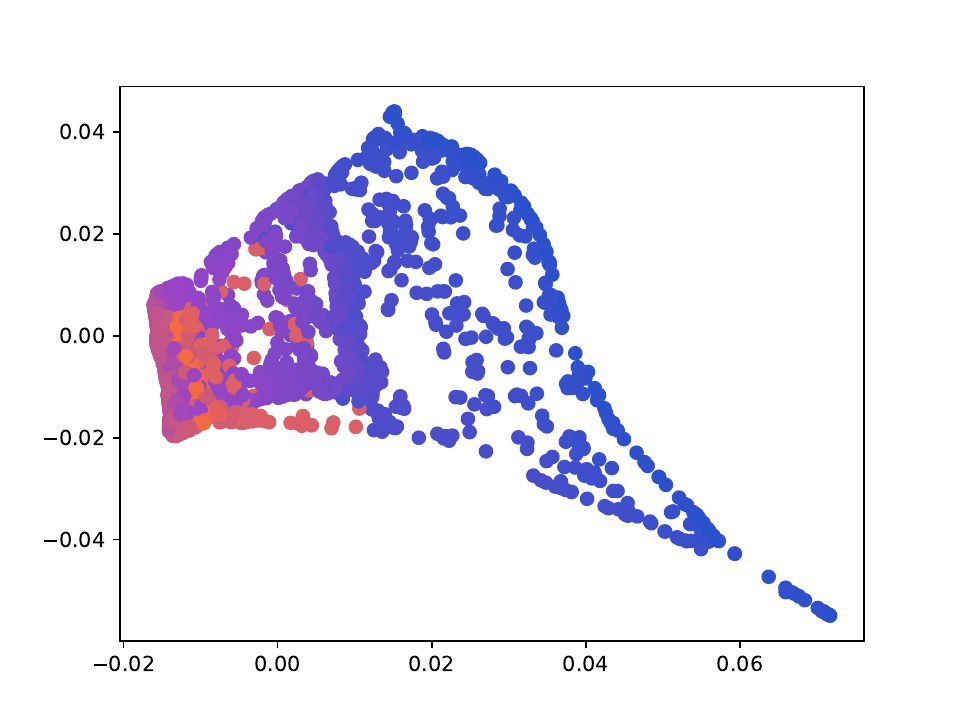}
      \end{subfigure}%
      \begin{subfigure}[t]{0.48\textwidth}
        \includegraphics[width=\textwidth]{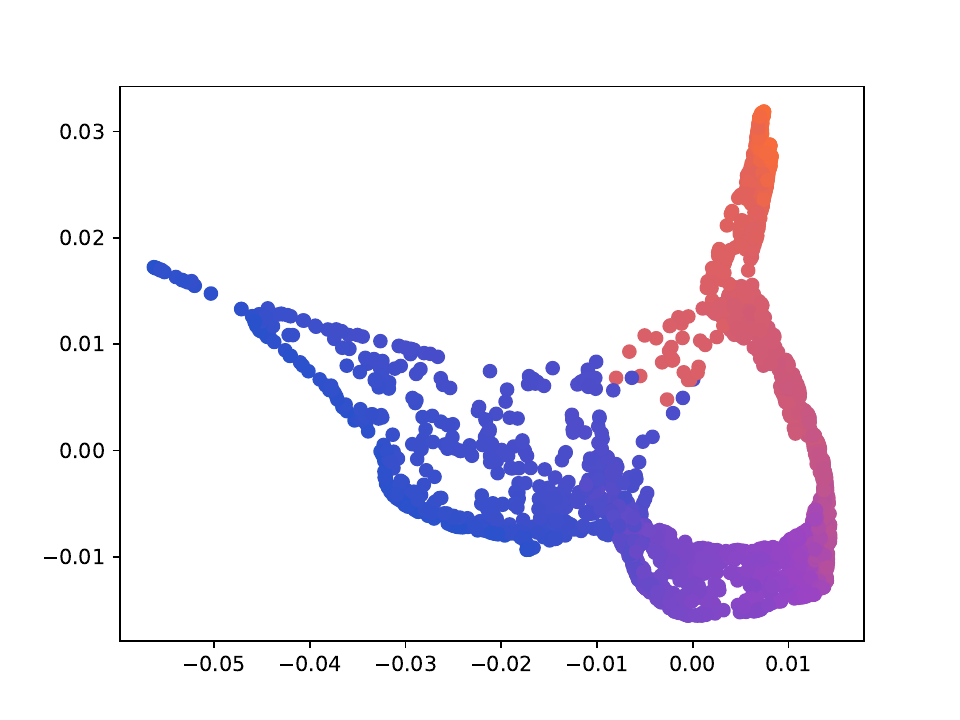}
      \end{subfigure}
    \end{minipage}
    \caption{Four \MDTrand embeddings}
  \end{subfigure} \quad
	\begin{subfigure}[b]{0.30\textwidth}
    \centering
    \includegraphics[width=\textwidth]{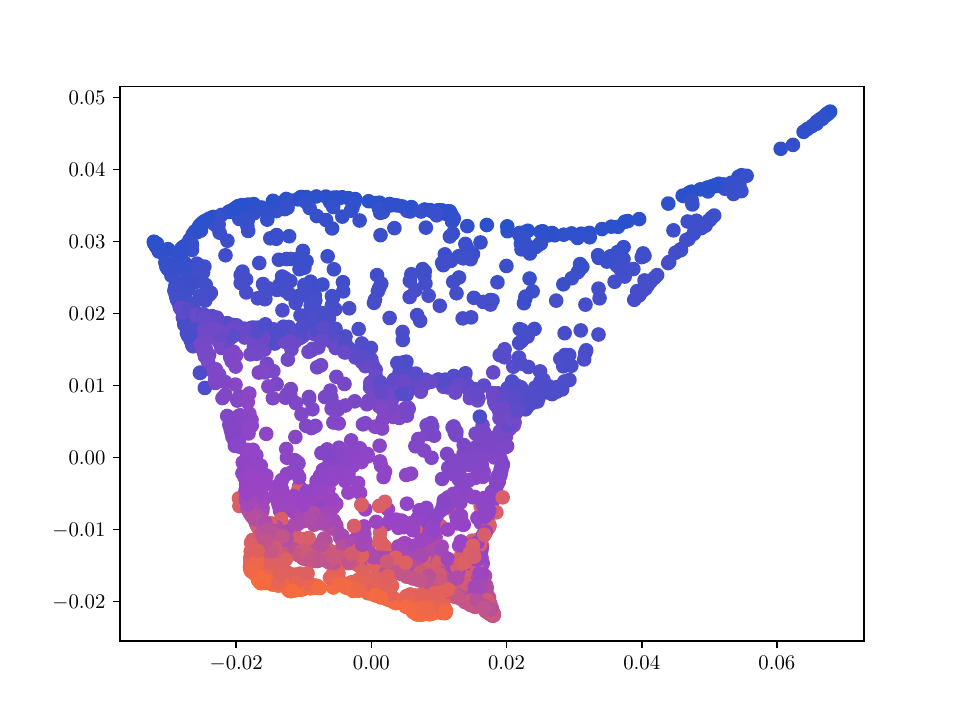}
    \caption{\MDT optimized for $Q_X$ ($t = 8$)}
  \end{subfigure}
  \begin{subfigure}[b]{0.30\textwidth}
    \centering
    \includegraphics[width=\textwidth]{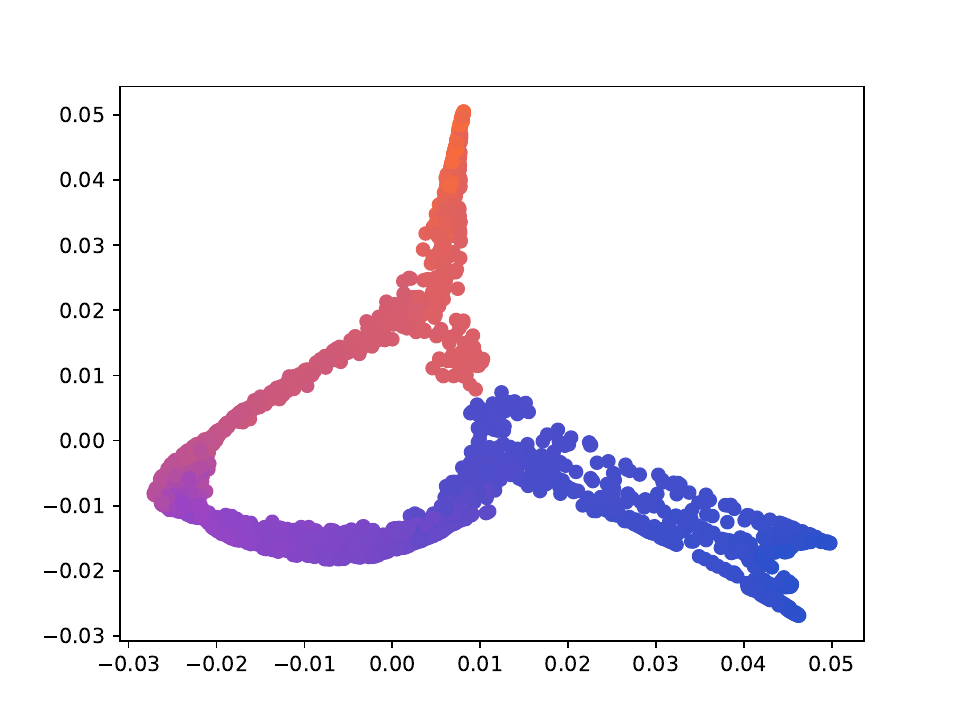}
    \caption{\MDT optimized for $Q_X$ ($t = 1$)}
  \end{subfigure}\\
  \caption{\textbf{Embeddings on the Deformed-Plane dataset.}~(a)~and~(b) show the two views of the dataset; (c)~\MVDM embedding (view I); (d)~\ADM embedding; (e)~\IDM embedding; (f)~embeddings obtained using four random \MDTs; (g)~the \MDT embedding obtained using a trajectory optimized for $Q_X$.~--~ Both \MVDM and \ADM struggle to recover the hidden manifold, with significant overlap between regions that should be distinct. \IDM captures an embedding with an 'omega' shape, essentially capturing the connectivity of the original manifold, although a slight overlap between the orange and blue points exists. The embeddings from randomly sampled \MDTs are diverse, including a case (bottom-right in (f)) that exhibits no overlap, and others with strong overlap. By contrast, optimizing the proposed criterion $Q_X$ yields a representation closer to the original manifold. At $t=1$, there is no overlap between regions, but the orange and blue regions are close to each others. At $t=8$, though, those regions are better separated, at the cost of a slight overlap.
  }
  \label{fig:deformed_manifold}
\end{figure*}

\inlinetitle{Datasets}{.}
Helix-A and Helix-B \citep{lindenbaum_multi-view_2020} are two synthetic datasets used to evaluate manifold learning algorithms, as they present challenges in capturing the underlying geometry of the data from multiple perspectives . They consist of $n = 1500$ points sampled from a $2$D spherical structure. The points are then embedded into a $3$D space using two different nonlinear transformations, resulting in two distinct views. Helix-A's datapoints are generated as $x_i = \{x_{i,1}, x_{i,2}\}_{i=1}^n$, where $x_{i, 1}$ are evenly spread in $[0, 2 \pi)$ and $x_{i, 2} = x_{i, 1} + \pi/2 \! \mod  2\pi$. View $1$ and view $2$ are generated by $\Phi_1$ and $\Phi_2$, respectively, those deformations are defined as follows; for $i \in \{1,...,n\}$:
\begin{align*}\textstyle
  & \Phi_1(x_i) = \begin{bmatrix}
  4 \cos(0.9 x_{i,1}) + 0.3 \cos(20 x_{i,1}) \\
  4 \sin(0.9 x_{i,1}) + 0.3 \sin(20 x_{i,1}) \\
  0.1(6.3 x_{i,1}^2 - x_{i,1}^3)
  \end{bmatrix},
  \\
  & \Phi_2(x_i) = \begin{bmatrix}
  4 \cos(0.9 x_{i,2}) + 0.3 \cos(20 x_{i,2}) \\
  4 \sin(0.9 x_{i,2}) + 0.3 \sin(20 x_{i,2}) \\
  0.1(6.3 x_{i,2}^2 - x_{i,2}^3)
  \end{bmatrix}.
\end{align*}
Helix-B stems from the previous one, where the definition of the views rely on the same input points $\{x_i\}_{i=1}^n$. The transformations are defined for $i \in \{1,...,n\}$ as:
\begin{equation*}
  \Phi_1(x) = 4 \begin{bmatrix}
   \cos(5 x_{i,1}) \\
   \sin(5 x_{i,1}) \\
   x_{i,1}
  \end{bmatrix},
  \, \textnormal{ and } \,
  \Phi_2(x) = 4 \begin{bmatrix}
   \cos(5 x_{i,2}) \\
   \sin(5 x_{i,2}) \\
   x_{i,2}
  \end{bmatrix}.
\end{equation*}
The Deformed-Plane dataset consists of points originally sampled uniformly on a plane. The plane is then deformed by two non-linear, non-bijective transformations that introduce connectivity not present in the original flat plane. \cref{fig:deformed_manifold} shows a visualization of this dataset.  Specifically, points $x = (x_1, x_2) \in \R^2$ are sampled as $x_1 \sim \gU(0, 3\pi/2)$ and $x_2 \sim \gU(0, 21)$, and two views are generated via the transformations $\Phi_1$ and $\Phi_2$, defined by
\begin{align*}
  \Phi_1(x) &=
  \begin{bmatrix}
    x_1 \cos(0.65x_1) \\
    0.2 x_2 + 0.3 \sin(x_1) \\
    x_1 \cos(x_1)
  \end{bmatrix},\\[2mm]
  \Phi_2(x) &=
  \begin{bmatrix}
    x_2 + \sin(2 x_1) + 0.4 \cos(x_2) \\
    x_1 + \sin(x_2) + 0.3 \cos(2x_1) \\
    \sin(x_1) + \cos(x_2)
  \end{bmatrix}.
\end{align*}

\begin{figure*}
  \centering
	\begin{subfigure}[t]{0.30\textwidth}
    \centering
		\includegraphics[width=\textwidth]{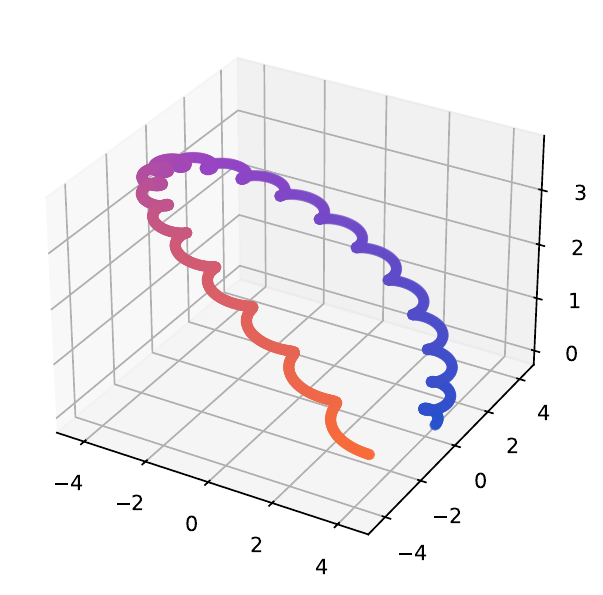}%
    \caption{View $1$}
  \end{subfigure} \quad%
\begin{subfigure}[t]{0.30\textwidth}
    \centering
		\includegraphics[width=\textwidth]{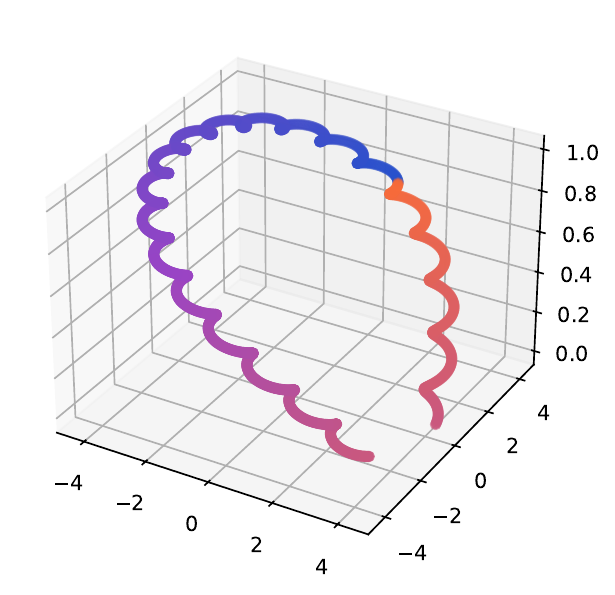}%
    \caption{View $2$}
  \end{subfigure}
  \\
  \begin{subfigure}[b]{0.30\textwidth}
    \centering
		\includegraphics[width=\textwidth]{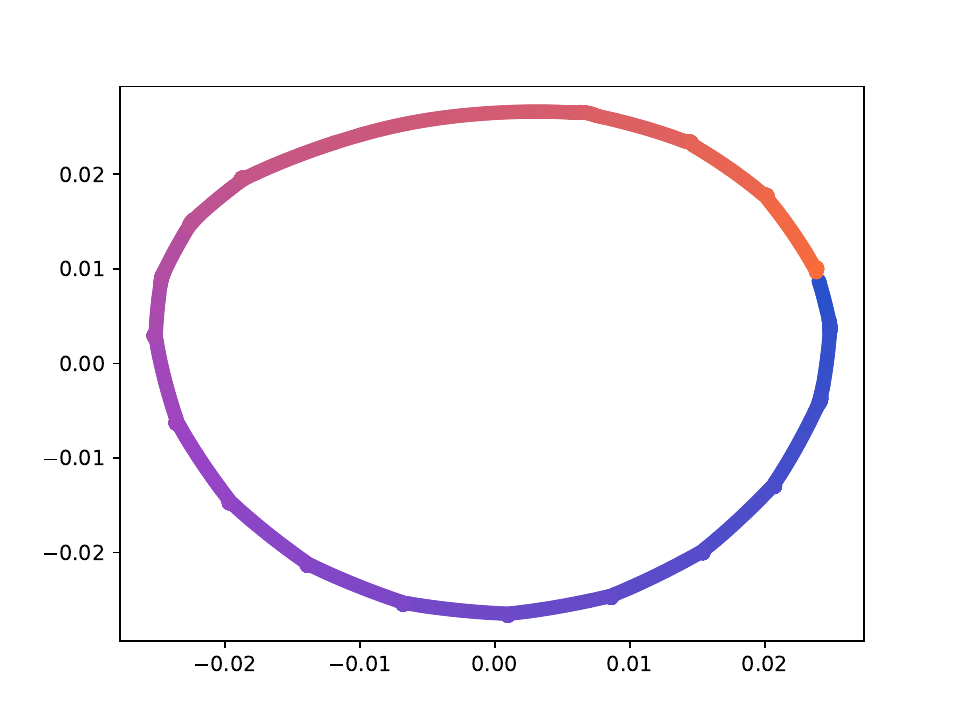}
    \caption{Multi-View Diffusion (\MVDM) -- View 1}
  \end{subfigure}\quad%
  \begin{subfigure}[b]{0.30\textwidth}
    \centering
		\includegraphics[width=\textwidth]{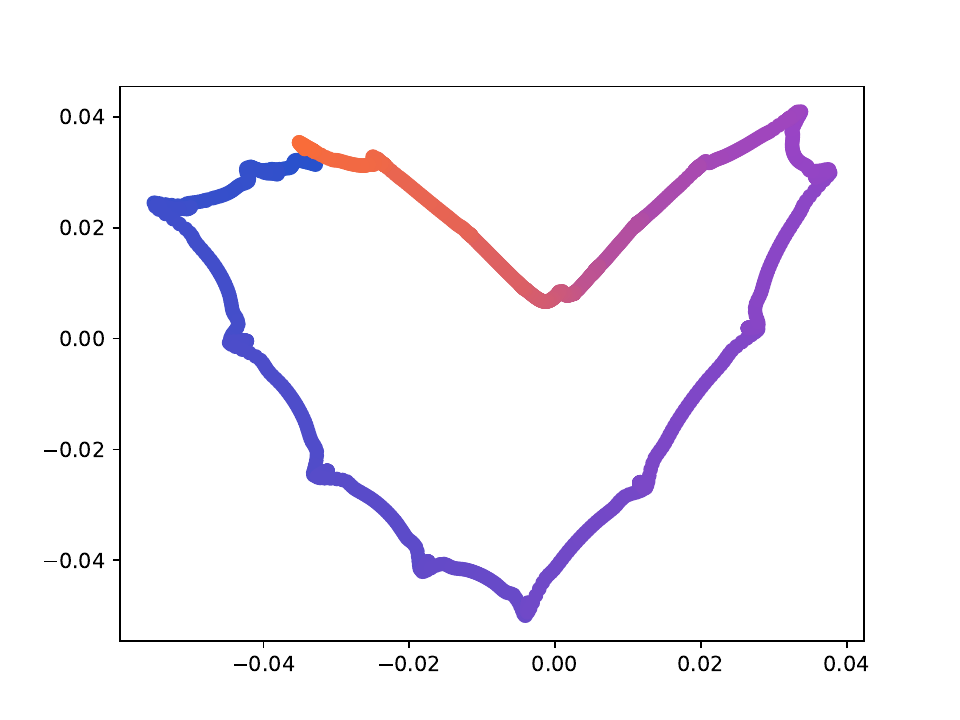}
    \caption{Alternating Diffusion (\ADM)}
  \end{subfigure}\quad%
  \begin{subfigure}[b]{0.30\textwidth}
    \centering
    \includegraphics[width=\textwidth]{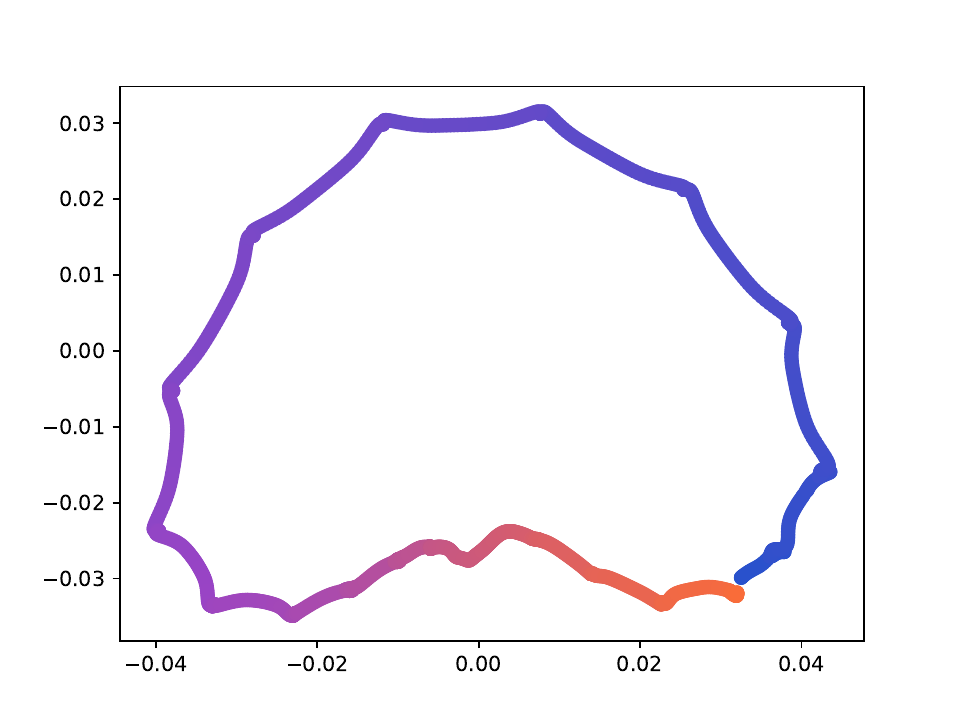}
    \caption{Integrated Diffusion (\IDM)}
  \end{subfigure}
  \\
   \begin{subfigure}[b]{0.30\textwidth}
    \centering
    \begin{minipage}{\textwidth}
      \centering
      \begin{subfigure}[t]{0.48\textwidth}
        \includegraphics[width=\textwidth]{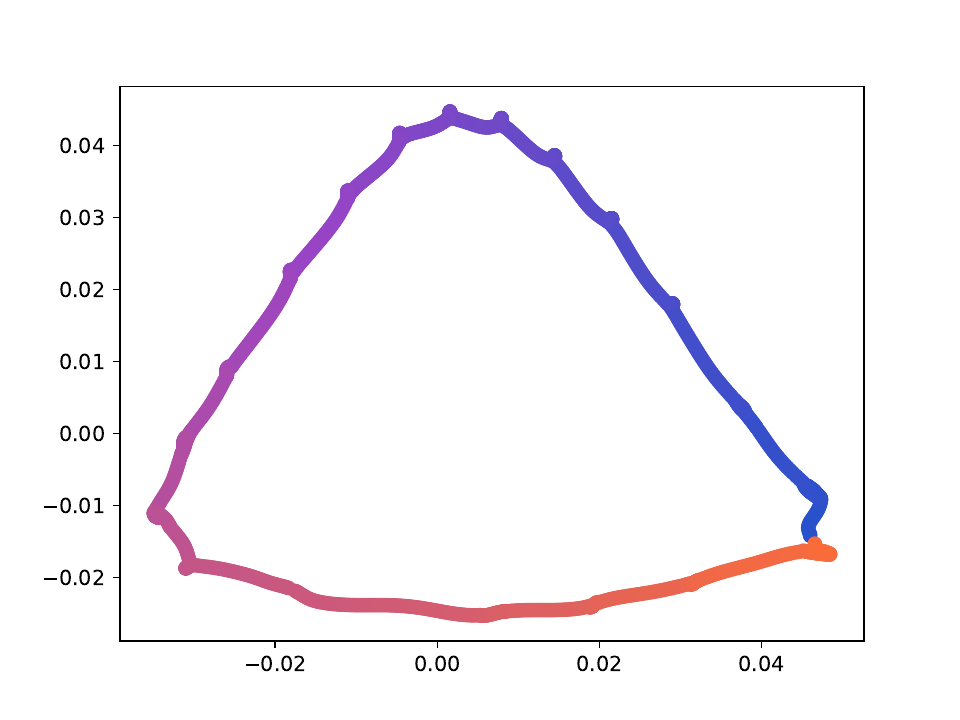}
      \end{subfigure}%
      \begin{subfigure}[t]{0.48\textwidth}
        \includegraphics[width=\textwidth]{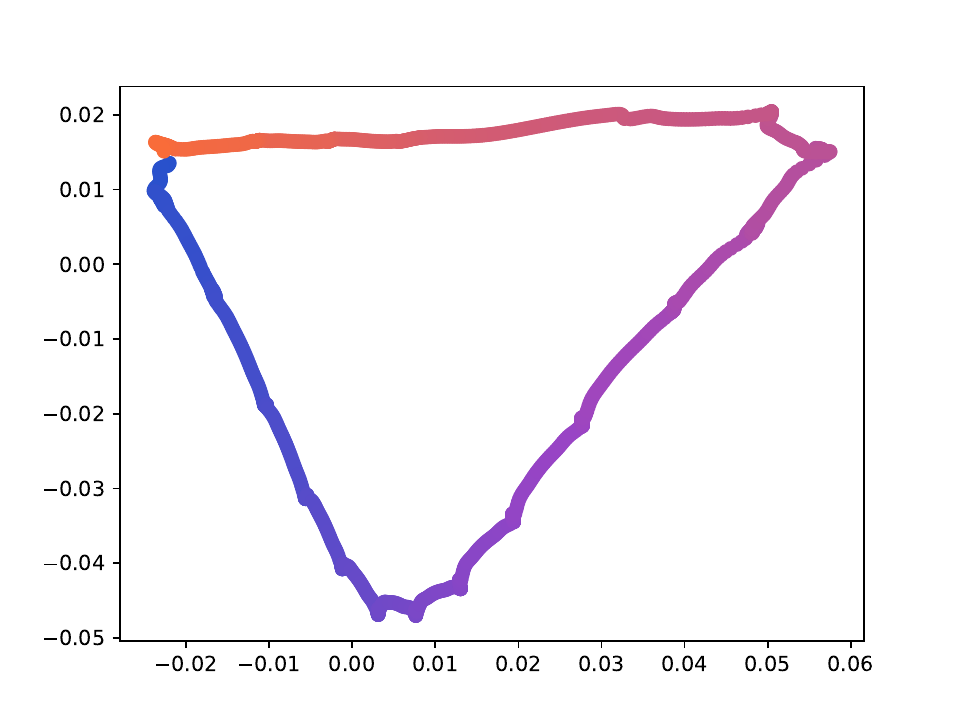}
      \end{subfigure} \\
      \begin{subfigure}[t]{0.48\textwidth}
        \includegraphics[width=\textwidth]{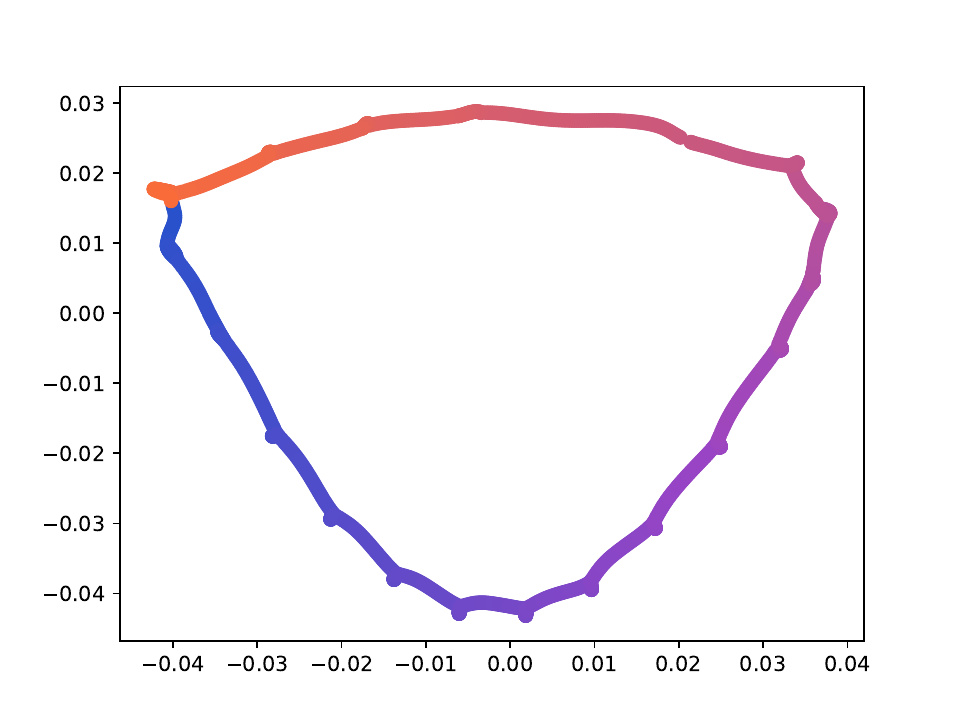}
      \end{subfigure}%
      \begin{subfigure}[t]{0.48\textwidth}
        \includegraphics[width=\textwidth]{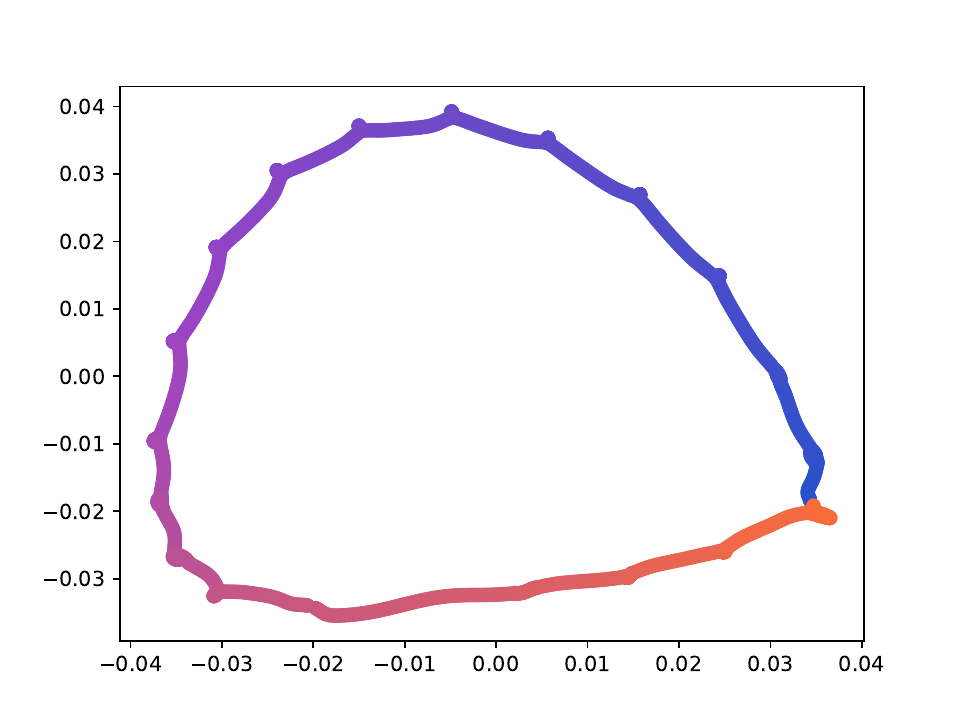}
      \end{subfigure}
    \end{minipage}
    \caption{Four \MDTrand embeddings}
  \end{subfigure} \quad
	\begin{subfigure}[b]{0.30\textwidth}
    \centering
		\includegraphics[width=\textwidth]{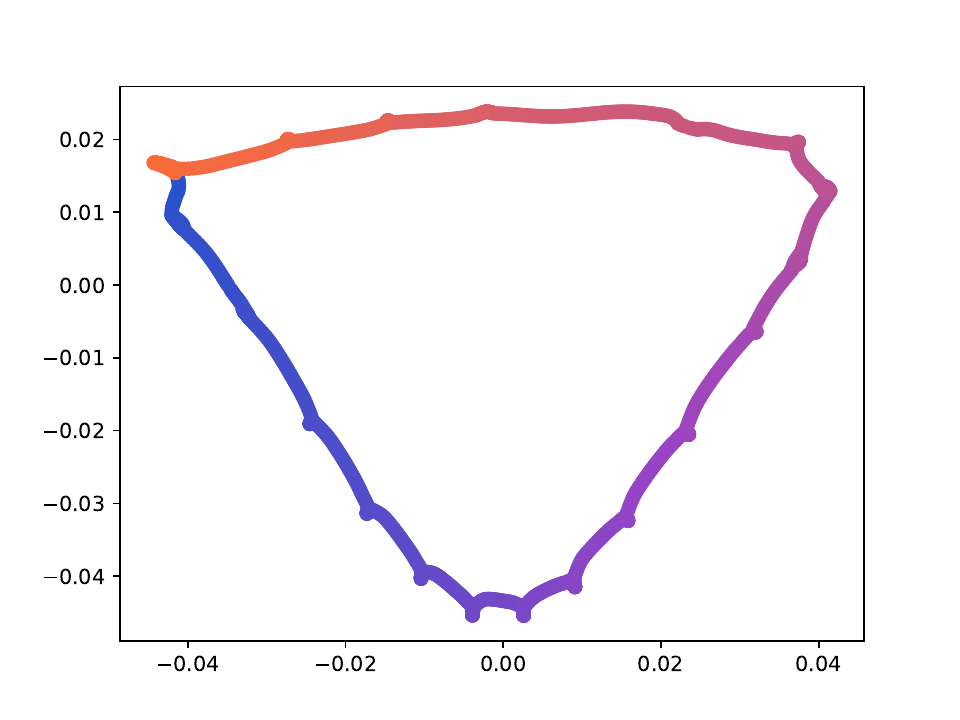}
    \caption{\MDT optimized for $Q_X$}
  \end{subfigure}\\

  \caption{\textbf{Embeddings on the Helix-A dataset.}~(a)~and~(b) show the two views of the dataset; (c)~\MVDM embedding (view I); (d)~\ADM embedding; (e)~\IDM embedding; (f)~embeddings obtained using four random \MDTs; (g)~the \MDT embedding obtained using a trajectory optimized for $Q_X$.~--~Methods that are based on the SVD of their related diffusion operator (\ADM, \IDM, and \MDTs) produce less smoothed embeddings compared to \MVDM, which relies on the eigendecomposition. This is particularly evident in the case of \ADM, which generates a pointy embedding. However, all methods seems to recover the circular structure of the underlying manifold in a rather meaninful manner. Moreover, we can observe that the embeddings obtained from randomly sampled \MDTs are similar, indicating that many trajectories in this case lead to satisfactory embeddings.
  }
  \label{fig:helixa}
\end{figure*}

\begin{figure*}[]
  \centering
    \begin{subfigure}[b]{0.30\textwidth}
    \centering
    \includegraphics[width=\textwidth]{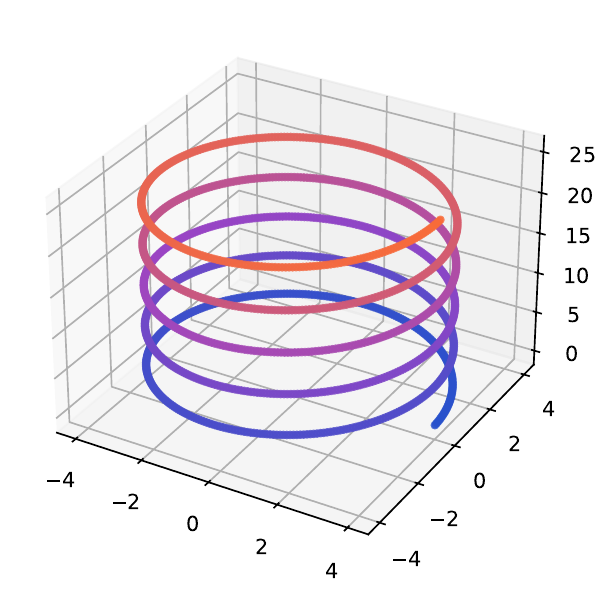}
    \caption{View $1$}
  \end{subfigure}\quad%
  \begin{subfigure}[b]{0.30\textwidth}
    \centering
    \includegraphics[width=\textwidth]{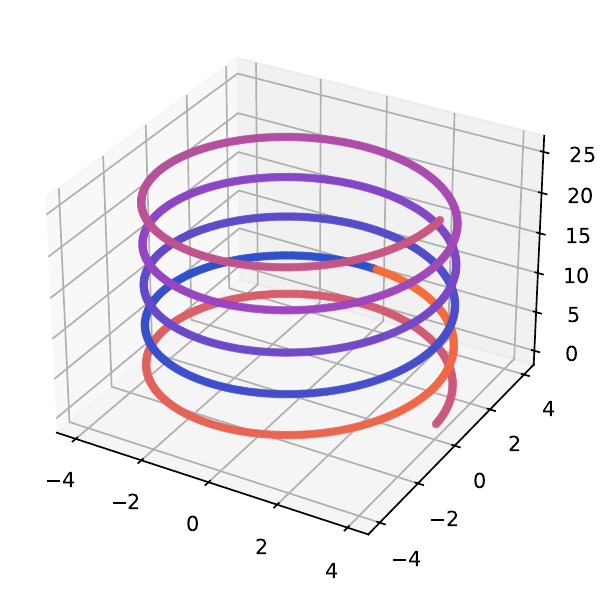}
    \caption{View $2$}
  \end{subfigure}
  \\
  \begin{subfigure}[b]{0.30\textwidth}
    \centering
		\includegraphics[width=\textwidth]{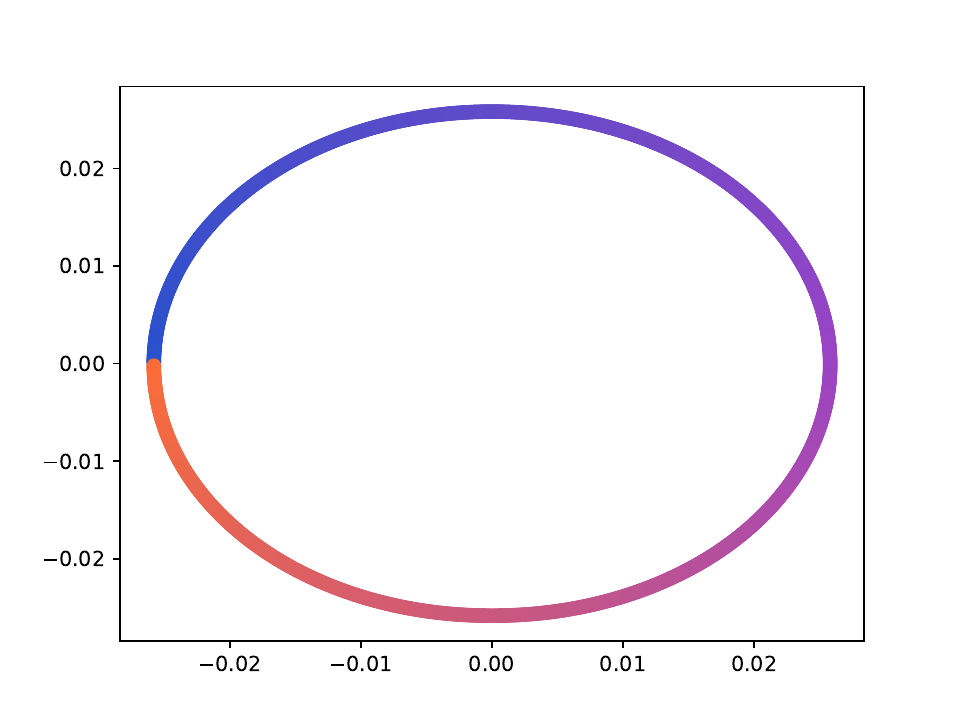}
    \caption{Multi-View Diffusion (\MVDM) -- View 1}
  \end{subfigure}\quad%
  \begin{subfigure}[b]{0.30\textwidth}
    \centering
		\includegraphics[width=\textwidth]{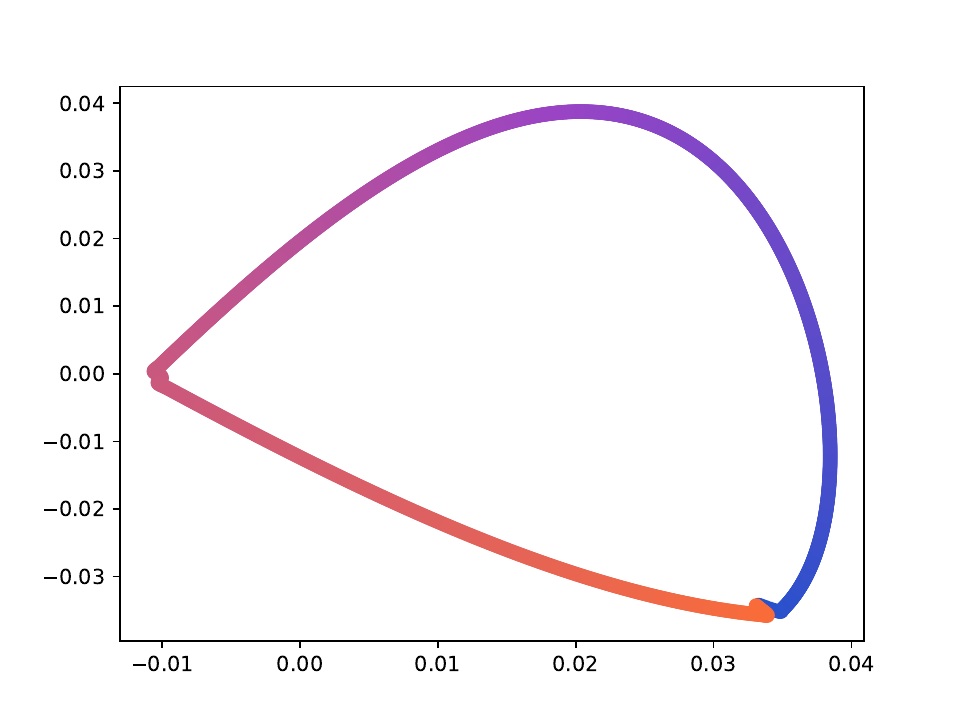}
    \caption{Alternating Diffusion (\ADM)}
  \end{subfigure}\quad%
   \begin{subfigure}[b]{0.30\textwidth}
    \centering
    \includegraphics[width=\textwidth]{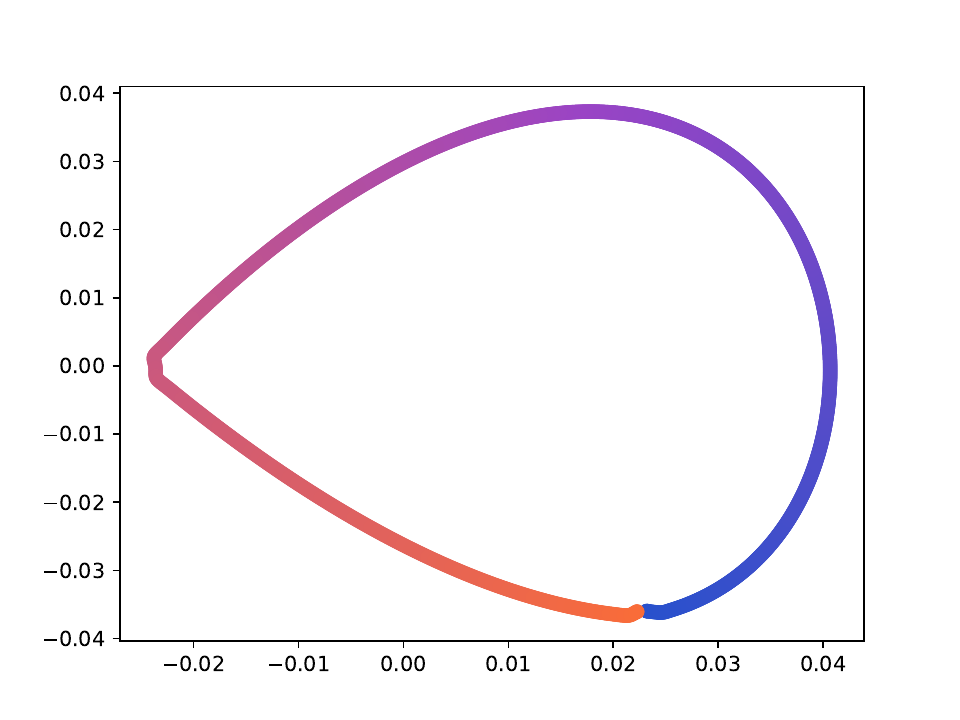}
    \caption{Integrated Diffusion (\IDM)}
  \end{subfigure}%
  \\
  \begin{subfigure}[b]{0.30\textwidth}
    \centering
    \begin{minipage}{\textwidth}
      \centering
      \begin{subfigure}[t]{0.48\textwidth}
        \includegraphics[width=\textwidth]{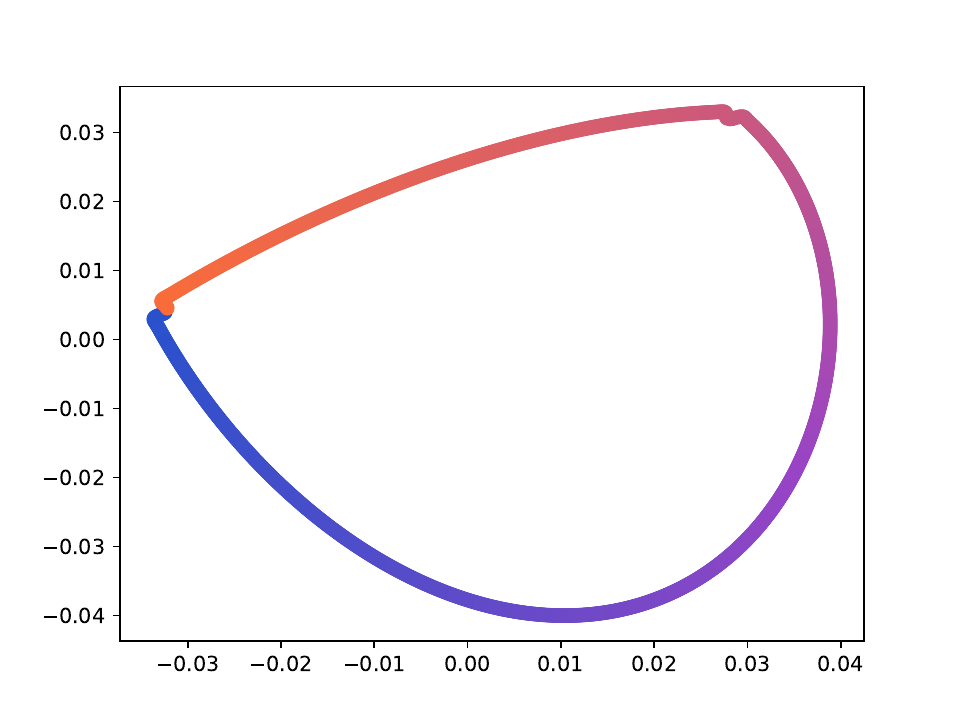}
      \end{subfigure}%
      \begin{subfigure}[t]{0.48\textwidth}
        \includegraphics[width=\textwidth]{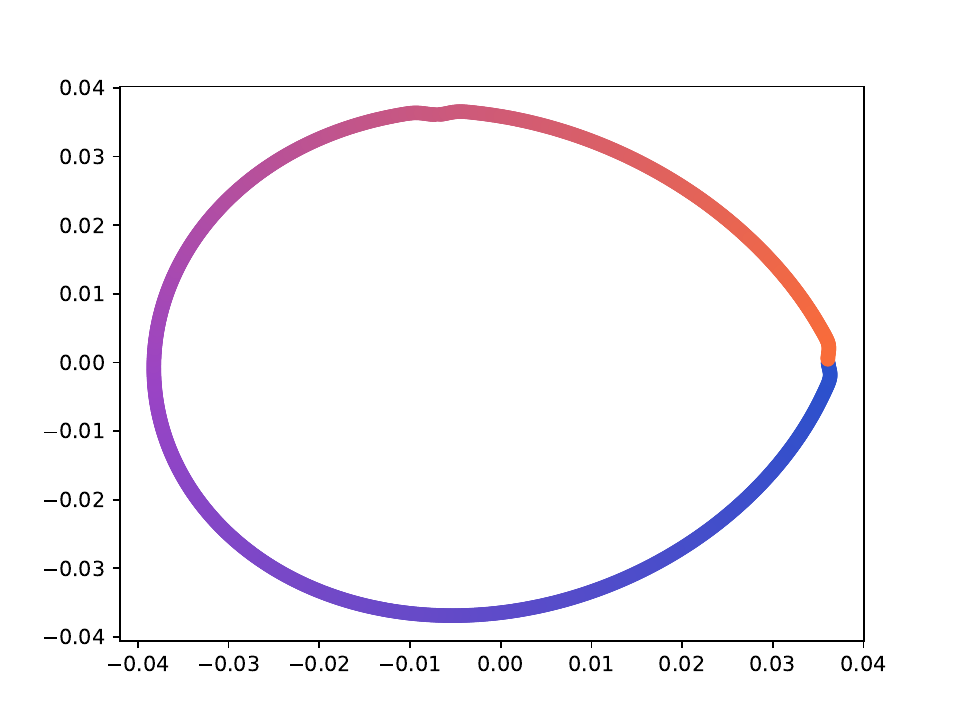}
      \end{subfigure} \\
      \begin{subfigure}[t]{0.48\textwidth}
        \includegraphics[width=\textwidth]{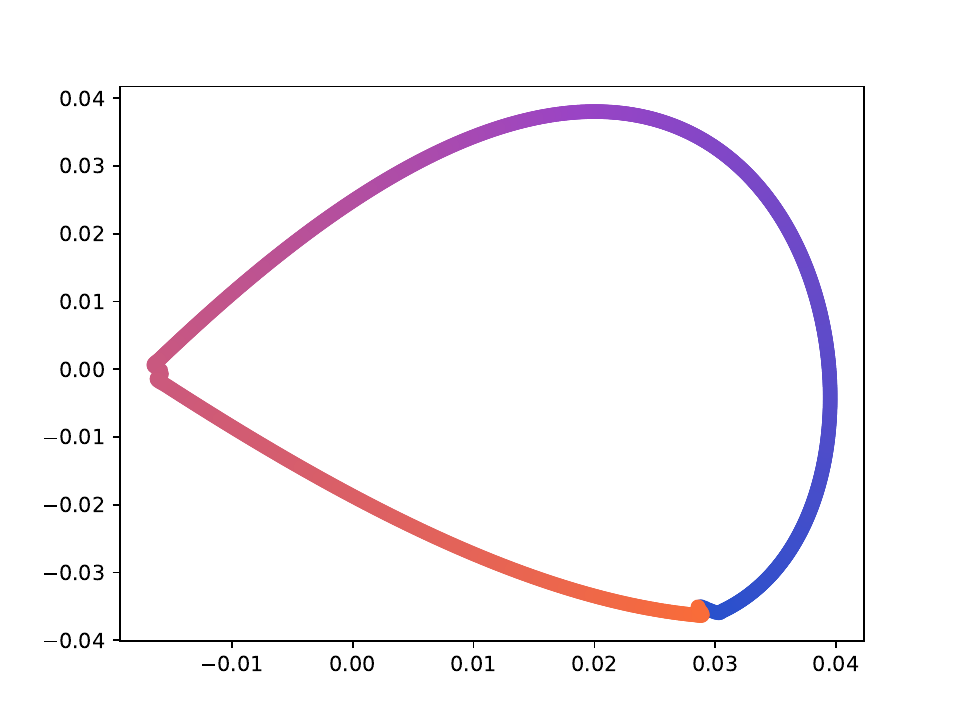}
      \end{subfigure}%
      \begin{subfigure}[t]{0.48\textwidth}
        \includegraphics[width=\textwidth]{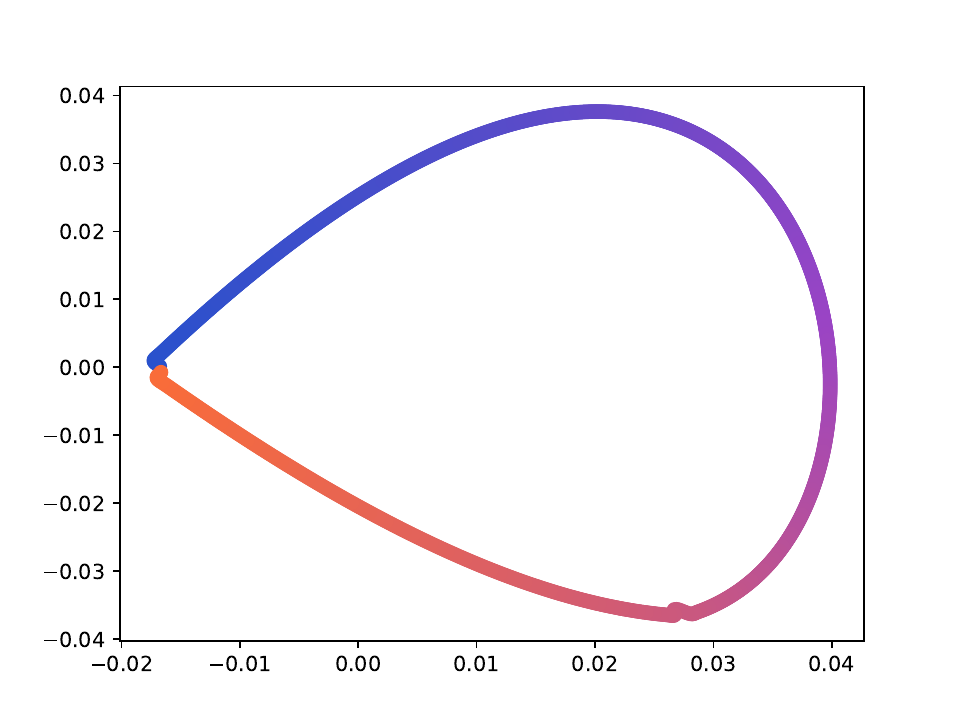}
      \end{subfigure}
    \end{minipage}
    \caption{Four \MDTrand embeddings}
  \end{subfigure} \quad%
	\begin{subfigure}[b]{0.30\textwidth}
    \centering
    \includegraphics[width=\textwidth]{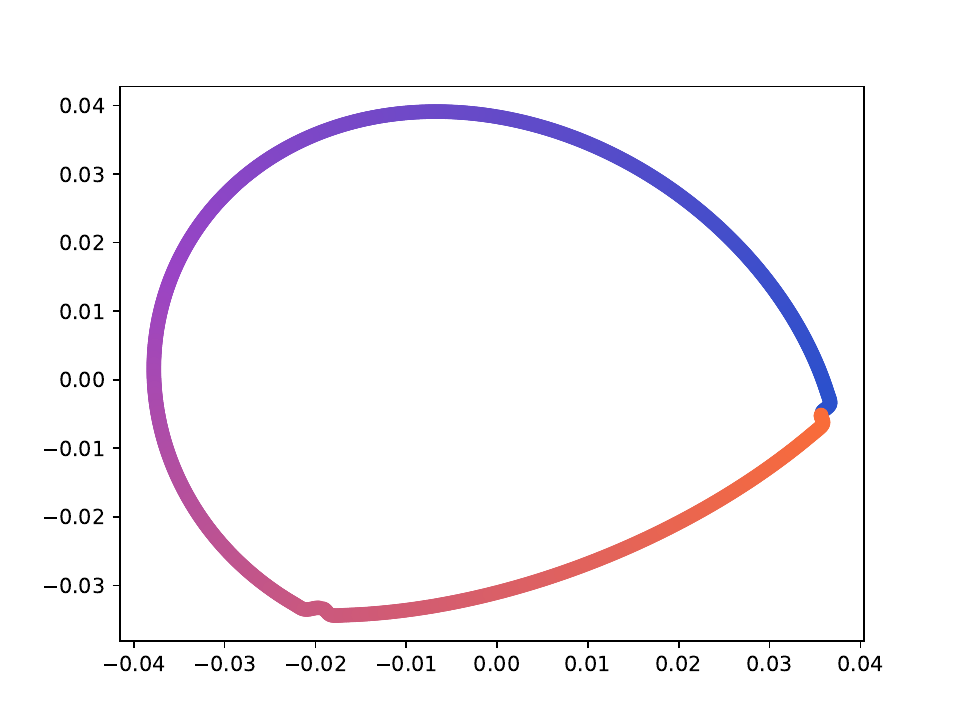}
    \caption{\MDT optimized for $Q_X$}
  \end{subfigure}\\

  \caption{\textbf{Embeddings on the Helix-B dataset.} (a)~and~(b) show the two views of the dataset; (c) first view of \MVDM (view I) ;(d) \ADM ;(e) \IDM; (f)~embeddings obtained using four random \MDTs; (g)~the \MDT embedding obtained using a trajectory optimized for $Q_X$.~--~ In this context, the methods perform comparably well. While some methods obtain pointy embeddings, others produce rounder ones, though all recover a circular structure.
  }

	 \label{fig:helixb}
\end{figure*}

\inlinetitle{Results}{.}
For the Helix-A and Helix-B datasets, the compared methods are used to generate two-dimensional embeddings of the original $3$D helices. The resulting embeddings are shown in \cref{fig:helixa} and \cref{fig:helixb}. In both cases the embeddings produced successfully capture the circular structure of the underlying manifold. Specifically, methods that rely on the eigen-decomposition tend to produce smoother embeddings that the ones relying on the SVD of the operator. This is particularly seen in the case of the Helix~A dataset, where \MVDM produces a very smooth embedding, while \ADM produces a pointy one. Those datasets serve as sanity checks for the multi-view manifold learning task.

For the Deformed-Plane dataset, both views exhibit clear limitations when attempting to recover the original underlying manifold. This highlights that combining these views requires careful consideration, leveraging only the parts of each view that are believed to be accurate. However, due to these inherent limitations, no view can perfectly recover the original manifold structure, as the transformation from the original manifold to the observed representation is non-bijective. The results for this dataset are presented in \cref{fig:deformed_manifold}. \MVDM and \ADM struggle to recover the underlying manifold, resulting in significant overlap between regions that should be distinct. \IDM captures an embedding with an 'omega' shape, effectively capturing the connectivity of the original manifold, although a slight overlap between the orange and blue points exists.
 Randomly sampled \MDTs produce similar embeddings, also showing strong overlap, except for the fourth case that produces a representation without any overlap. Two embeddings produces by \MDTCST are shown, one using the time parameter $t=1$, and another using $t=8$. Both recover representations that are `meaningful'. When using $t=1$, we obtain a representation that admits no overlap between the regions, but where the blue and orange regions are mapped close to each other. In the case of $t=8$, those regions are better separated, at the cost of a slight overlap.

Overall, the results highlight the merits of \MDT and the advantages of using the proposed contrastive loss for trajectory optimization in multi-view manifold learning.

\subsection{Application to data clustering}\label{sec:experiments:clustering}
\inlinetitle{External and internal quality indices}{.}~%
When \emph{true class labels} are available giving the true class of the datapoints in a benchmark dataset, the quality of a produced clustering partition can be evaluated through supervised measures. We specifically use the \emph{Adjusted Mutual Information} (AMI) measure \citep{JMLR:v11:vinh10a}, which is adjusted for the effect of agreement due to chance. Higher AMI values are preferred as they indicate partitions that match better with the partition that is induced by the true labels. We report the average value and standard deviation of those measures over $100$ runs of a clustering method in each case.

Regarding the learning of \MDTs, an internal quality index is needed to guide the process. Each such internal index imposes its own bias, in particular, optimal partitions according to a given index may not align with the optimal partition according to another index. As stated in \cref{sec:determine-sequence}, we consider the Calinski-Harabasz (CH) index~\citep{desgraupes_clustering_2016}, which is a widely used internal clustering validation measure. The CH index evaluates the quality of a clustering partition by considering the ratio of between-cluster dispersion to within-cluster dispersion. Higher CH values indicate better-defined clusters, as they reflect a larger separation between clusters relative to the compactness within clusters. When learning \MDT trajectories for clustering, we aim to maximize the CH index. An analysis of the correlation between AMI and CH on real-life datasets (see details in \appref{app:datasets}) shows that optimizing CH generally leads to improved AMI scores, although this relationship is not strictly monotonic.

\definecolor{mvdmcol}{HTML}{F7B5D4}
\definecolor{adcol}{HTML}{ACD86E}
\definecolor{padcol}{HTML}{B9E28C}
\definecolor{idmcol}{HTML}{3BB273}
\definecolor{comdcol}{HTML}{F49F0A}
\definecolor{crmvsccol}{HTML}{FFD333}
\definecolor{mvsccol}{HTML}{FFE070}
\definecolor{crdcol}{HTML}{E67E22}
\definecolor{affaddcol}{HTML}{E67E22}
\definecolor{affmulcol}{HTML}{E67E22}
\definecolor{dmbcol}{HTML}{6A4C93}
\definecolor{mdtchsdcol}{HTML}{0B93C1}
\definecolor{mdttreecol}{HTML}{07A0C3}
\definecolor{mdtconvcol}{HTML}{2851CC}
\definecolor{mdtrandcol}{HTML}{4469DA}
\definecolor{mdtbscdcol}{HTML}{07A0C3}

\begin{table*}
  \centering
  \begin{adjustbox}{width=\linewidth,center}
\begin{tabular}{l|ccccccccc|r}
\Xhline{1pt}
\multirow{2}{*}{\textbf{Methods}} & \multicolumn{9}{c|}{\textbf{Datasets}} & \multirow{2}{*}{\textbf{PRR}} \\
\cline{2-10}
&
\textbf{\textsc{K-MvMNIST}}* & \textbf{\textsc{L-MvMNIST}}* & \textbf{\textsc{Olivetti}} & \textbf{\textsc{Yale}} & \textbf{\textsc{100Leaves}} & \textbf{\textsc{L-Isolet}} & \textbf{\textsc{MSRC}} & \textbf{\textsc{Multi-Feat}}* & \textbf{\textsc{Caltech101-7}}* \\
\Xhline{0.75pt}
\textcolor{crdcol}{\rule{1em}{0.75em}} \CrD \ \ \ \ \ \citep{wang_unsupervised_2012} \smallg{(2012)} & 66.58 \mystd{1.32} & 59.51 \mystd{1.04} & 71.83 \mystd{1.19} & 56.16 \mystd{1.65} & 44.37 \mystd{0.68} & \mybest{79.79 \mystd{0.92}} & 48.32 \mystd{1.10} & 76.93 \mystd{1.44} & \mybest{65.92 \mystd{2.68}} & 0.93 \\
\textcolor{comdcol}{\rule{1em}{0.75em}}  \ComD\ \ \citep{shnitzer_recovering_2018} \smallg{(2018)} & 65.40 \mystd{2.06} & 61.82 \mystd{1.72} & 75.63 \mystd{1.02} & — & — & — & — & — & — & 0.99 \\
\textcolor{adcol}{\rule{1em}{0.75em}}  \ADM  \ \ \ \ \ \ \ \ \ \ \ \ \ \ \citep{katz_alternating_2019} \smallg{(2019)} & 66.25 \mystd{1.65} & 60.98 \mystd{1.99} & 75.08 \mystd{0.97} & 57.99 \mystd{1.22} & 80.01 \mystd{0.69} & 77.24 \mystd{0.56} & 50.03 \mystd{1.43} & 82.98 \mystd{1.39} & 56.71 \mystd{4.08} & 0.99 \\
\textcolor{mvdmcol}{\rule{1em}{0.75em}}  \MVDM \ \ \ \ \ \ \ \ \ \ \citep{lindenbaum_multi-view_2020} \smallg{(2020)} & \mybest{68.79 \mystd{2.28}} & 61.57 \mystd{1.55} & 75.82 \mystd{1.24} & 56.59 \mystd{0.62} & 82.43 \mystd{0.42} & 8.03 \mystd{0.64} & 48.84 \mystd{2.76} & 82.51 \mystd{1.95} & 56.05 \mystd{3.62} & 0.90 \\
\textcolor{idmcol}{\rule{1em}{0.75em}}  \IDM \ \ \ \ \ \ \ \ \ \ \ \, \,  \ \citep{kuchroo_multimodal_2022} \smallg{(2022)} & \mysecond{68.76 \mystd{1.89}} & 61.51 \mystd{0.97} & 74.98 \mystd{1.05} & 52.44 \mystd{1.03} & 73.82 \mystd{0.52} & \mysecond{79.77 \mystd{0.58}} & 35.16 \mystd{1.83} & 83.43 \mystd{1.07} & 45.77 \mystd{3.07} & 0.93 \\
\textcolor{padcol}{\rule{1em}{0.75em}}  \textsc{p-AD} \, \, \, \, \,   \,\citep{kuchroo_multimodal_2022} \smallg{(2022)} & \mysecond{68.76 \mystd{1.31}} & \mysecond{62.09 \mystd{1.03}} & 75.52 \mystd{0.96} & 57.74 \mystd{1.18} & 69.94 \mystd{0.68} & 78.77 \mystd{0.59} & 40.97 \mystd{2.07} & 81.88 \mystd{1.32} & 58.18 \mystd{3.18} & 0.97 \\
\Xhline{0.75pt}
\textcolor{mdtchsdcol}{\rule{1em}{0.75em}}  \MDTchs & 68.53 \mystd{1.83} & \mybest{62.24 \mystd{1.08}} & \mybest{76.91 \mystd{1.07}} & \mybest{59.40 \mystd{0.93}} & \mybest{91.35 \mystd{0.83}} & 77.91 \mystd{0.63} & \mybest{76.18 \mystd{2.64}} & \mysecond{85.37 \mystd{1.01}} & 63.56 \mystd{3.42} & \mybest{1.08} \\
\textcolor{mdtbscdcol}{\rule{1em}{0.75em}}  \MDTbsc & 68.39 \mystd{1.21} & 61.77 \mystd{0.92} & 75.05 \mystd{1.33} & 58.57 \mystd{1.31} & 61.73 \mystd{0.40} & 79.61 \mystd{0.65} & 73.67 \mystd{3.48} & \mybest{87.32 \mystd{0.67}} & \mysecond{64.05 \mystd{3.09}} & 1.03 \\
\textcolor{mdtrandcol}{\rule{1em}{0.75em}}  \MDTrand & 68.20 \mystd{1.98} & 61.33 \mystd{1.47} & 75.95 \mystd{1.52} & 57.14 \mystd{1.74} & 70.51 \mystd{6.01} & 77.78 \mystd{0.70} & 61.27 \mystd{10.84} & 80.28 \mystd{4.25} & 58.57 \mystd{7.18} & 1.00 \\
\textcolor{mdtconvcol}{\rule{1em}{0.75em}} \MDTcvx & 68.51 \mystd{1.53} & 61.69 \mystd{1.12} & \mysecond{76.49 \mystd{1.24}} & \mysecond{58.68 \mystd{1.11}} & \mysecond{86.19 \mystd{4.66}} & 78.00 \mystd{0.62} & \mysecond{74.62 \mystd{3.40}} & 84.84 \mystd{1.87} & 63.46 \mystd{3.64} & \mysecond{1.07} \\
\Xhline{1pt}
\end{tabular}
  \end{adjustbox}
  \caption{\textbf{Clustering results across multiple datasets.} For each column, the highest and second-highest mean scores are shown in bold and underlined, respectively. Standard deviations are given in parentheses. An asterisk (*) indicates a dataset where methods fails to outperform the `best' single-view baseline. The Performance Ratio to Random (PRR, \cref{eq:PRR}) at the rightmost column quantifies each method's average performance as a ratio to the discrete \MDTrand baseline, with values above $1$ indicating superior performance. `—' indicates that \ComD results are not available for those datasets, due to the two-view limitation of the method.}
  \label{tab:results_real_ds}
\end{table*}

\inlinetitle{Datasets}{.}~%
The Multiple Features dataset (Multi-Feat) contains $2000$ handwritten digits that are extracted from Dutch utility maps\footnote{UCI repository at: \href{http://archive.ics.uci.edu/dataset/72/multiple+features}{http://archive.ics.uci.edu/dataset/72/multiple+features}.}. Each digit is represented using six distinct feature sets: Fourier coefficients of the character shapes, profile correlations, Karhunen-Loève coefficients, pixel averages in $2 \times 3$ windows, Zernike moments, and morphological descriptors.

\begin{figure}[t!]
  \begin{subfigure}[t]{0.450\textwidth}
    \centering
		\includegraphics[width=\textwidth,viewport=35pt 20pt 630pt 390pt,clip]{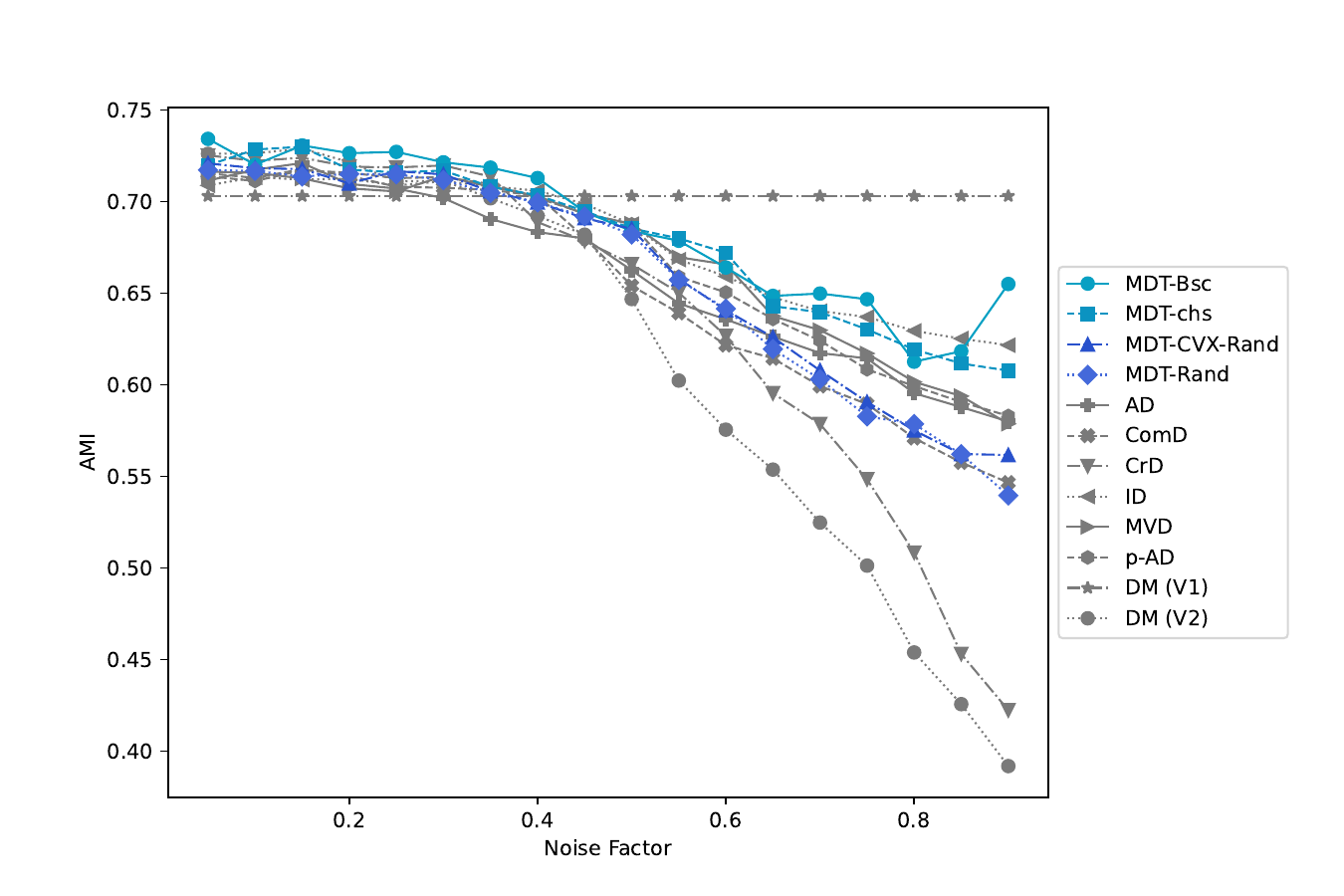}
    \caption{K-MvMNIST}
    \label{fig:scalnoise:Kmnist}
  \end{subfigure} \hfill \\
	\begin{subfigure}[t]{0.450\textwidth}
    \centering
		\includegraphics[width=\textwidth,viewport=35pt 20pt 630pt 390pt,clip]{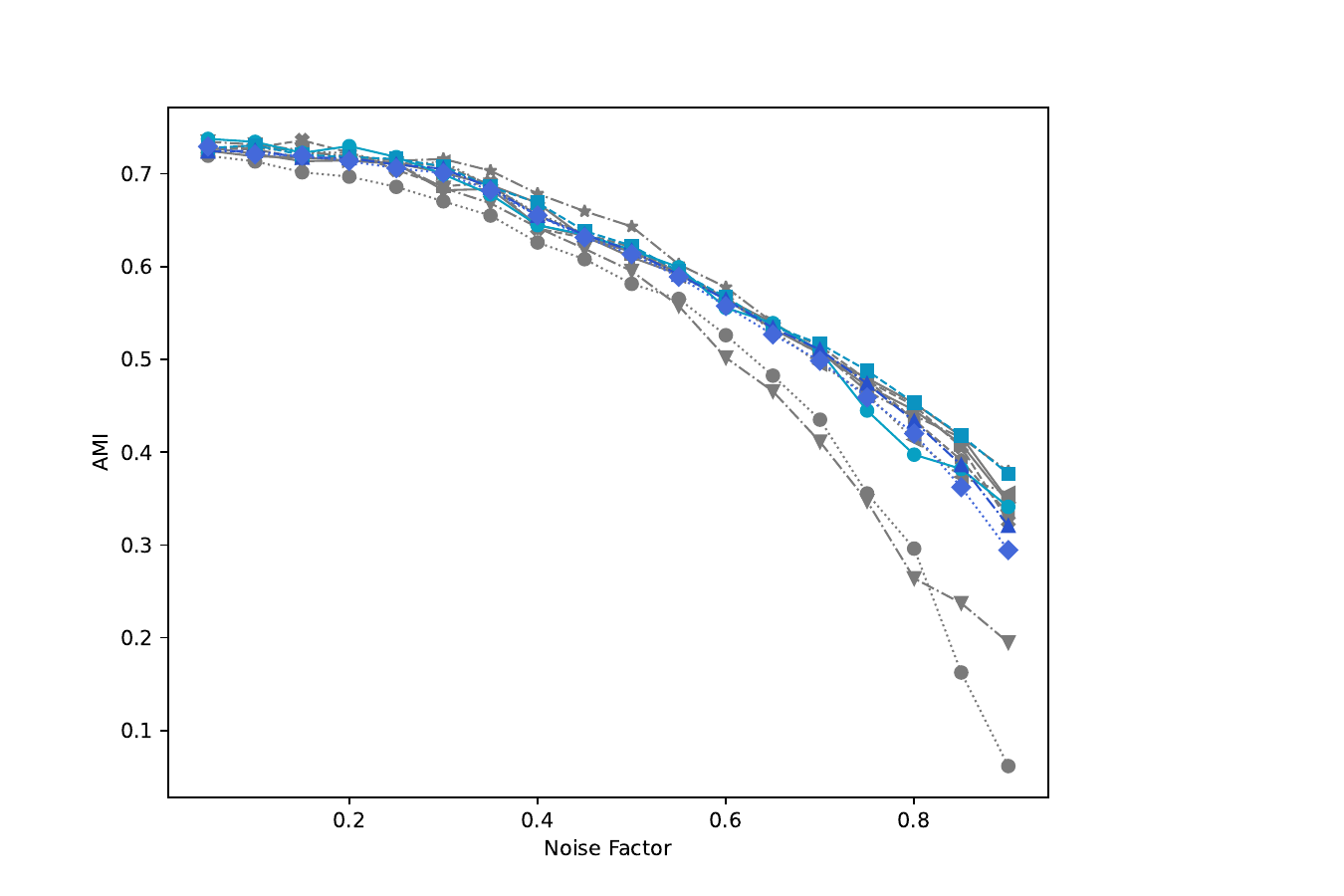}
    \caption{L-MvMNIST}
    \label{fig:scalnoise:Lmnist}
  \end{subfigure}
  \caption{\textbf{Robustness and power.}
	The clustering performance of the compared methods as the noise grows for L/K-Mv-MNIST. }
  \label{fig:scalnoise}
\end{figure}

K-MvMNIST \citep{kuchroo_multimodal_2022} and L-MvMNIST \citep{lindenbaum_multi-view_2020} are two multi-view variants of the MNIST dataset, which are inspired by prior work. K-MvMNIST has $2$-views, one consist of the original images with an added fixed Gaussian noise $\mathcal{N}(0, 0.3)$, and the other view is obtained by adding a different Gaussian $\gN(0, s)$, where $s \in [0,1)$ is a robustness parameter. The impact of the second Gaussian noise on clustering results is shown in \cref{fig:scalnoise:Kmnist}. L-MvMNIST follows a similar setup, it defines a robustness parameter $s \in [0,1)$, then the first view is created by adding Gaussian noise $\gN(0, s)$ to the original image, the second view by randomly dropping each pixel with a $s\%$ probability. Unlike prior work \cite{lindenbaum_multi-view_2020} that used only digits $2$ and $3$ drawn from the infiniMNIST dataset, we use here the original MNIST dataset, downsampled to $6000$ samples. In both cases we set by default $s = 0.5$, while the importance of this parameter is analyzed in \cref{fig:scalnoise:Kmnist} and \cref{fig:scalnoise:Lmnist}.

The L-Isolet dataset \cite{lindenbaum_multi-view_2020} is a multi-view dataset generated from single point-cloud view using different kernels. Its $1600$ datapoints are observed in either $3$ views:
\begin{align*}
  \kappa_1 (x_i, x_j) & = {\textstyle\exp\left(-\frac{\norm{x_i - x_j}^2}{2 \sigma^2} \right)}, \\
  \kappa_2 (x_i, x_j) & = {\textstyle\exp\left(-\frac{\norm{x_i - x_j}}{\sigma}\right)}, \\
  \kappa_3 (x_i, x_j) & = {\textstyle\exp \left(\frac{\rmT_{i, j} - 1}{2 \sigma}\right)},
\end{align*}
where $\rmT_{ij}$ is the correlation between $x_i$ and $x_j$, and
$\sigma$ follows the max-min heuristic. This dataset evaluates whether clustering methods can detect informative kernels and merge heterogeneous structural information.

\newpage
The next five datasets are all available online\footnote{Olivetti is available online at:~\\\href{https://www.cl.cam.ac.uk/research/dtg/attarchive/facedatabase.html}{{https://www.cl.cam.ac.uk/research/dtg/attarchive/facedatabase.html}}; \\the four other datasets are available at:~\\\href{https://github.com/ChuanbinZhang/Multi-view-datasets/}{https://github.com/ChuanbinZhang/Multi-view-datasets/}}.
$\bullet$~The Olivetti dataset consists of grayscale facial images represented through two feature descriptors: a HOG transform of the datapoints, and a $150$ principal component analysis (PCA) representation, offering complementary texture and structural information.
$\bullet$~
Yale faces dataset consists of $165$ gray-scale face images belonging to $15$ subjects with each subject containing $11$ images. Each image is represented using three feature sets: a Gabor filter-based descriptor, a Local binary patterns feature, and a raw pixel intensity vector. $\bullet$~%
The 100leaves dataset is multi-class dataset where each leaf sample is characterized through several shape and texture-oriented feature views, capturing different morphological aspects of the same object. In particular, three views are used: shape descriptors, fine scale margin and texture histogram.
$\bullet$~
A subset of the Microsoft Research in Cambridge dataset (MSRC) is used, which contains images represented through five complementary feature views: Color moments, Histogram of oriented gradients feature, GIST feature, Local binary patterns feature and CENTRIST feature.
$\bullet$~
The Caltech101-7 dataset contains $1474$ images belonging to seven classes, which are faces, motorbikes, dollar bill, Garfield, stop sign, and windsor chair. Each image is represented through six feature views: Gabor-feature, wavelet moments, CENTRIST-feature, Histogram of oriented gradients , GIST-feature, Local binary patterns feature.

\begin{figure}[t]
  \centering
  	{\scriptsize\hspace{1.5em}--------------- \MDT ---------------\ \ --------------------- literature ---------------------}\\
  \includegraphics[width=\columnwidth]{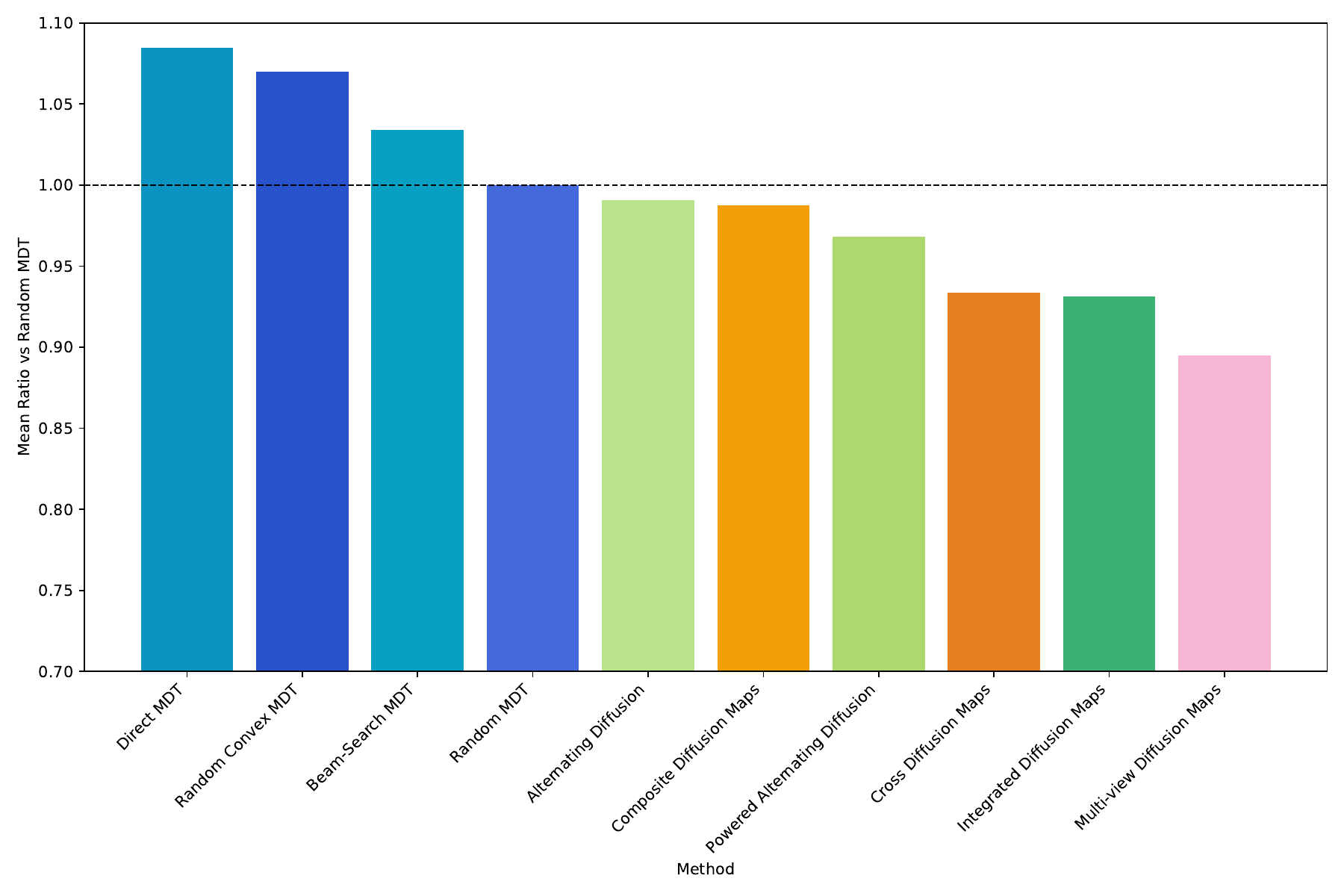}
  \vspace{-2em}
   \caption{\textbf{Clustering performance ratio comparing different methods to the discrete \MDTrand (\ie sampled from $\Psetc$).} Each bar shows the PRR index (\cref{eq:PRR}), which is the ratio of the AMI obtained by a method to that obtained by \MDTrand. Ratio values below $1$ indicate worse average clustering performance than \MDTrand, while values above $1$ indicate better performance. The results highlight that the three first designed \MDT variants outperform the plain random \MDTrand counterpart, all other diffusion-based methods, some of which encompassed by the \MDT framework, come after and importantly they do not perform better than our \MDTrand baseline.}
  \label{fig:perf_index}
\end{figure}

\inlinetitle{Random \MDTs as a baseline}{.}~%
Beyond their mere performance, \MDTs also provide a valuable baseline tool for analyzing multi-view diffusion behavior. Since the \MDT framework is general and parameterized only by the operator set ($\Pset$) and the optimization strategy to determine a trajectory, one can instantiate \MDTs with minimal assumptions to check how view interactions influence learning outcomes. In this sense, \MDTrand can serve as a `neutral' diffusion baseline: random or uniformly sampled trajectories represent an unbiased mixture of views, while convex or optimized variants reveal how adaptive view weighting affects manifold recovery or clustering. Through this lens, \MDTrand constitutes a principled and interpretable reference point for evaluating multi-view fusion strategies, which is particularly useful for diagnostic or benchmarking purposes, \eg when assessing the sensitivity of downstream tasks to the order or frequency of cross-view diffusion. To employ this idea we introduce the Performance Ratio to Random (PRR) index, which is the AMI ratio of a method to the \MDTrand baseline in the canonical set $\Psetc$, which treats all views and their combinations evenly. The average PRR across datasets is reported in \cref{fig:perf_index} and \cref{tab:results_real_ds}:
\begin{equation}\label{eq:PRR}
    \text{PRR}(\text{method}) = \frac{\text{AMI}(\text{method})}{\text{AMI}(\MDTrand)}.
\end{equation}

\inlinetitle{Results}{.}~%
\cref{tab:results_real_ds} compares a range of multi-view clustering approaches across nine datasets of varying difficulty and structure. Experiments are made over $100$ runs, and reports mean AMI and standard deviation. Additional analysis on the noise-factor-based datasets is given in \cref{fig:scalnoise:Kmnist,fig:scalnoise:Lmnist}.

\noindent\emph{Comparative performance.}~
In \cref{tab:results_real_ds}, \MDT variants based on the convex set $\Psetcvx$ (\MDTcvx, \MDTchs) consistently achieve strong clustering performance across datasets, often outperforming established methods such as \MVDM, \ADM, \pADM, and \IDM. This underscores the importance of carefully designing $\Pset$ for \MDTs' success. Notably, \MDTchs attains the highest AMI on five of the nine datasets and ranks second on one. Although \MDTcvx does not achieve the top score on any dataset, it performs consistently well, suggesting that convex interpolation of views provides a robust strategy. As \MDTbsc can evaluate trajectories of varying lengths, it can adaptively select diffusion scales for each dataset, which is advantageous when a fixed diffusion time heuristic fails to capture the appropriate scale. However, this flexibility can result in sub-optimal choices, notably when the internal criterion doesn't align with the ground-truth labels, as illustrated by the 100Leaves dataset, where \MDTbsc underperforms noticeably.

Overall, these results highlight the flexibility and effectiveness of the \MDT framework when paired with thoughtfully designed trajectory spaces and optimization strategies. Looking at the PRR index in \cref{fig:perf_index}, we observe that the three first designed \MDT variants outperform the plain random \MDTrand counterpart, all other diffusion-based methods, some of which encompassed by the \MDT framework, come after and importantly they do not perform better than our \MDTrand baseline. This further emphasizes the strength of the \MDT framework and the importance of trajectory design.

\noindent\emph{Robustness to noise.}
\cref{fig:scalnoise} illustrates how different methods behave as the noise factor $s$ increases. In K-MvMNIST, only the second view is affected by noise, so methods that treat both views equally show a faster decline in performance compared to those employing weighting strategies (e.g., \IDM, \MDTcvx, \MDTbsc), which seems to attain similar performance decay with the addition of noise. In L-MvMNIST, the quality of both views deteriorates substantially with increasing $s$, and all methods experience a corresponding drop in AMI scores.

These experiments demonstrate the robustness of \MDTs when paired with appropriate trajectory designs and optimization strategies: they can leverage the more informative views while reducing the impact of noisier or less reliable ones.

\section{Conclusions}

This work introduced Multi-view Diffusion Trajectories \\ (\MDTs), a unified operator-based framework for constructing diffusion geometries across multiple data views. By modeling multi-view diffusion as a time-inhomogeneous process driven by sequences of view-specific operators, \MDTs generalize a family of existing multi-view diffusion methods, including Alternating Diffusion and Integrated Diffusion, while offering significantly greater flexibility. In that sense, our framework provides a common language in which these methods can be analyzed, compared, and extended.

We established fundamental properties of \MDTs under mild assumptions and formulated trajectory-dependent diffusion distances and embeddings via singular value decompositions of the induced operators. This provides a principled extension of classical diffusion maps to settings where the diffusion law evolves over time. The proposed trajectory space further enables, for the first time, operator learning within diffusion-based multi-view fusion, opening an avenue for view-adaptive diffusion processes that can selectively emphasize complementary structures across views.

Empirical evaluation on both manifold learning and clustering tasks shows that \MDTs are expressive and practically effective. Both discrete and convex variants achieve competitive or superior performance compared to established multi-view diffusion approaches, showing the benefit of structured operator sequences over fixed fusion rules. Notably, randomly sampled trajectories (\MDTrand), relying only on a chosen time parameter, achieve high performance, highlighting the richness of the MDT operator space and indicating that even simple exploration strategies can yield informative multi-view geometries. Moreover, \MDTrand serves as a useful baseline for evaluating diffusion-based multi-view methods, as it is not consistently outperformed by more sophisticated approaches in practice.

Overall, \MDTs provide a coherent, extensible, and computationally tractable framework for multi-view diffusion. Their flexibility in operator design and compatibility with principled learning criteria make them a powerful foundation for future developments in multi-view representation learning. The unified view offered by \MDTs can guide the design of new multi-view diffusion models and contribute to a deeper understanding of how heterogeneous data sources can be fused through iterative geometric processes. Potential directions include integrating supervised objectives into trajectory learning, improving scalability via graph sparsification, and exploring connections with neural or contrastive graph-based methods. Exploring continuous-time MDT formulations, could underpin strong connexions analogous to heat kernels.

\section{Acknowledgments}
This work was supported by the Industrial Data Analytics and Machine Learning Chair hosted at ENS Paris-Saclay, and by the Fondation Mathématiques Jacques Hadamard hosted at Université Paris-Saclay.

\inlinetitle{CRediT contribution}{.} \textbf{Gwendal Debaussart-Joniec}: Conceptualization, Methodology, Investigation, Formal analysis, Writing - original draft, Writing - review \& editing, Visualization, Validation, Software, Data curation; \textbf{Argyris Kalogeratos}: Conceptualization, Methodology, Writing - original draft, Writing - review \& editing, Supervision, Validation, Methodology

\inlinetitle{Code availability}{.}~The code implementing \MDT and all compared methods, as well as the datasets, will all be released in an open-source repository upon publication.

\inlinetitle{Declaration of competing interest}{.}~The authors declare that they have no known competing financial interests or personal relationships that could have appeared to influence the work reported in this paper.

{
\small
\bibliographystyle{apalike}
\bibliography{bib}

\begin{thebibliography}{}

\bibitem[Berline et~al., 1992]{berline_heat_1992}
Berline, N., Getzler, E., and Vergne, M. (1992).
\newblock {\em Heat Kernels and Dirac Operators}.
\newblock Springer, Berlin, Heidelberg.

\bibitem[Butler et~al., 2022]{butler_convolutional_2022}
Butler, L., Parada-Mayorga, A., and Ribeiro, A. (2022).
\newblock Convolutional {Learning} on {Multigraphs}.

\bibitem[Coifman and Lafon, 2006]{coifman_diffusion_2006}
Coifman, R.~R. and Lafon, S. (2006).
\newblock Diffusion maps.
\newblock {\em Applied and Computational Harmonic Analysis}, 21(1):5--30.

\bibitem[Costa et~al., 2005]{costa_discrete_2005}
Costa, O. L.~V., Marques, R.~P., and Fragoso, M.~D. (2005).
\newblock {\em Discrete-Time Markov Jump Linear Systems}.
\newblock Springer.

\bibitem[de~Sa, 2018]{de_sa_spectral_2018}
de~Sa, V.~R. (2018).
\newblock Spectral {Clustering} with {Two} {Views}.
\newblock page~8.

\bibitem[Desgraupes, 2016]{desgraupes_clustering_2016}
Desgraupes, B. (2016).
\newblock Clustering indices.

\bibitem[Dijk et~al., 2018]{dijk_recovering_2018}
Dijk, D.~v., Sharma, R., Nainys, J., Yim, K., Kathail, P., Carr, A.~J., Burdziak, C., Moon, K.~R., Chaffer, C.~L., Pattabiraman, D., Bierie, B., Mazutis, L., Wolf, G., Krishnaswamy, S., and Pe’er, D. (2018).
\newblock Recovering gene interactions from single-cell data using data diffusion.
\newblock {\em Cell}, 174.

\bibitem[Ding and Wu, 2020]{ding_spectral_2020}
Ding, X. and Wu, H.-T. (2020).
\newblock On the spectral property of kernel-based sensor fusion algorithms of high dimensional data.

\bibitem[Gagolewski et~al., 2021]{gagolewski_are_2021}
Gagolewski, M., Bartoszuk, M., and Cena, A. (2021).
\newblock Are cluster validity measures (in) valid?
\newblock {\em Information Sciences}, 581:620--636.

\bibitem[Gleich, 2015]{gleich_pagerank_2014}
Gleich, D.~F. (2015).
\newblock Page{R}ank beyond the web.
\newblock {\em SIAM Review}, 57(3):321--363.

\bibitem[Grigoryan, 2009]{grigoryan_heat_2009}
Grigoryan, A. (2009).
\newblock {\em Heat {Kernel} and {Analysis} on {Manifolds}}.
\newblock {AMS}/{IP} studies in advanced mathematics. American Mathematical Soc.

\bibitem[Grover and Leskovec, 2016]{grover_node2vec_2016}
Grover, A. and Leskovec, J. (2016).
\newblock node2vec: {Scalable} {Feature} {Learning} for {Networks}.

\bibitem[Haghverdi et~al., 2015]{haghverdi_diffusion_2015}
Haghverdi, L., Buettner, F., and Theis, F.~J. (2015).
\newblock {Diffusion maps for high-dimensional single-cell analysis of differentiation data}.
\newblock {\em Bioinformatics}, 31(18):2989--2998.

\bibitem[Hu et~al., 2024]{hu_comprehensive_2024}
Hu, H., Wang, X., Zhang, Y., Chen, Q., and Guan, Q. (2024).
\newblock A comprehensive survey on contrastive learning.
\newblock {\em Neurocomputing}.

\bibitem[Hu et~al., 2012]{hu_survey_2012}
Hu, J., Wang, Y., Zhou, E., Fu, M.~C., and Marcus, S.~I. (2012).
\newblock A survey of some model-based methods for global optimization.
\newblock In Hernández-Hernández, D. and Minjárez-Sosa, J.~A., editors, {\em Optimization, {Control}, and {Applications} of {Stochastic} {Systems}: {In} {Honor} of {Onésimo} {Hernández}-{Lerma}}, pages 157--179. Birkhäuser Boston, Boston.

\bibitem[Huang and Chen, 2022]{huang_random_2022}
Huang, F. and Chen, H. (2022).
\newblock random walks on multiplex networks.
\newblock {\em Journal of Statistical Mechanics: Theory and Experiment}, 2022:103404.

\bibitem[Igel et~al., 2007]{igel_evolutionary_2007}
Igel, C., Hansen, N., and Roth, S. (2007).
\newblock Covariance matrix adaptation for multi-objective optimization.
\newblock {\em Evolutionary Computation}.

\bibitem[Jones, 2001]{jones_direct_2001}
Jones, D.~R. (2001).
\newblock {\em Direct global optimization algorithm}, pages 431--440.
\newblock Springer US.

\bibitem[Katz et~al., 2019]{katz_alternating_2019}
Katz, O., Talmon, R., Lo, Y.-L., and Wu, H.-T. (2019).
\newblock Alternating diffusion maps for multimodal data fusion.
\newblock {\em Information Fusion}, 45.

\bibitem[Kuchroo et~al., 2022]{kuchroo_multimodal_2022}
Kuchroo, M., Godavarthi, A., Tong, A., Wolf, G., and Krishnaswamy, S. (2022).
\newblock Multimodal {Data} {Visualization} and {Denoising} with {Integrated} {Diffusion}.

\bibitem[Langville and Meyer, 2004]{Langville2004Deeper}
Langville, A.~N. and Meyer, C.~D. (2004).
\newblock Deeper inside {P}age{R}ank.
\newblock {\em Internet Mathematics}, 1(3).

\bibitem[Lederman and Talmon, 2018]{lederman_learning_2018}
Lederman, R.~R. and Talmon, R. (2018).
\newblock Learning the geometry of common latent variables using alternating-diffusion.
\newblock {\em Applied and Computational Harmonic Analysis}.

\bibitem[Lederman et~al., 2015]{lederman_alternating_2015}
Lederman, R.~R., Talmon, R., Wu, H.-t., Lo, Y.-L., and Coifman, R.~R. (2015).
\newblock Alternating diffusion for common manifold learning with application to sleep stage assessment.
\newblock In {\em IEEE International Conference on Acoustics, Speech and Signal Processing}.

\bibitem[Lindenbaum et~al., 2020]{lindenbaum_multi-view_2020}
Lindenbaum, O., Yeredor, A., Salhov, M., and Averbuch, A. (2020).
\newblock Multi-view diffusion maps.
\newblock {\em Information Fusion}, 55:127--149.

\bibitem[Martinez-Cantin, 2014]{martinez_bayesopt_2014}
Martinez-Cantin, R. (2014).
\newblock {BayesOpt}: A bayesian optimization library for nonlinear optimization, experimental design and bandits.
\newblock {\em Journal of Machine Learning Research}.

\bibitem[Nadler et~al., 2008]{barth_diffusion_2008}
Nadler, B., Lafon, S., Coifman, R., and Kevrekidis, I.~G. (2008).
\newblock Diffusion maps - a probabilistic interpretation for spectral embedding and clustering algorithms.

\bibitem[Nadler et~al., 2006]{nadler_diffusion_2006}
Nadler, B., Lafon, S., Coifman, R.~R., and Kevrekidis, I.~G. (2006).
\newblock Diffusion maps, spectral clustering and reaction coordinates of dynamical systems.
\newblock {\em Applied and Computational Harmonic Analysis}, 21:113--127.

\bibitem[Pan and Kang, 2021]{pan_multi-view_2021}
Pan, E. and Kang, Z. (2021).
\newblock Multi-view {Contrastive} {Graph} {Clustering}.
\newblock Issue: arXiv:2110.11842 arXiv: 2110.11842 [cs].

\bibitem[Roy et~al., 2020]{roy_learning_2020}
Roy, A., Kumar, V., Mukherjee, D., and Chakraborty, T. (2020).
\newblock Learning {Multigraph} {Node} {Embeddings} {Using} {Guided} {Lévy} {Flights}.
\newblock In {\em Advances in {Knowledge} {Discovery} and {Data} {Mining}}, pages 524--537.

\bibitem[Satopaa et~al., 2011]{satopaa_finding_2011}
Satopaa, V., Albrecht, J., Irwin, D., and Raghavan, B. (2011).
\newblock Finding a "kneedle" in a haystack: Detecting knee points in system behavior.
\newblock In {\em International Conference on Distributed Computing Systems Workshops}.

\bibitem[Seneta, 1981]{seneta_nonnegative_1981}
Seneta, E. (1981).
\newblock {\em Non-negative Matrices and Markov Chains}.
\newblock Springer.

\bibitem[Serré et~al., 2025]{serre_stein_2025}
Serré, G., Kalogeratos, A., and Vayatis, N. (2025).
\newblock Stein boltzmann sampling: A variational approach for global optimization.
\newblock In {\em International Conference on Artificial Intelligence and Statistics}.

\bibitem[Sevi et~al., 2025]{sevi_generalized_2025}
Sevi, H., Debaussart-Joniec, G., Hacini, M., Jonckheere, M., and Kalogeratos, A. (2025).
\newblock Generalized dirichlet energy and graph laplacians for clustering directed and undirected graphs.

\bibitem[Sevi et~al., 2022]{sevi2022clustering}
Sevi, H., Jonckheere, M., and Kalogeratos, A. (2022).
\newblock Clustering for directed graphs using parametrized random walk diffusion kernels.

\bibitem[Shnitzer et~al., 2018]{shnitzer_recovering_2018}
Shnitzer, T., Ben-Chen, M., Guibas, L., Talmon, R., and Wu, H.-T. (2018).
\newblock Recovering hidden components in multimodal data with composite diffusion operators.

\bibitem[Talmon and tieng Wu, 2017]{talmon_latent_2017}
Talmon, R. and tieng Wu, H. (2017).
\newblock Latent common manifold learning with alternating diffusion: analysis and applications.

\bibitem[Valentini et~al., 2023]{valentini_hetnode2vec_2023}
Valentini, G., Casiraghi, E., Cappelletti, L., Fontana, T., Reese, J., and Robinson, P. (2023).
\newblock Het-node2vec: second order random walk sampling for heterogeneous multigraphs embedding.

\bibitem[Vinh et~al., 2009]{vinh_information_2009}
Vinh, N.~X., Epps, J., and Bailey, J. (2009).
\newblock Information theoretic measures for clusterings comparison: is a correction for chance necessary?

\bibitem[Vinh et~al., 2010]{JMLR:v11:vinh10a}
Vinh, N.~X., Epps, J., and Bailey, J. (2010).
\newblock Information theoretic measures for clusterings comparison: Variants, properties, normalization and correction for chance.
\newblock {\em Journal of Machine Learning Research}, 11(95):2837--2854.

\bibitem[Wang et~al., 2012]{wang_unsupervised_2012}
Wang, B., Jiang, J., Wang, W., Zhou, Z.-H., and Tu, Z. (2012).
\newblock Unsupervised metric fusion by cross diffusion.
\newblock In {\em IEEE Conference on Computer Vision and Pattern Recognition}, pages 2997--3004.

\bibitem[Wolfowitz, 1963]{wolfowitz_product_1963}
Wolfowitz, J. (1963).
\newblock Products of indecomposable, aperiodic, stochastic matrices.
\newblock {\em Proceedings of the American Mathematical Society}, 14.

\bibitem[Yeh et~al., 2025]{yeh_landmark_2024}
Yeh, S.-Y., Wu, H.-T., Talmon, R., and Tsui, M.-P. (2025).
\newblock Landmark alternating diffusion.
\newblock {\em SIAM Journal on Mathematics of Data Science}, 7(2):621--642.

\bibitem[Ángela Fernández et~al., 2015]{fernandez_diffusion_2015}
Ángela Fernández, González, A.~M., Díaz, J., and Dorronsoro, J.~R. (2015).
\newblock Diffusion maps for dimensionality reduction and visualization of meteorological data.
\newblock {\em Neurocomputing}, 163:25--37.

\end{thebibliography}
}

\appendix

\section{Proofs} \label{app:proofs}

\noindent \emph{Proof of \cref{prop:W_irreducible}.}
Since $\rmP$ and $\rmQ$ are stochastic matrices, $\rmP \mathbf{1} = \rmQ \mathbf{1} = \mathbf{1}$. Hence $(\rmP\rmQ) \mathbf{1} = \rmP \mathbf{1} = \mathbf{1}$. Which, by definition means that $\rmP \rmQ$ is row-stochastic. Moreover,
\begin{equation*}
  [\rmP \rmQ]_{ij} = \rmP_{ii} \rmQ_{ij} + \Ind{i\neq j}\rmP_{ij} \rmQ_{jj} + \sum_{k\notin \{i,j\}} \rmP_{ik} \rmQ_{kj}, \quad \forall i,j.
\end{equation*}
In particular, this quantity is positive if $\rmQ_{ij}$ or $\rmP_{ij}$ is positive. This means that the product keeps \emph{at least} the connectivity of both $\rmP$ and $\rmQ$, hence it inherits the aperiodicity and the irreducibility properties, which are by definition related to the connectivity of the matrix \citep{seneta_nonnegative_1981}.
\qed
\\ \\
\noindent \emph{Proof of \cref{cor:W_irreducible}.} Let us start by prooving the first part of the corollary. Using \cref{prop:W_irreducible}, we have that for any $t$, $\rmW^{(t)}$ is a product of stochastic, irreducible and aperiodic matrices, thus it is also stochastic, irreducible and aperiodic. Using Perron-Froebenius theorem ensures that the stationary distribution $\pi_t$ exists for any $t$. Moreover, we have that for any $t$, this stationary distribution is unique and satisfies $\pi_t^\top \rmW^{(t)} = \pi_t^\top$.
The second part follows from \citep[Thm.1]{wolfowitz_product_1963}, which states:
\begin{addmargin}[1em]{0em}
  \textbf{Theorem.}
  Let $\gS = \{\rmA_1 \dots \rmA_k\}$ be square stochastic matrices of the same size, such that any product of elements of $\gS$ is stochastic, irreducible and aperiodic. Then, for any $\epsilon \geq 0$ there exist $N_\epsilon$ such that for any matrix product $\rmP$ of length $n \geq N_\epsilon$
  \begin{equation}
    \max_j \max_{i_1, i_2} \norm{\rmP_{i_1 j} - \rmP_{i_2 j}} \leq \epsilon.
  \end{equation}
\end{addmargin}
Indeed, using \cref{prop:W_irreducible} gives us that every product of elements of $\Pset$ remains stochastic irreducible and aperiodic. Using this theorem then gives us the fact that that the limit $\rmW^{(\infty)}$ exists and is such that $\rmW^{(\infty)}_{i_1 j} = \rmW^{(\infty)}_{i_2 j}$ for every $i_1, i_2$. In particular, this is a definition of an ergodic law for inhomogeneous Markov chains (see e.g \citep{seneta_nonnegative_1981}). And we there exists $\pi_\infty$ such that $\rmW^{(\infty)}=\mathbf{1}\pi_\infty^\top$.
\qed
\begin{table*}[t!]
  \centering
  \begin{adjustbox}{width=\linewidth,center}
\begin{tabular}{l|ccccccccc}
\Xhline{1pt}
\multirow{2}{*}{\textbf{Methods}} & \multicolumn{9}{c}{\textbf{Datasets}}  \\
\cline{2-10}
&
\textbf{\textsc{K-MvMNIST}} & \textbf{\textsc{L-MvMNIST}} & \textbf{\textsc{Olivetti}} & \textbf{\textsc{Yale}} & \textbf{\textsc{100Leaves}} & \textbf{\textsc{L-Isolet}} & \textbf{\textsc{MSRC}} & \textbf{\textsc{Multi-Feat}} & \textbf{\textsc{Caltech101-7}} \\
\Xhline{0.75pt}
True Labels & 6.64 & 5.75 & 5.35 & 3.94 & \mybest{5.87} & 6.51 & \mysecond{3.72} & 5.33 & 3.05 \\
\Xhline{0.25pt}
\textcolor{crdcol}{\rule{1em}{0.75em}} \CrD \ \ \ \ \ \citep{wang_unsupervised_2012} \smallg{(2012)} & 6.92 \mystd{0.10} & 6.02 \mystd{0.14} & 5.41 \mystd{0.08} & 4.80 \mystd{0.19} & 2.16 \mystd{0.04} & 6.91 \mystd{0.17} & 2.69 \mystd{0.09} & 4.92 \mystd{0.20} & 3.55 \mystd{0.10}  \\
\textcolor{comdcol}{\rule{1em}{0.75em}}  \ComD\ \ \citep{shnitzer_recovering_2018} \smallg{(2018)} & 7.03 \mystd{0.07} & \mysecond{6.27 \mystd{0.06}} & \mybest{6.03 \mystd{0.03}} & — & — & — & — & — & —  \\
\textcolor{adcol}{\rule{1em}{0.75em}}  \ADM  \ \ \ \ \ \ \ \ \ \ \ \ \ \ \citep{katz_alternating_2019} \smallg{(2019)} & 7.02 \mystd{0.06} & 6.26 \mystd{0.06} & \mysecond{6.00 \mystd{0.04}} & \mysecond{5.16 \mystd{0.04}} & 5.16 \mystd{0.06} & \mybest{7.84 \mystd{0.05}} & 2.96 \mystd{0.07} & 5.40 \mystd{0.04} & 3.53 \mystd{0.04}  \\
\textcolor{mvdmcol}{\rule{1em}{0.75em}}  \MVDM \ \ \ \ \ \ \ \ \ \ \citep{lindenbaum_multi-view_2020} \smallg{(2020)} & 7.03 \mystd{0.06} & 6.24 \mystd{0.05} & 5.97 \mystd{0.03} & 5.12 \mystd{0.05} & 5.41 \mystd{0.04} & 2.84 \mystd{0.03} & 3.02 \mystd{0.07} & 5.40 \mystd{0.05} & 3.52 \mystd{0.05} \\
\textcolor{idmcol}{\rule{1em}{0.75em}}  \IDM \ \ \ \ \ \ \ \ \ \ \ \, \,  \ \citep{kuchroo_multimodal_2022} \smallg{(2022)} & 7.04 \mystd{0.04} & 6.20 \mystd{0.07} & 5.92 \mystd{0.06} & 4.96 \mystd{0.09} & 4.98 \mystd{0.05} & 7.42 \mystd{0.11} & 2.54 \mystd{0.07} & 5.22 \mystd{0.08} & 3.29 \mystd{0.08}  \\
\textcolor{padcol}{\rule{1em}{0.75em}}  \textsc{p-AD} \, \, \, \, \,   \,\citep{kuchroo_multimodal_2022} \smallg{(2022)} & — & — & 5.95 \mystd{0.06} & 5.14 \mystd{0.07} & 4.07 \mystd{0.08} & 7.65 \mystd{0.09} & 2.50 \mystd{0.08} & 5.24 \mystd{0.13} & 3.51 \mystd{0.06}  \\
\Xhline{0.75pt}
\textcolor{mdtchsdcol}{\rule{1em}{0.75em}}  \MDTchs & \mybest{7.06 \mystd{0.04}} & \mybest{6.28 \mystd{0.06}} & 5.96 \mystd{0.06} & \mybest{5.20 \mystd{0.04}} & \mysecond{5.45 \mystd{0.06}} & \mysecond{7.75 \mystd{0.09}} & \mybest{3.76 \mystd{0.07}} & \mybest{5.43 \mystd{0.03}} & \mybest{3.61 \mystd{0.04}} \\
\textcolor{mdtbscdcol}{\rule{1em}{0.75em}}  \MDTbsc & 7.03 \mystd{0.06} & 6.24 \mystd{0.06} & 5.89 \mystd{0.07} & 5.12 \mystd{0.09} & 4.37 \mystd{0.05} & 7.47 \mystd{0.12} & 3.59 \mystd{0.08} & \mybest{5.43 \mystd{0.05}} & 3.58 \mystd{0.06}  \\
\textcolor{mdtrandcol}{\rule{1em}{0.75em}}  \MDTrand & 7.00 \mystd{0.06} & 6.24 \mystd{0.06} & 5.94 \mystd{0.06} & 5.08 \mystd{0.04} & 4.64 \mystd{0.23} & 7.74 \mystd{0.08} & 3.42 \mystd{0.32} & 5.24 \mystd{0.14} &  3.52 \mystd{0.10 }\\
\textcolor{mdtconvcol}{\rule{1em}{0.75em}} \MDTcvx & \mysecond{7.05 \mystd{0.04}} & \mysecond{6.27 \mystd{0.05}} & 5.97 \mystd{0.06} & 5.15 \mystd{0.04} & 5.19 \mystd{0.20} & \mysecond{7.75 \mystd{0.09}}& 3.70 \mystd{0.11} & \mysecond{5.41 \mystd{0.07}} & \mysecond{3.60 \mystd{0.05}} \\
\Xhline{1pt}
\end{tabular}
  \end{adjustbox}
  \caption{\textbf{Clustering results -- CH across multiple datasets.} For each column, the highest and second-highest mean CH scores are shown in bold and underlined, respectively. Standard deviations are given in parentheses. `—' indicates that \ComD results are not available for those datasets, due to the two-view limitation of the method.}
  \label{tab:ch_score}
\end{table*}

\noindent \emph{Proof of \cref{prop:diff_dist_map}.}
Let $t \in \sN^*$, define $\Wnorm^{(t)} \defeq \rmPi_t^{1/2} \rmW^{(t)} \rmPi_t^{-1/2}$ and its SVD $\Wnorm^{(t)} = \rmM \rmLambda \rmN$ and denote $\rmW^{(t)} = \rmU_t \mSigma_t \rmV_t^\top$ the svd of $\rmW^{(t)}$. Note that the SVD of $\Wnorm$ and the one of $\rmW$ are related as follow: $\mLambda = \mSigma$, $\rmU = \rmPi_t^{-1/2} \rmM$ and $\rmV^\top = \rmN^\top \rmPi_t^{-1/2}$.
  \begin{equation}
    \begin{split}
      \gD_{\rmW^{(t)}}(x_i,x_j) &= \norm{(e_i - e_j)^\top \rmW^{(t)}}_{1/\pi_\iter} \\
      &= \norm{(e_i - e_j)^\top \rmW^{(t)} \rmPi_\iter^{-1/2}}_2 \\
      &= \norm{(e_i - e_j)^\top\rmPi_\iter^{-1/2}\Wnorm^{(t)}}_2 \\
      &= \norm{(e_i - e_j)^\top\rmPi_\iter^{-1/2} \rmM \rmSigma_\iter \rmN^\top}_2 \\
      &= \norm{(e_i - e_j)^\top \rmU_\iter \rmSigma_\iter}_2 \\
      &= \norm{(e_i - e_j)^\top\rmPsi^{(t)}}_2.
    \end{split}
  \end{equation}
  The $4$-th equation comes from the fact that, by definition of an SVD, $\rmN$ is semi-orthogonal matrix, thus $\norm{x^\top \rmN^\top}_2 = \norm{x}_2$ by the isometry property.
\qed

\section{Details for the clustering evaluation} \label{app:datasets}

\begin{figure}[t!]
\centering
\begin{subfigure}[t]{0.22\textwidth}
    \centering
    \includegraphics[width=\textwidth]{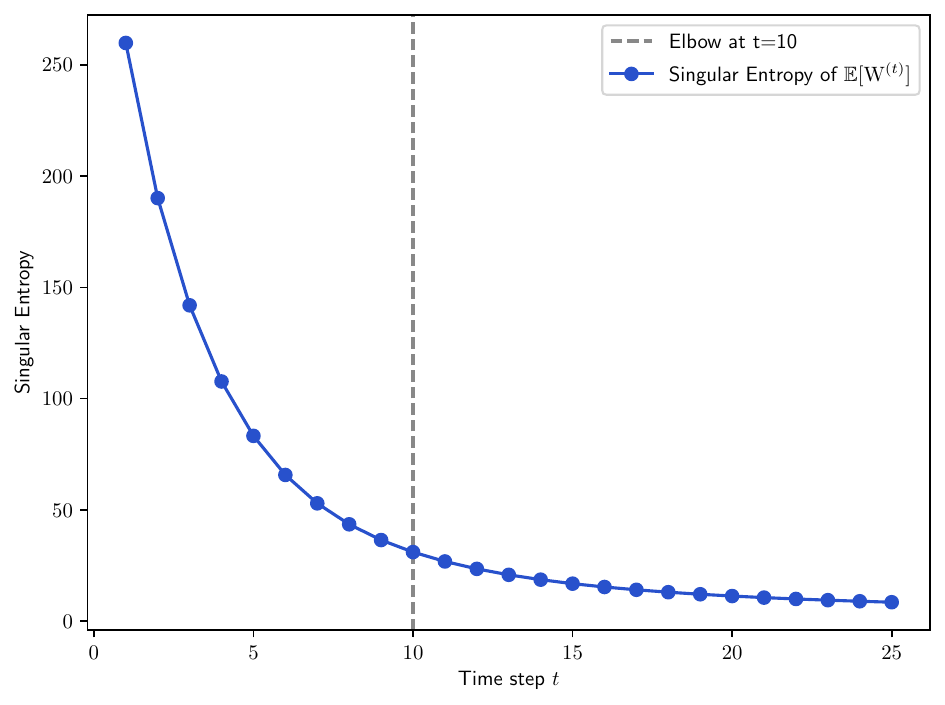}
    \caption{K-MvMNIST}
  \end{subfigure}
  \begin{subfigure}[t]{0.22\textwidth}
    \centering
    \includegraphics[width=\textwidth]{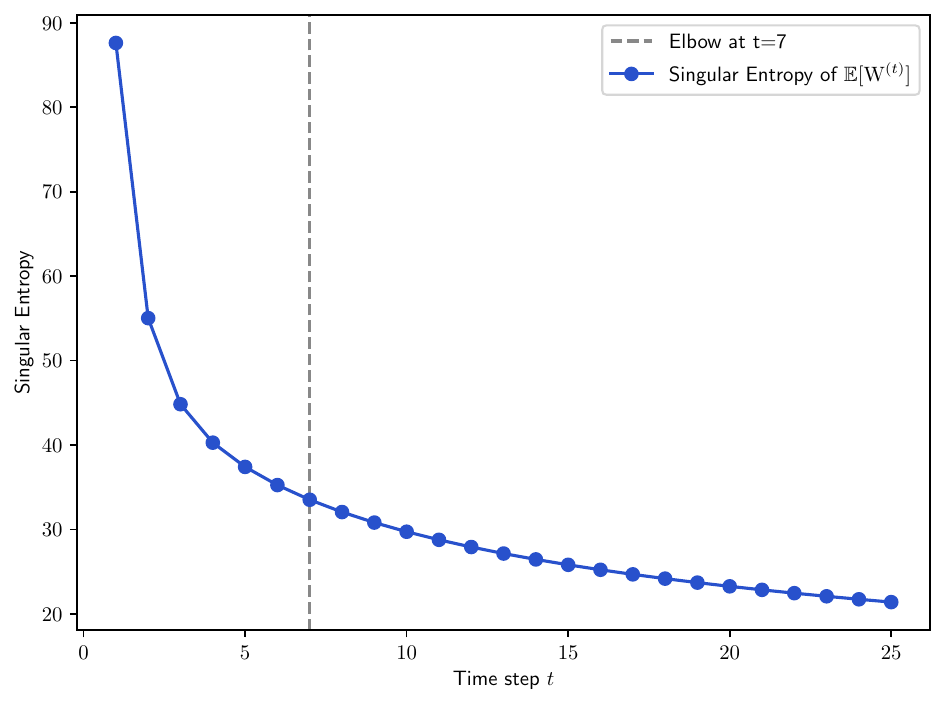}
    \caption{Helix A}
  \end{subfigure}
  \caption{\textbf{Singular entropy curves.} The curves refer to the expected random trajectory on different datasets. The selected $t$ parameter (elbow point) is indicated by a vertical dashed line.}
  \label{fig:singular_entropy}
\end{figure}

\inlinetitle{Singular entropy of the expected random trajectory}{.}
As we use the elbow of the singular entropy curve to select the $t$ parameter, we provide in \cref{fig:singular_entropy} the singular entropy curves for the expected random trajectory on two different datasets used in \cref{sec:experiments}. The same exponential decay behavior is observed on all datasets, and the selected $t$ parameter (elbow point) is indicated by a vertical dashed line.

\begin{figure*}[h!]
  \centering
  \begin{subfigure}[b]{0.30\textwidth}
    \centering
    \includegraphics[width=\linewidth]{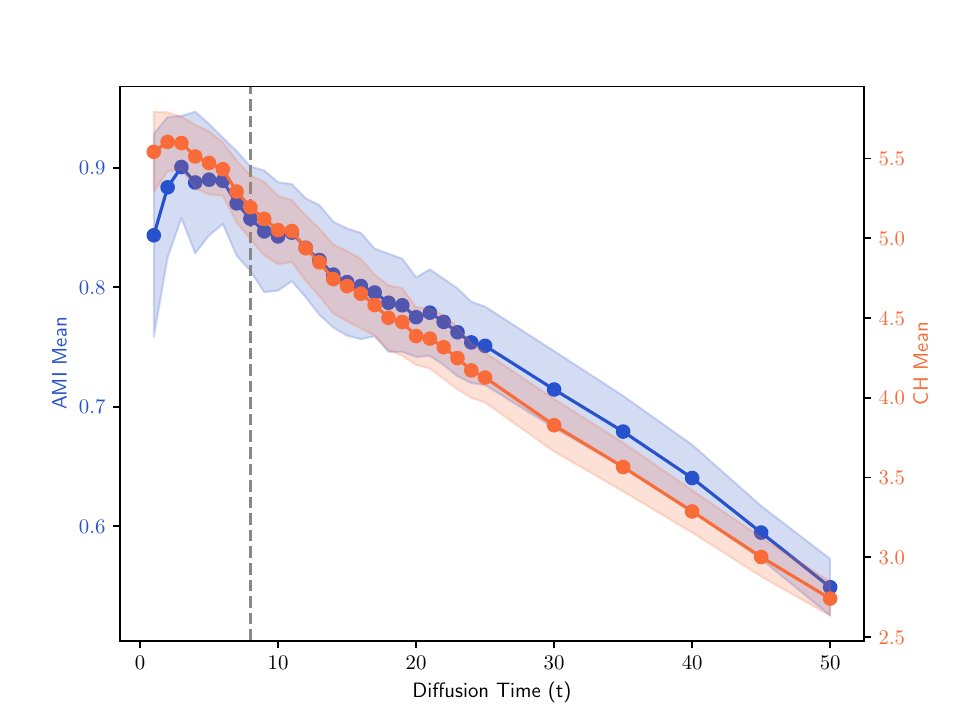}
    \caption{100Leaves}
  \end{subfigure} \hfill%
  \begin{subfigure}[b]{0.30\textwidth}
    \centering
    \includegraphics[width=\linewidth]{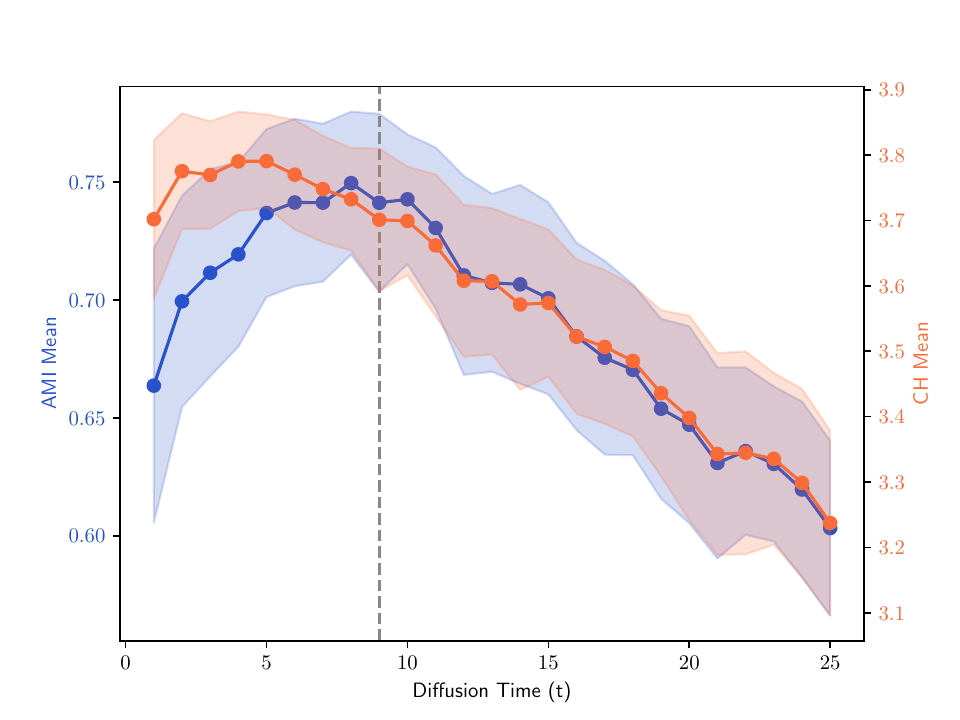}
    \caption{MSRC}
  \end{subfigure} \hfill
  \begin{subfigure}[b]{0.30\textwidth}
    \centering
    \includegraphics[width=\linewidth]{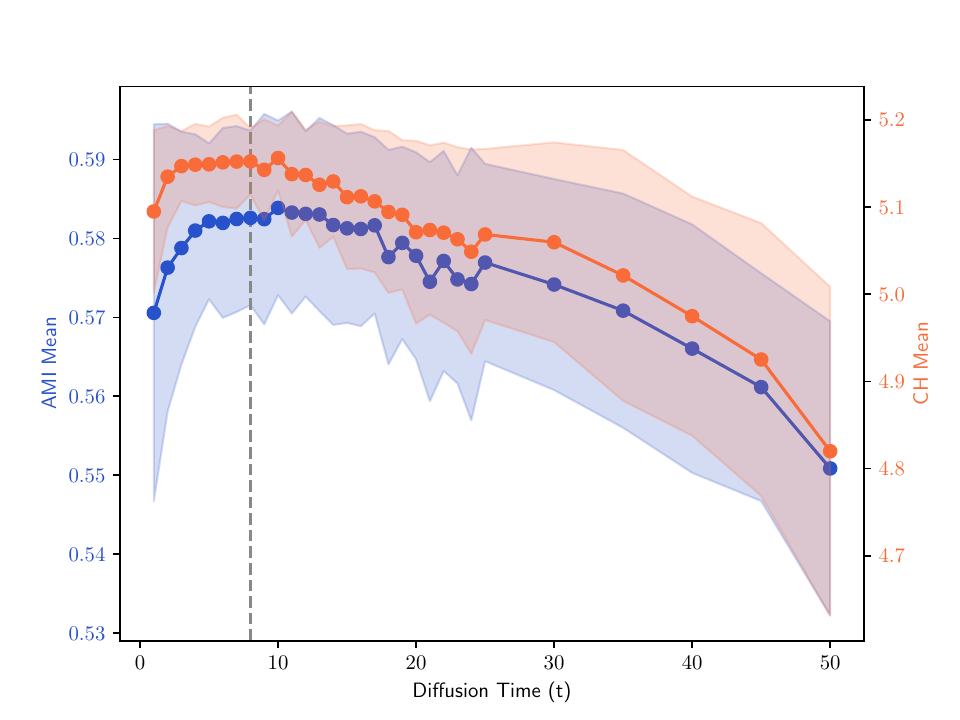}
    \caption{Yale}
  \end{subfigure} \hfill \\
  \begin{subfigure}[b]{0.30\textwidth}
    \centering
    \includegraphics[width=\linewidth]{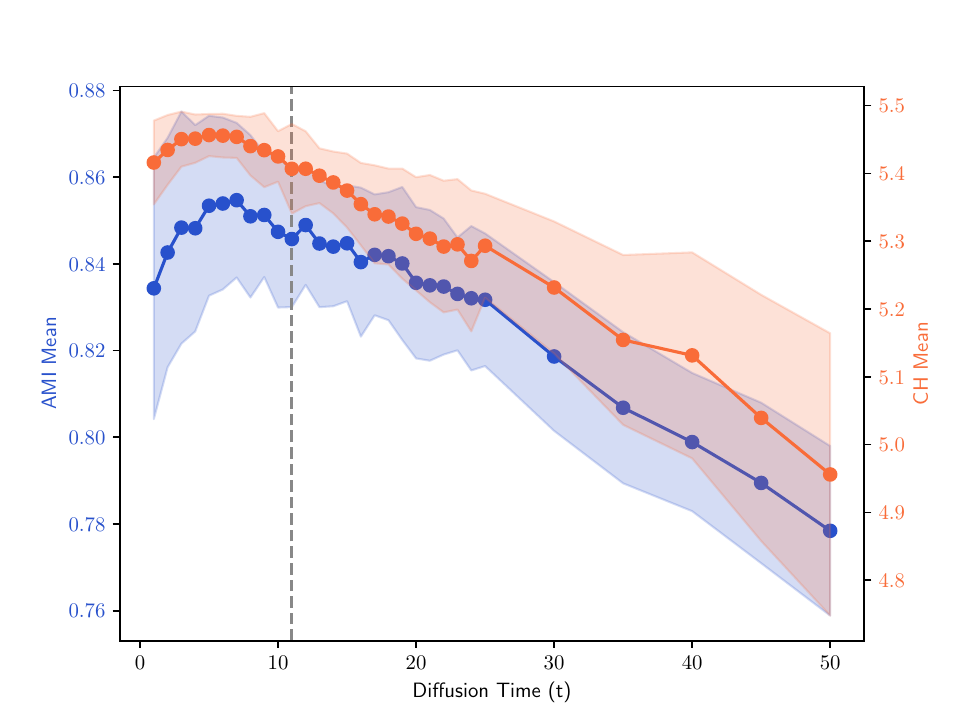}
    \caption{Multi-Feat}
  \end{subfigure} \hfill%
  \begin{subfigure}[b]{0.30\textwidth}
    \centering
    \includegraphics[width=\linewidth]{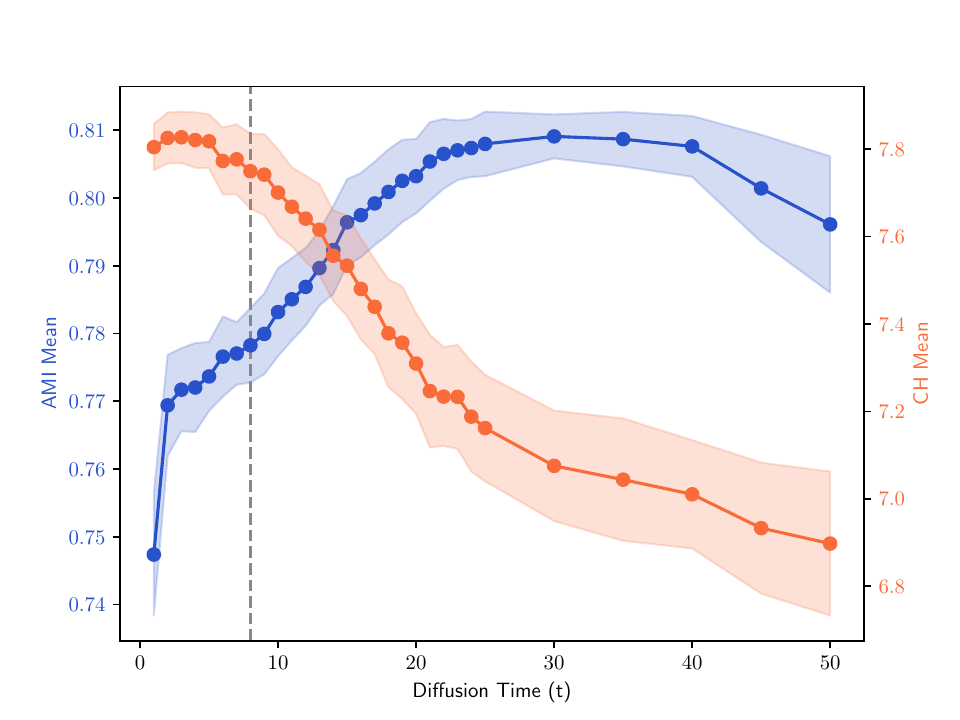}
    \caption{L-Isolet}
  \end{subfigure} \hfill%
  \begin{subfigure}[b]{0.30\textwidth}
    \centering
    \includegraphics[width=\linewidth]{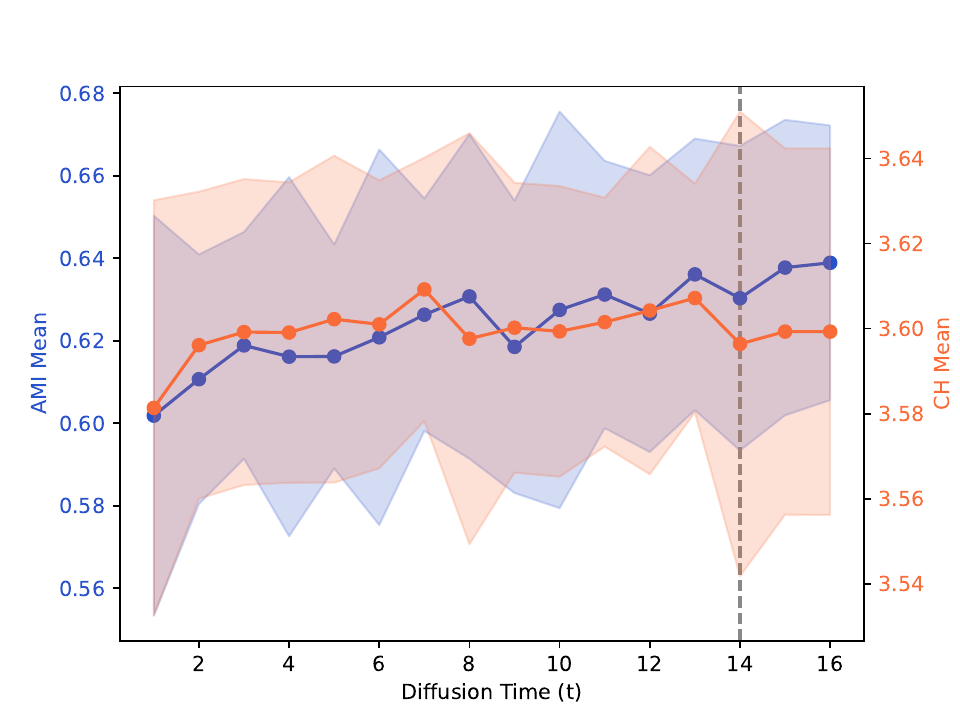}
    \caption{Caltech101-7}
  \end{subfigure} \hfill \\
  \caption{\textbf{Clustering results -- AMI and CH in different datasets.} Each point corresponds to an \MDTcvx trajectory length ($t$), with its associated mean taken over $100$ randomly sampled \MDTs, standard deviation is shown in colored bands around the mean.}
  \label{fig:ch_score}
\end{figure*}

\inlinetitle{Correlation between internal criterion and clustering quality}{.}~%
We examine two aspects of the relationship between the internal criterion (CH index) and clustering quality as measured by AMI. First, \cref{tab:ch_score} reports the CH scores for the ground-truth labels alongside those produced by the evaluated methods, allowing inspecting how closely the internal criterion matches the actual clustering structure. Second, \cref{fig:ch_score} shows how the CH and AMI vary with trajectory length, where we plot both metrics for randomly selected \MDTs in $\Psetcvx$ and track how their averages evolve over time. This showcases how optimizing CH may impact AMI, and reveals how \MDTs' clustering quality is affected by the choice of trajectory length. The figure also marks the time parameter adopted in \cref{sec:experiments:clustering}, which typically falls in a region where both AMI and CH are high. Although not always optimal, this tuning is a reasonable compromise that clearly avoids ranges associated with low AMI.

\begin{figure*}[t]
  \centering
  \begin{subfigure}[t]{0.25\textwidth}
    \centering
		\includegraphics[width=\textwidth]{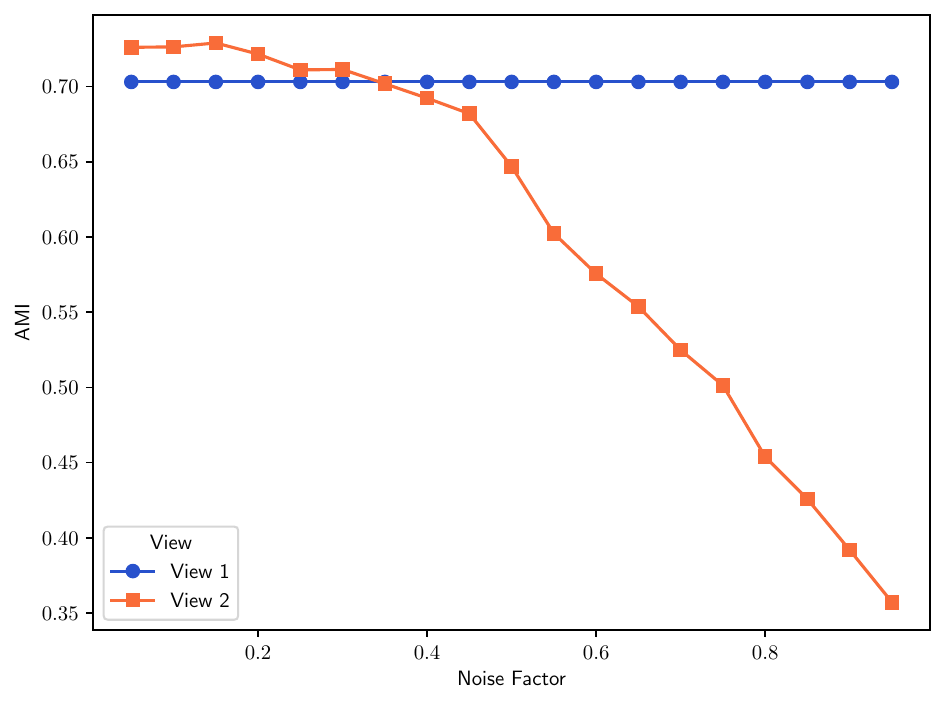}
    \caption{K-MvMNIST}
    \label{fig:viewperf:Kmnist}
  \end{subfigure}\hfill
  \begin{subfigure}[t]{0.25\textwidth}
    \centering
		\includegraphics[width=\textwidth]{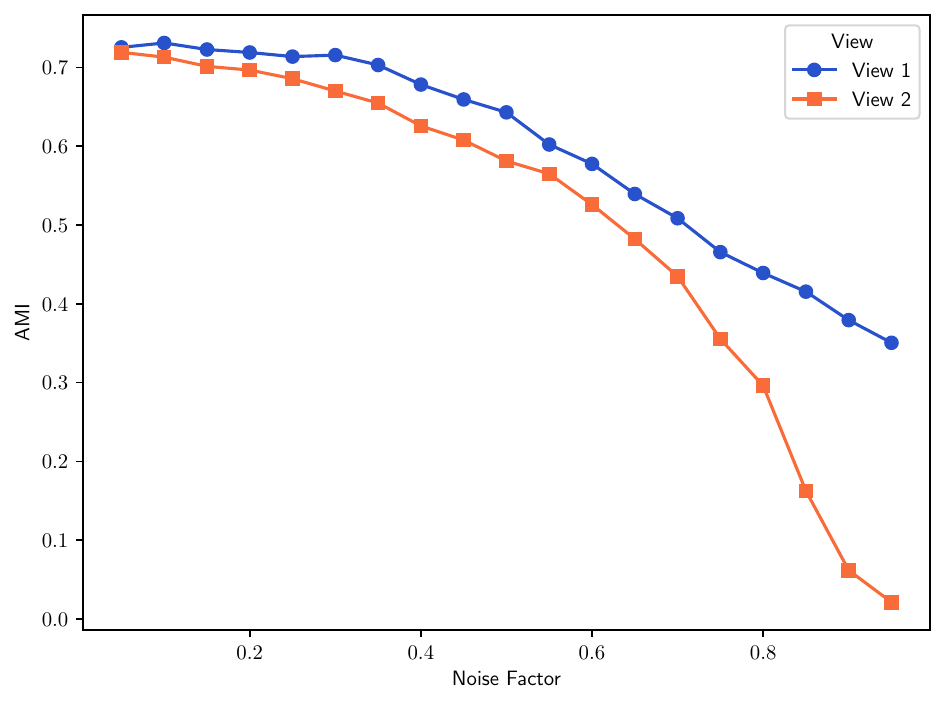}
    \caption{L-MvMNIST}
    \label{fig:viewperf:Lmnist}
  \end{subfigure}\hfill
  \begin{subfigure}[t]{0.25\textwidth}
    \centering
		\includegraphics[width=\textwidth]{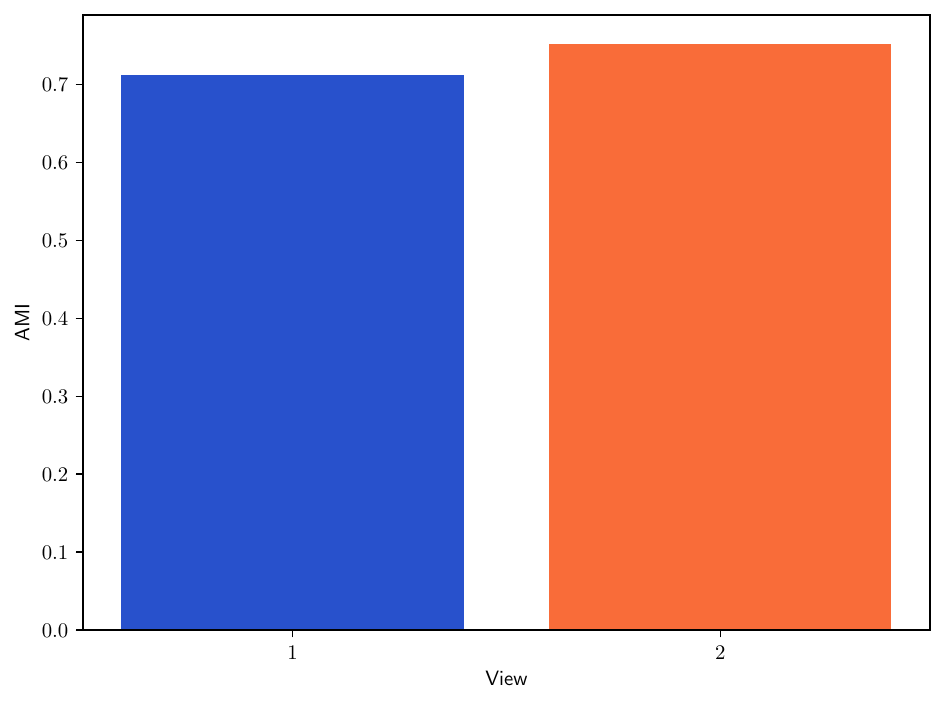}
    \caption{Olivetti}
    \label{fig:viewperf:olivetti}
  \end{subfigure}\hfill \\
  \begin{subfigure}[t]{0.25\textwidth}
    \centering
		\includegraphics[width=\textwidth]{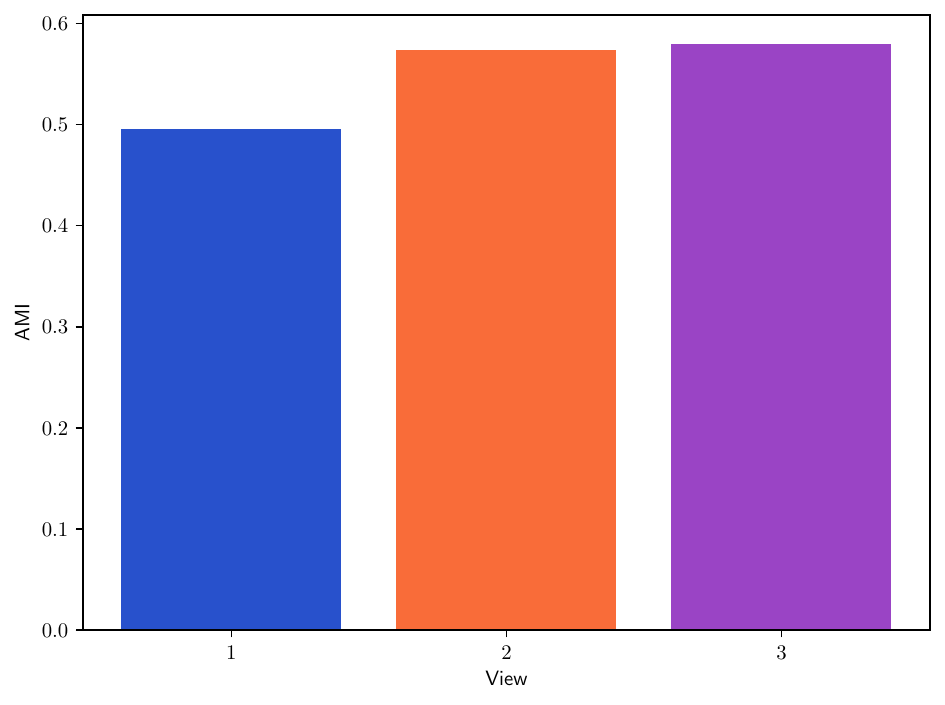}
    \caption{Yale}
    \label{fig:viewperf:yale}
  \end{subfigure} \hfill
  \begin{subfigure}[t]{0.25\textwidth}
    \centering
		\includegraphics[width=\textwidth]{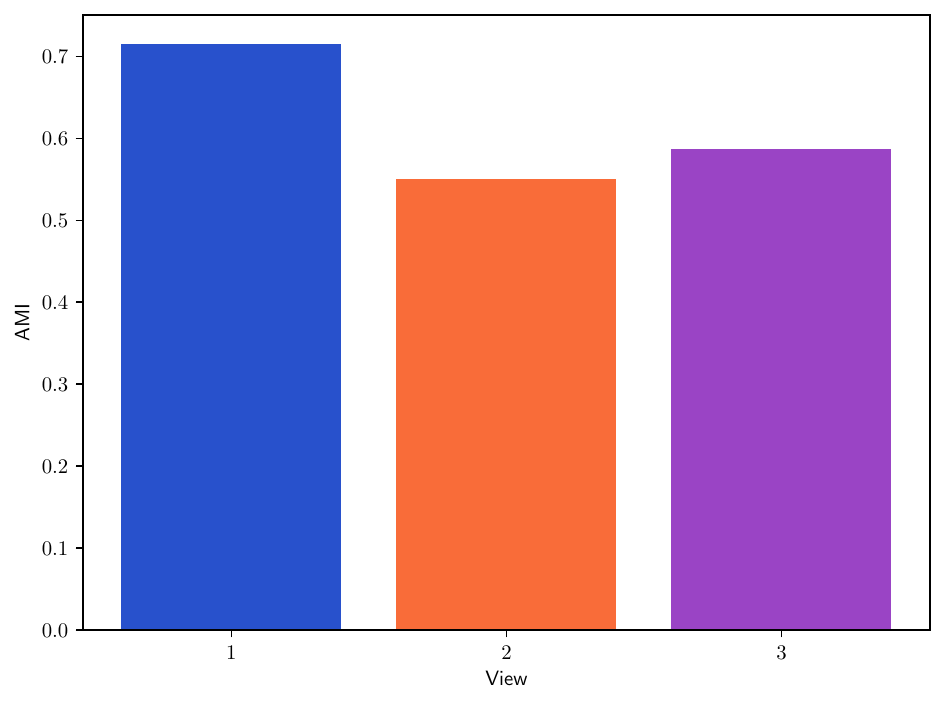}
    \caption{100Leaves}
    \label{fig:viewperf:leaves}
  \end{subfigure} \hfill
  \begin{subfigure}[t]{0.25\textwidth}
    \centering
		\includegraphics[width=\textwidth]{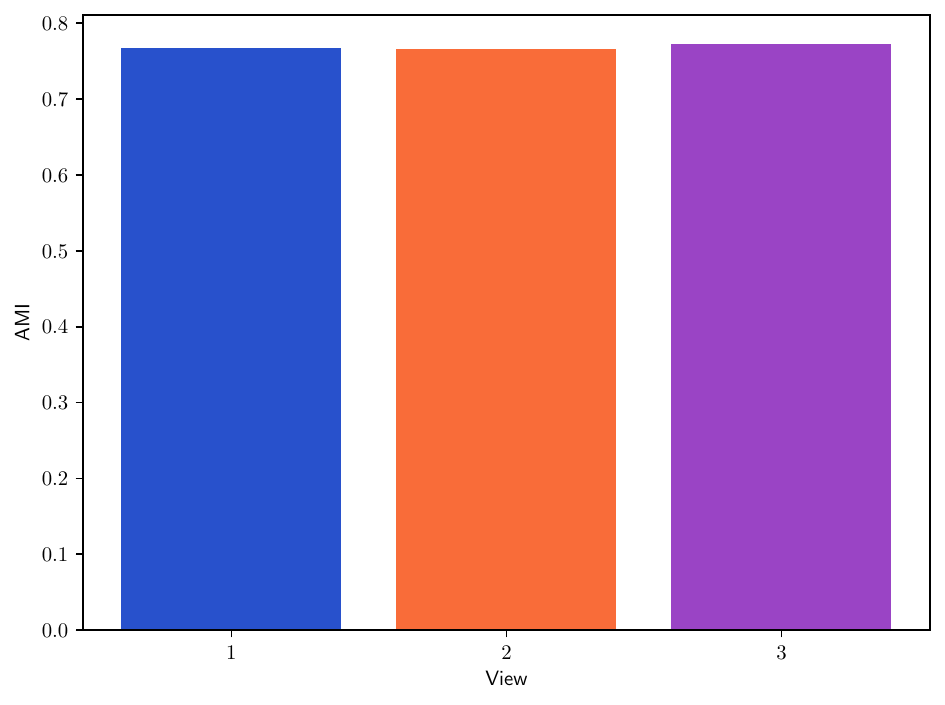}
    \caption{L-Isolet}
    \label{fig:viewperf:lisolet}
  \end{subfigure}\hfill \\
  \begin{subfigure}[t]{0.24\textwidth}
    \centering
		\includegraphics[width=\textwidth]{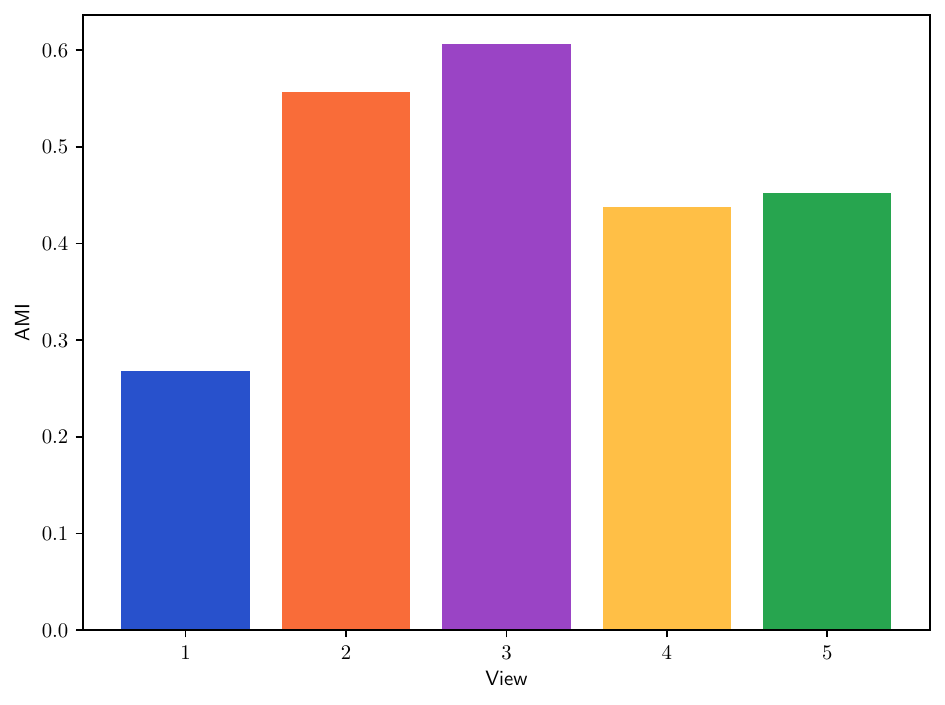}
    \caption{MSRC}
    \label{fig:viewperf:msrc}
  \end{subfigure}\hfill%
  \begin{subfigure}[t]{0.25\textwidth}
    \centering
		\includegraphics[width=\textwidth]{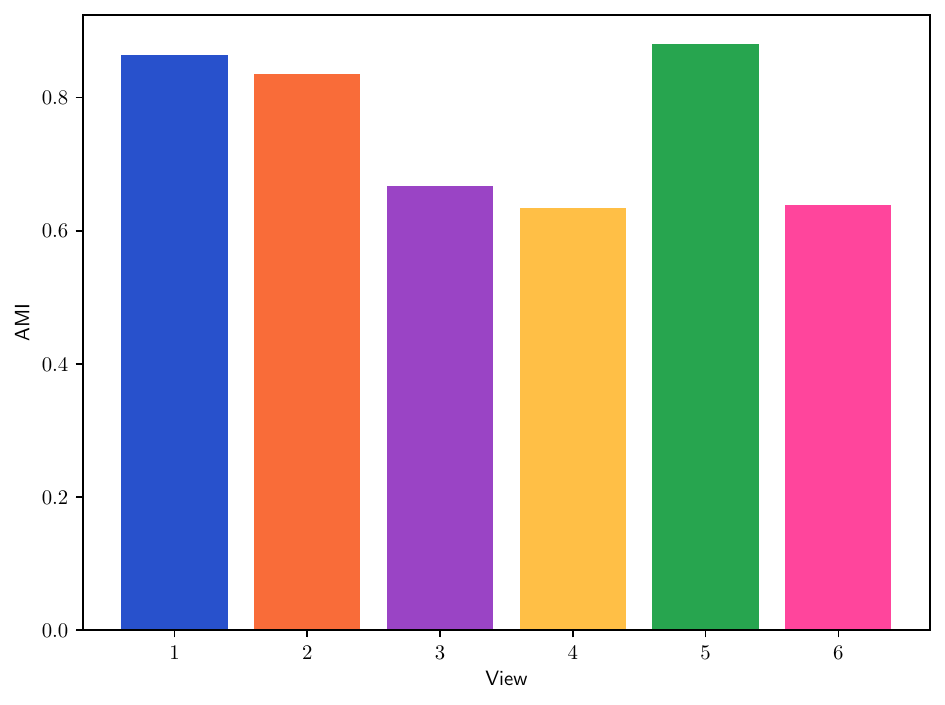}
    \caption{Multi-Feat}
    \label{fig:viewperf:multiple_feat}
  \end{subfigure}\hfill
  \begin{subfigure}[t]{0.25\textwidth}
    \centering
		\includegraphics[width=\textwidth]{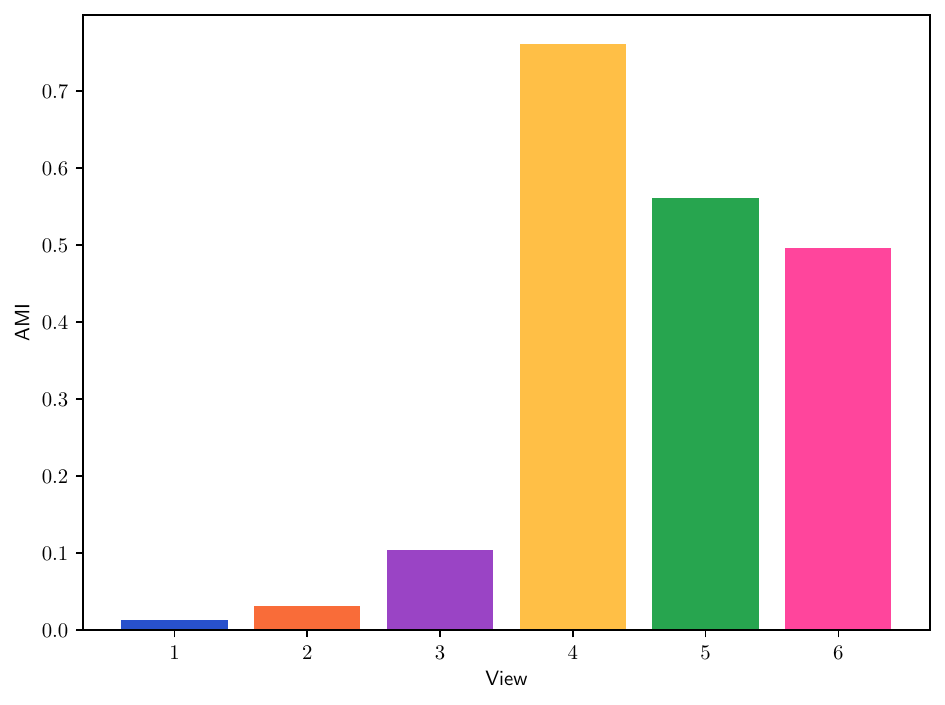}
    \caption{Caltech101-7}
    \label{fig:viewperf:caltech101-7}
  \end{subfigure}\hfill
  \caption{\textbf{Quality of individual views in the context of clustering performance.} AMI using each individual view separately on different datasets. (a) and (b) show the results for K-MvMNIST and L-MvMNIST according to the `noise-factor' $s$.}
  \label{fig:viewperf}
\end{figure*}

\inlinetitle{Reference clustering quality per view}{.} \label{app:viewperf}
To better understand the clustering performance of different methods on multi-view datasets, we report in \cref{fig:viewperf} clustering results (AMI) achieved when using each individual view separately. This analysis provides insights about how informative each view is and how it may contribute to the overall clustering performance for each method.
In particular, we can see that the Caltech101 dataset admits a strong heterogeneity in view quality, with view $1$-$3$ being significantly less informative. 
Similarly, in K-MvMNIST and L-MvMNIST (\cref{fig:viewperf:Kmnist,fig:viewperf:Lmnist}), we can see how the noise-factor $s$ affects the quality of the second view, which in turn impacts the overall clustering performance of methods that rely on both views equivalently. In the MSRC dataset (\cref{fig:viewperf:msrc}), view $1$ seems less informative than the others. In Olivetti, L-Isolet, Yale, and 100Leaves datasets (\cref{fig:viewperf:olivetti,fig:viewperf:lisolet,fig:viewperf:yale,fig:viewperf:leaves}), all views seem to provide similar clustering quality.

\end{document}